\documentclass[10pt]{article}

\usepackage{booktabs}

\usepackage{graphicx}

\usepackage{mathtools}
\usepackage{amsmath}
\usepackage{amssymb}
\usepackage{bm}

\DeclarePairedDelimiter{\ceil}{\lceil}{\rceil}

\usepackage{times}
\usepackage[english]{babel}

\usepackage{authblk}
\usepackage{amsmath}
\usepackage{amssymb}
\usepackage{amsthm}
\usepackage{comment}
\usepackage{graphicx}
\usepackage{subcaption}
\usepackage{hyperref}

\usepackage{tikz}
\usetikzlibrary{spy,calc}

\usepackage[mathscr]{euscript}

\usepackage[]{algorithm2e}

\usepackage[thinc]{esdiff}

\usepackage{float}

\newif\ifblackandwhitecycle
\gdef\patternnumber{0}

\pgfkeys{/tikz/.cd,
    zoombox paths/.style={
        draw=orange,
        very thick
    },
    black and white/.is choice,
    black and white/.default=static,
    black and white/static/.style={ 
        draw=white,   
        zoombox paths/.append style={
            draw=white,
            postaction={
                draw=black,
                loosely dashed
            }
        }
    },
    black and white/static/.code={
        \gdef\patternnumber{1}
    },
    black and white/cycle/.code={
        \blackandwhitecycletrue
        \gdef\patternnumber{1}
    },
    black and white pattern/.is choice,
    black and white pattern/0/.style={},
    black and white pattern/1/.style={    
            draw=white,
            postaction={
                draw=black,
                dash pattern=on 2pt off 2pt
            }
    },
    black and white pattern/2/.style={    
            draw=white,
            postaction={
                draw=black,
                dash pattern=on 4pt off 4pt
            }
    },
    black and white pattern/3/.style={    
            draw=white,
            postaction={
                draw=black,
                dash pattern=on 4pt off 4pt on 1pt off 4pt
            }
    },
    black and white pattern/4/.style={    
            draw=white,
            postaction={
                draw=black,
                dash pattern=on 4pt off 2pt on 2 pt off 2pt on 2 pt off 2pt
            }
    },
    zoomboxarray inner gap/.initial=5pt,
    zoomboxarray columns/.initial=2,
    zoomboxarray rows/.initial=2,
    subfigurename/.initial={},
    figurename/.initial={zoombox},
    zoomboxarray/.style={
        execute at begin picture={
            \begin{scope}[
                spy using outlines={%
                    zoombox paths,
                    width=\imagewidth / \pgfkeysvalueof{/tikz/zoomboxarray columns} - (\pgfkeysvalueof{/tikz/zoomboxarray columns} - 1) / \pgfkeysvalueof{/tikz/zoomboxarray columns} * \pgfkeysvalueof{/tikz/zoomboxarray inner gap} -\pgflinewidth,
                    height=\imageheight / \pgfkeysvalueof{/tikz/zoomboxarray rows} - (\pgfkeysvalueof{/tikz/zoomboxarray rows} - 1) / \pgfkeysvalueof{/tikz/zoomboxarray rows} * \pgfkeysvalueof{/tikz/zoomboxarray inner gap}-\pgflinewidth,
                    magnification=3,
                    every spy on node/.style={
                        zoombox paths
                    },
                    every spy in node/.style={
                        zoombox paths
                    }
                }
            ]
        },
        execute at end picture={
            \end{scope}
            \node at (image.north) [anchor=north,inner sep=0pt] {\subcaptionbox{\label{\pgfkeysvalueof{/tikz/figurename}-image}}{\phantomimage}};
            \node at (zoomboxes container.north) [anchor=north,inner sep=0pt] {\subcaptionbox{\label{\pgfkeysvalueof{/tikz/figurename}-zoom}}{\phantomimage}};
     \gdef\patternnumber{0}
        },
        spymargin/.initial=0.5em,
        zoomboxes xshift/.initial=1,
        zoomboxes right/.code=\pgfkeys{/tikz/zoomboxes xshift=1},
        zoomboxes left/.code=\pgfkeys{/tikz/zoomboxes xshift=-1},
        zoomboxes yshift/.initial=0,
        zoomboxes above/.code={
            \pgfkeys{/tikz/zoomboxes yshift=1},
            \pgfkeys{/tikz/zoomboxes xshift=0}
        },
        zoomboxes below/.code={
            \pgfkeys{/tikz/zoomboxes yshift=-1},
            \pgfkeys{/tikz/zoomboxes xshift=0}
        },
        caption margin/.initial=4ex,
    },
    adjust caption spacing/.code={},
    image container/.style={
        inner sep=0pt,
        at=(image.north),
        anchor=north,
        adjust caption spacing
    },
    zoomboxes container/.style={
        inner sep=0pt,
        at=(image.north),
        anchor=north,
        name=zoomboxes container,
        xshift=\pgfkeysvalueof{/tikz/zoomboxes xshift}*(\imagewidth+\pgfkeysvalueof{/tikz/spymargin}),
        yshift=\pgfkeysvalueof{/tikz/zoomboxes yshift}*(\imageheight+\pgfkeysvalueof{/tikz/spymargin}+\pgfkeysvalueof{/tikz/caption margin}),
        adjust caption spacing
    },
    calculate dimensions/.code={
        \pgfpointdiff{\pgfpointanchor{image}{south west} }{\pgfpointanchor{image}{north east} }
        \pgfgetlastxy{\imagewidth}{\imageheight}
        \global\let\imagewidth=\imagewidth
        \global\let\imageheight=\imageheight
        \gdef\columncount{1}
        \gdef\rowcount{1}
        
    },
    image node/.style={
        inner sep=0pt,
        name=image,
        anchor=south west,
        append after command={
            [calculate dimensions]
            node [image container,subfigurename=\pgfkeysvalueof{/tikz/figurename}-image] {\phantomimage}
            node [zoomboxes container,subfigurename=\pgfkeysvalueof{/tikz/figurename}-zoom] {\phantomimage}
        }
    },
    color code/.style={
        zoombox paths/.append style={draw=#1}
    },
    connect zoomboxes/.style={
    spy connection path={\draw[draw=none,zoombox paths] (tikzspyonnode) -- (tikzspyinnode);}
    },
    help grid code/.code={
        \begin{scope}[
                x={(image.south east)},
                y={(image.north west)},
                font=\footnotesize,
                help lines,
                overlay
            ]
            \foreach \x in {0,1,...,9} { 
                \draw(\x/10,0) -- (\x/10,1);
                \node [anchor=north] at (\x/10,0) {0.\x};
            }
            \foreach \y in {0,1,...,9} {
                \draw(0,\y/10) -- (1,\y/10);                        \node [anchor=east] at (0,\y/10) {0.\y};
            }
        \end{scope}    
    },
    help grid/.style={
        append after command={
            [help grid code]
        }
    },
}

\newcommand\phantomimage{%
    \phantom{%
        \rule{\imagewidth}{\imageheight}%
    }%
}
\newcommand\zoombox[2][]{
    \begin{scope}[zoombox paths]
        \pgfmathsetmacro\xpos{
            (\columncount-1)*(\imagewidth / \pgfkeysvalueof{/tikz/zoomboxarray columns} + \pgfkeysvalueof{/tikz/zoomboxarray inner gap} / \pgfkeysvalueof{/tikz/zoomboxarray columns} ) + \pgflinewidth
        }
        \pgfmathsetmacro\ypos{
            (\rowcount-1)*( \imageheight / \pgfkeysvalueof{/tikz/zoomboxarray rows} + \pgfkeysvalueof{/tikz/zoomboxarray inner gap} / \pgfkeysvalueof{/tikz/zoomboxarray rows} ) + 0.5*\pgflinewidth
        }
        \edef\dospy{\noexpand\spy [
            #1,
            zoombox paths/.append style={
                black and white pattern=\patternnumber
            },
            every spy on node/.append style={#1},
            x=\imagewidth,
            y=\imageheight
        ] on (#2) in node [anchor=north west] at ($(zoomboxes container.north west)+(\xpos pt,-\ypos pt)$);}
        \dospy
        \pgfmathtruncatemacro\pgfmathresult{ifthenelse(\columncount==\pgfkeysvalueof{/tikz/zoomboxarray columns},\rowcount+1,\rowcount)}
        \global\let\rowcount=\pgfmathresult
        \pgfmathtruncatemacro\pgfmathresult{ifthenelse(\columncount==\pgfkeysvalueof{/tikz/zoomboxarray columns},1,\columncount+1)}
        \global\let\columncount=\pgfmathresult
        \ifblackandwhitecycle
            \pgfmathtruncatemacro{\newpatternnumber}{\patternnumber+1}
            \global\edef\patternnumber{\newpatternnumber}
        \fi
    \end{scope}
}

\newcommand{\zoombarycentric}[1]{\begin{tikzpicture}[zoomboxarray, zoomboxarray columns=1, zoomboxarray rows=1, zoomboxarray inner gap=0.5cm, connect zoomboxes, zoombox paths/.append style={ultra thick}]
    \node [image node] { \includegraphics[width=0.45\textwidth]{#1} };
    \zoombox[magnification=5]{0.600,0.48}
\end{tikzpicture}}

\DeclareFontFamily{OT1}{pzc}{}
\DeclareFontShape{OT1}{pzc}{m}{it}{<-> s * [1.10] pzcmi7t}{}
\DeclareMathAlphabet{\mathpzc}{OT1}{pzc}{m}{it}

\newtheorem{definition}{Definition}
\newtheorem{theorem}{Theorem}

\DeclareFontFamily{OT1}{pzc}{}
\DeclareFontShape{OT1}{pzc}{m}{it}{<-> s * [1.10] pzcmi7t}{}
\DeclareMathAlphabet{\mathpzc}{OT1}{pzc}{m}{it}

\newcommand{\ppintensity}{\lambda}
\newcommand{\ppcumintensity}{\Lambda}
\newcommand{\timeinverse}{\tau}
\newcommand{\arrivalprocess}{\mathscr{N}}
\newcommand{\poissonprocess}{\mathscr{P}}

\DeclareMathOperator*{\E}{\mathbb{E}}
\newcommand{\nrgbdistricts}{379}
\newcommand{\forceofinf}{\phi}
\newcommand{\forceofinfbetween}{\mathring{\phi}}
\newcommand{\infectedpotential}{\mathcal{I}}
\newcommand{\susceptiblecontribution}{\mu}
\newcommand{\spectralradius}{\Upsilon}
\newcommand{\patches}{\mathcal{P}}
\newcommand{\realnumbers}{\mathbb{R}}
\newcommand{\mdpstatespace}{\mathcal{S}}
\newcommand{\mdpactionspace}{\mathcal{A}}
\newcommand{\mdptransition}{T}
\newcommand{\mdprewardfn}{R}
\newcommand{\epiweeks}{\mathpzc{w}}
\newcommand{\simplexset}{\mathbb{S}}
\newcommand{\aitchisonmean}{C_A}
\newcommand{\mdpstate}{\mathbf{s}}
\newcommand{\mdpaction}{\mathbf{a}}
\newcommand{\schoolclosurebudget}{\mathpzc{b}}
\newcommand{\mobilitymatrix}{\mathcal{M}}
\newcommand{\probofinf}{\beta}
\newcommand{\recoveryrate}{\gamma}
\newcommand{\latencyrate}{\zeta}

\begin{document}

\title{Deep reinforcement learning for large-scale epidemic control}  

\author[1,2]{Pieter Libin}
\author[1]{Arno Moonens}
\author[1]{Timothy Verstraeten}
\author[1]{Fabian Perez-Sanjines}
\author[3]{Niel Hens}
\author[2]{Philippe Lemey}
\author[1]{Ann Now\'{e}}
\affil[1]{Artificial Intelligence Lab, Department of computer science, Vrije Universiteit Brussel, Brussels, Belgium}
\affil[2]{KU Leuven – University of Leuven, Department of Microbiology and Immunology, Rega Institute for Medical Research, Clinical and Epidemiological Virology, Leuven, Belgium}
\affil[3]{Interuniversity Institute of Biostatistics and statistical Bioinformatics, Data Science Institute, Hasselt University, Hasselt, Belgium}

\maketitle

 \begin{abstract}
Epidemics of infectious diseases are an important threat to public health and global economies. 
Yet, the development of prevention strategies remains a challenging process, as epidemics are non-linear and complex processes.
For this reason, we investigate a deep reinforcement learning approach to automatically learn prevention strategies in the context of pandemic influenza.
Firstly, we construct a new epidemiological meta-population model, with $\nrgbdistricts$ patches (one for each administrative district in Great Britain), that adequately captures the infection process of pandemic influenza.
Our model balances complexity and computational efficiency such that the use of reinforcement learning techniques becomes attainable.
Secondly, we set up a ground truth such that we can evaluate the performance of the ``Proximal Policy Optimization" algorithm to learn in a single district of this epidemiological model.
Finally, we consider a large-scale problem, by conducting an experiment where we aim to learn a joint policy to control the districts in a community of 11 tightly coupled districts, for which no ground truth can be established. 
This experiment shows that deep reinforcement learning can be used to learn mitigation policies in complex epidemiological models with a large state space.
Moreover, through this experiment, we demonstrate that there can be an advantage to consider collaboration between districts when designing prevention strategies.
 \end{abstract}
 \section{Introduction}
Epidemics of infectious diseases are an important threat to public health and global economies. The most efficient way to combat epidemics is through prevention. To develop prevention strategies and to implement them as efficiently as possible, a good understanding of the complex dynamics that underlie these epidemics is essential. To properly understand these dynamics, and to study emergency scenarios, epidemiological models are necessary. Such models enable us to make predictions and to study the effect of prevention strategies in simulation. The development of prevention strategies, which need to fulfil distinct criteria (i.a., prevalence, mortality, morbidity, cost), remains a challenging process. For this reason, we investigate a deep reinforcement learning (RL) approach to automatically learn prevention strategies in an epidemiological model.
The use of model-free deep reinforcement learning is particularly interesting, as it allows us to set up a learning environment in a complex epidemiological setting (i.e., large state space and non-linear dependencies) while imposing few assumptions on the policies to be learned.
In this work, we conduct our experiments in the context of pandemic influenza, where we aim to learn optimal school closure policies to mitigate the epidemic.

Pandemic preparedness is important, as influenza pandemics have made many victims in the (recent) past \cite{Paules2017} and the ongoing COVID-19 epidemic is yet another reminder of this fact \cite{zhu2020novel}.
Contrary to seasonal influenza epidemics, an influenza pandemic is caused by a newly emerging virus strain that can become pandemic by spreading rapidly among naive human hosts (i.e., human hosts with no prior immunity) worldwide \cite{Paules2017}.
This means that at the start of the pandemic no vaccine will be available and it will take several months before vaccine production can commence \cite{webby2003we}.
For this reason, learning optimal strategies of non-therapeutic intervention measures, such as school closure policies, is of great importance to mitigate pandemics \cite{markel2007nonpharmaceutical}.

To meet this objective, we consider a reinforcement learning approach. However, as the state-of-the-art of reinforcement learning techniques require many interactions with the environment in order to converge, our first contribution entails a realistic epidemiological model that still has a favourable computational performance.

Specifically, we construct a meta-population model that consists of a set of $\nrgbdistricts$ interconnected patches, where each patch corresponds to an administrative region in Great Britain and is internally represented by an age-structured stochastic compartmental model.
To conduct our experiments, we establish a Markov decision process with a state space that directly corresponds to our epidemiological model, an action space that allows us to open and close schools on a weekly basis, a transition function that follows the epidemiological model's dynamics, and a reward function that is targeted to the objective of reducing the attack rate (i.e., the proportion of the population that was infected).
In this work, we will use ``Proximal Policy Optimization" (PPO) \cite{schulman2017proximal} to learn the school closure policies.

First, we set up an experiment in an epidemiological model that covers a single administrative district.
This setting enables us to specify a ground truth that allows us to empirically assess the performance of the policies learned by PPO.
In this analysis, we consider different values for the basic reproductive number $R_0$ (Definition~\ref{def:r0}) and the population composition (i.e., proportion of adults, children, elderly, adolescents) of the district. 
Both parameters induce a significant change of the epidemic model's dynamics.

\begin{definition}[Basic reproductive number]
The basic reproductive number, $R_0$, is the number of infections that is, on average, generated by one single infected individual that is placed in an otherwise fully susceptible population.
\label{def:r0}
\end{definition}

Through these experiments, we demonstrate the potential of deep reinforcement learning algorithms to learn policies in the context of complex epidemiological models, opening the prospect to learn in even more complex stochastic models with large action spaces.
In this regard, we consider a large scale setting where we examine whether there is an advantage to consider the collaboration between districts when designing school closure policies.
We conduct an experiment in our epidemiological model with \nrgbdistricts{} districts and attempt to learn a joint policy to control the districts in the Cornwall-Devon community, a set of $11$ tightly coupled districts.
To this end, we assign an agent to each of the 11 districts of the Cornwall-Devon community and use a reinforcement learning approach to learn a joint policy.
We compare this joint policy to a non-collaborative policy (i.e., aggregated independent learners).

\section{Related work}
The closing of schools is an effective way to limit the spread of an influenza pandemic \cite{markel2007nonpharmaceutical}.
For this reason, the use of school closures as a mitigation strategy has been explored in variety of modelling studies \cite{halder2010developing,brown2011would,milne2013cost,de2018impact,ciavarella2016school,germann2019school,halloran2008modeling,haber2007effectiveness,eames2012measured}, of which the work presented in \cite{germann2019school} is the most recent and comprehensive study.

The concept to learn dynamic policies by formulating the decision problem as a Markov decision process (MDP) was first introduced in \cite{yaesoubi2011dynamic}.
The proposed technique was used to investigate dynamic tuberculosis case-finding policies in HIV/tuberculosis co-epidemics \cite{yaesoubi2013identifying}.
Later, the technique was extended towards a method to include cost-effectiveness in the analysis \cite{yaesoubi2016identifying}, and applied to investigate mitigation policies (i.e., school closures and vaccines) in the context of pandemic influenza in a simplified epidemiological model.
The work presented in \cite{yaesoubi2011dynamic,yaesoubi2016identifying} uses a policy iteration algorithm to solve the MDP.
To scale this approach to larger problem settings, we explore the use of on-line reinforcement learning techniques (e.g., TD-learning, policy gradient).
Note that the "Deep Q-networks" algorithm was recently used to investigate culling and vaccination in farms in a simple individual-based model to delay the spread of viruses in a cattle population \cite{probert2019context}.
However, to our best knowledge, the work presented in this manuscript is the first attempt to use deep reinforcement learning algorithms directly on a complex meta-population model.
Furthermore, we experimentally validate the performance of these algorithms using a ground truth, in a variety of model settings  (i.e., different census compositions and different $R_0$'s).
This is the first validation of this kind and it demonstrates the potential of on-line deep reinforcement learning techniques in the context of epidemic decision making.
Finally, we present a novel approach to investigate how intervention policies can be improved by enabling collaboration between different geographic districts, by formulating the setting as  a multi-agent problem, and by solving it using deep multi-agent reinforcement learning algorithms.
%TODO: add later
%Next to stateful reinforcement learning, the use of multi-armed bandits has recently been explored to assist decision makers to efficiently select the optimal prevention strategy \cite{libin2018bayesian,libin2019bayesian}.

\section{Epidemiological model}
We construct a meta-population model that consists of $\nrgbdistricts$ patches, where each patch represents one administrative region in Great Britain.
Great Britain consists of three countries with the following administrative regions: 325 districts in England, 22 unitary authorities in Wales and 32 council areas in Scotland.
Each patch consists of a stochastic age-structured compartmental model, which we describe in sub-section~\ref{sec:intra-patch-seir}, and the different patches are connected via a mobility model, as detailed in sub-section~\ref{sec:between_patch_model}.
In sub-section~\ref{sec:spatial_model_validation} we discuss how we validate and calibrate the model.
We analyse the model's computational complexity and discuss the model's performance in the Supplementary Information.

\subsection{Intra-patch model}
\label{sec:intra-patch-seir}
\begin{figure}
\center
    \includegraphics[width=.6\textwidth]{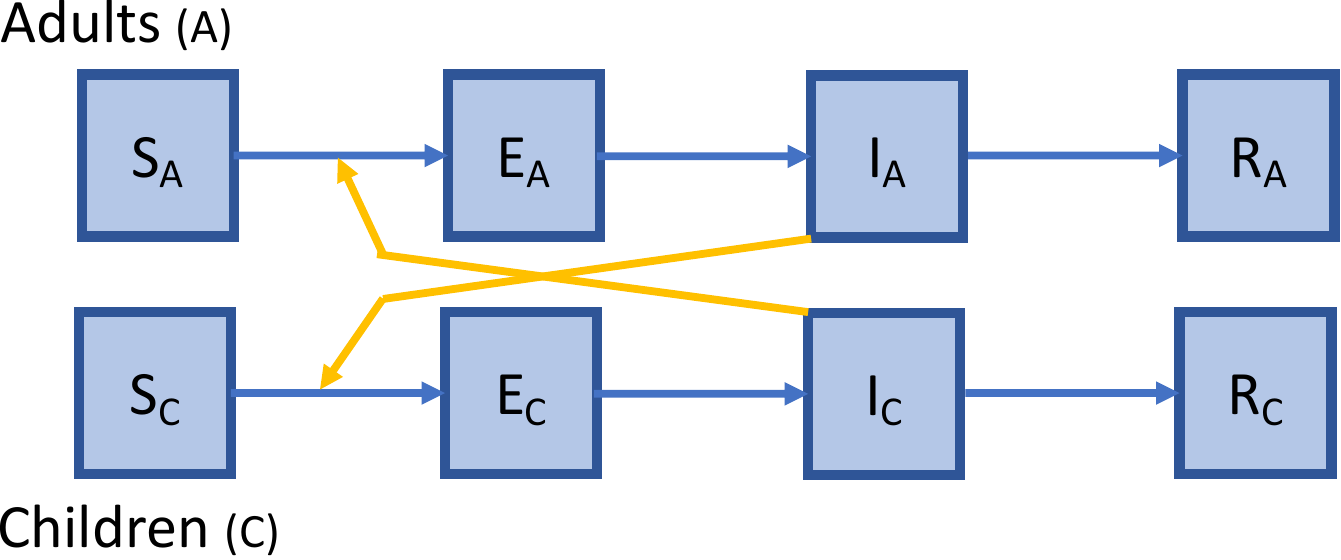}
  \caption{We depict an age-structured SEIR model that considers two age groups (i.e., adults and children). This model consists of two SEIR models, one for each age group, that are connected to represent mixing between the age groups (yellow arrows). Note that it is also possible to mix within the age groups. Note that we use two age groups in this figure to provide a clear visualization of the model. In our actual model, we consider four different age groups.}
  \label{fig:age_seir}
\end{figure}

We consider a stochastic compartmental SEIR model from which we sample trajectories.
We first describe the model in terms of ordinary differential equations (i.e., a deterministic representation) that we then transform to stochastic differential equations \cite{allen2008construction} to make a stochastic evaluation possible.
An SEIR model divides the population in a susceptible, exposed, infected and recovered compartment, and is commonly used to model influenza epidemics \cite{eames2012measured}.
More specifically, we consider an age-structured SEIR model (see Figure~\ref{fig:age_seir} for a visualization) with a set of $n$ disjoint age groups \cite{eames2012measured,fumanelli2012inferring}.
This model is formally described by this system of ordinary differential equations (ODEs), defined for each age group $i$:
\begin{equation}
\begin{split}
\diff{S_i}{t} &= -\forceofinf_i(t) S_i(t) \\
\diff{E_i}{t} &= \forceofinf_i(t) S_i(t) - \latencyrate E_i(t) \\ 
\diff{I_i}{t} &= \latencyrate E_i(t) - \recoveryrate I_i(t) \\
\diff{R_i}{t} &= \recoveryrate I_i(t).
\end{split}
\end{equation}

Every susceptible individual in age group $i$ is subject to an age-specific and time-dependent force of infection:
\begin{equation}
\forceofinf_i(t) = \sum_{j=1}^{n}\probofinf M_{ij}(t)\frac{I_j(t)}{N_j(t)},
\end{equation}
which depends on:
\begin{itemize}
\item The probability of transmission $\probofinf$ when a contact occurs. 
\item The time-dependent contact matrix $M$, where $M_{ij}(t)$ is the average frequency of contacts that an individual in age group $i$ has with an individual in age group $j$.
\item The frequency at which contacts with infected individuals (in age group $j$) occur: $I_j(t)/N_j(t)$.
\end{itemize}
Once exposed, individuals move to the infected state according to the latency rate $\latencyrate$. 
Individuals recover from infection (i.e., get better or die) at a recovery rate $\recoveryrate$.

We omit vital dynamics (i.e., births and deaths that are not caused by the epidemic) in this SEIR model, as the epidemic's time scale is short and we therefore assume that births and deaths will have a limited influence on the epidemic process \cite{towers2012social}. 
Therefore, at any time: 
\begin{equation}
N_i(t) = S_i(t) + E_i(t) + I_i(t) + R_i(t),
\end{equation}
where the total population size $N_i$ corresponds to age-specific census data.
Our model considers 4 age groups: children (0-4 years), adolescents (5-18 years), adults (19-64 years) and elderly (65 years and older).

Note that the contact frequency $M_{ij}(t)$ is time-dependent, in order to model school closures, i.e., a different contact matrix is used for school term and school holiday. Following \cite{eames2012measured}, we consider \textit{conversational contacts}, i.e., contacts for which physical touch is not required.
As we aim to model the effectiveness of school closure interventions, we use the United Kingdom contact matrices presented in \cite{eames2012measured}, which encodes a contact matrix for both school term and school holiday.
These contact matrices are the result of an internet-based social contact survey completed by a cohort of participants \cite{eames2012measured}.
The contact matrices encode for the same age groups as mentioned before: children, adolescents, adults and elderly.

We defined the SEIR model in terms of a system of ODEs which implies a deterministic evaluation of the system.
However, for predictions, stochastic models are preferred, as they to account for stochastic variation and allow us to quantify uncertainty \cite{king2015avoidable}.
In order to sample trajectories from this set of differential equations,  we transform the system of ODEs to a system of stochastic differential equations (SDEs), using the transformation procedure presented in \cite{allen2008construction}.
This procedure introduces stochastic noise to the system by adding a Wiener process to each transition in the ODE.
We evaluate the SDE at discrete time steps using the Euler-Maruyama approximation method \cite{allen2008construction}.

Each compartmental model is representative of one of the 
administrative districts and as such the compartmental model is parametrised with the census data of the respective district, i.e., population counts stratified by age groups.
We use the 2011 United Kingdom census data made available by NOMIS \footnote{\url{https://www.nomisweb.co.uk}}.  
We present more details on the census data in the Supplementary Information.

\subsection{Inter-patch model}
\label{sec:between_patch_model}
Our model, that is comprised of a set of connected SEIR patches, is inspired by the recent BBC pandemic model \cite{klepac2018contagion}.
The BBC pandemic model was in its turn motivated by the model presented in \cite{gog2014spatial}.

At each time step, our model checks whether a patch $p$ becomes infected.
This is modulated by the patch's force of infection, which combines the potential of the infected patches in the system, weighted by a mobility model, that represents the commuting of adults between the different patches:
\begin{equation}
\label{eq:between_force_of_infection}
\forceofinfbetween_{p}(t) = \sum_{p' \in \patches} \mobilitymatrix_{p'p} \cdot \probofinf \cdot  \left(S^{\text{A}}_{p}(t)\right)^{\susceptiblecontribution} \cdot \infectedpotential_{p'}(t),
\end{equation}
where $\patches$ is the set of patches in the model, $\mobilitymatrix_{p'p}$ is the mobility flux between patch $p'$ and $p$, $\probofinf$ is the probability of transmission on a contact, $S^{\text{A}}_{p}(t)$ is the susceptible population of adults in patch $p$ at time $t$ and its contribution is modulated by parameter $\susceptiblecontribution$ (range in $[0,1]$), and $\infectedpotential_{p'}(t)$ is the infectious potential of patch $p'$ at time $t$.
We define this infectious potential as,
\begin{equation}
\infectedpotential_{p'}(t) = I^{\text{A}}_{p'}(t) \cdot M_{\text{A}\text{A}},
\end{equation} 
where $I^{\text{A}}_{p'}(t)$ is the size at time $t$ of the infectious adult population in patch $p'$ and $M_{\text{A}\text{A}}$ is the average number of contacts between adults, as specified in the contact matrix (see sub-section~\ref{sec:intra-patch-seir}).

$\mobilitymatrix$ is a matrix based on the mobility dataset provided by NOMIS\footnote{We use the NOMIS WU03UK dataset that was released in 2011.}. This dataset describes the amount of commuting between the districts in Great Britain.

In general, this inter-patch model is constructed from first principles i.e., census data, a mobility model, the number of infected individuals and the transmission potential of the virus.
However, for the parameter $\susceptiblecontribution$ that modulates the contribution of the susceptibles in the receptive patch (while it is commonly used in literature \cite{gog2014spatial,eggo2010spatial,kissler2019geographic}) no such intuition is readily available.
Therefore, this parameter is typically fitted to match the properties of the epidemic that is under investigation \cite{gog2014spatial,eggo2010spatial,kissler2019geographic}.
We will calibrate this parameter such that it can be used for a range of $R_0$ values, as detailed in the next sub-section.

Given this time-dependent force of infection, we model the event that a patch becomes infected with a non-homogeneous Poisson process \cite{wang2018characterizing}.
As the process' intensity depends on how the model (i.e., the set of all patches) evolves, we cannot sample the time at which a patch becomes infected a priori.
Therefore, we determine this time of infection using the time scale transformation algorithm \cite{cinlar2013introduction}. Details about this procedure can be found in Supplementary Information. 
Following \cite{klepac2018contagion}, we assume that a patch will become infected only once.

By using this time scale transformation algorithm and evaluating the stochastic differential equation at discrete time steps, we produced a model with favourable performance, i.e., in our experiments we can run about $2$ simulation runs per second on a MacBook Pro (CPU: 2,3 GHz Intel Core i5).
We analyse the model's computational complexity in the Supplementary Information.

\subsection{Calibration and validation}
\label{sec:spatial_model_validation}
Our objective is to construct a model that is representative for contemporary Great Britain with respect to population census and mobility trends.
This model will be used to study school closure intervention strategies for future influenza pandemics. 
While in many studies \cite{kissler2019geographic,gog2014spatial,eggo2010spatial} a model is created specifically to fit one epidemic case, we aim for a model that is robust with respect to different epidemic parameters, most importantly $R_0$, the basic reproduction number.

To validate our model according to these goals, we conduct two experiments. 
In the first experiment, we compare our patch model to an SEIR compartmental model that uses the same contact matrix and age structure, but with homogeneous spatial mixing (i.e., no spatial structure).
While we do not expect our model to behave exactly like the compartmental model, as the patches and the mobility network that connects them induces a different dynamic, we do observe similar trends with respect to the epidemic curve and peak day.
This experiment also enables us to calibrate the $\susceptiblecontribution$ parameter.
We present a detailed description of this analysis and report the results in Supplementary Information.
In the second experiment we show that our model is able to reproduce the trends that were observed during the 2009 influenza pandemic, commonly known as the swine-origin influenza pandemic, i.e., A(H1N1)v2009, that originated in Mexico.
The 2009 influenza pandemic in Great Britain is an interesting case to validate our model for two reasons.
Firstly, the pandemic occurred quite recently and thus our model's census and mobility scheme are a good fit, as both the datasets on which we base our census and mobility model were released in 2011.
Secondly, due to the time when the virus entered Great Britain, the summer holiday started 11 weeks after the emergence of the epidemic. 
The timing of the holidays had a severe impact on the progress of the epidemic and resulted in a epidemic curve with two peaks. 
This characteristic epidemic curve enables us to demonstrate the predictive power of our age-structured contact model with support for school closures.
%Thirdly, the number of symptomatic cases that occurred in Great Britain during the 2009 pandemic was recorded meticulously and is publicly available \cite{kubiak20122009}, such that we can compare our model's output to this dataset.
In Figure~\ref{fig:gb_2009_bestfit}, we show a set of model realisations in conjunction with the symptomatic case data, which shows that we were able to closely match the epidemic trends observed during the British pandemic in 2009 (details on this case study in the Supplementary Information).
%This model was parametrized with a basic reproductive number of $1.4$ and infectious period of $2.6$, to fit the observed data.
%The reproductive number is in good concordance with the general consensus that the virus responsible for the 2009 pandemic exhibited a moderate infectiousness.
%While the infectious period slightly differs from the value reported by \cite{balcan2009seasonal} (i.e., $2.5$ days), it lies well within the confidence bounds reported in this study (confidence interval: 1.1-4.0 days).
Note that our model reports the number of infections while the British Health Protection Agency only recorded symptomatic cases. 
Therefore we scale the epidemic curve with a factor of $\frac{1}{4}$. 
This large number of asymptomatic cases produced by our model is in line with earlier serological surveys \cite{miller2010incidence} and with previous modelling studies \cite{kubiak20122009}.

%cd /Users/plibin/research/spatial-pandemic-experiments/spatial/uk_2009
%./run_all.sh
%
%(or on the Thinking VSC)
%wsub -A lp_hiv_networks -batch run_all.pbs -data run_all.csv
%
%python plot_trajectories.py --R0s "1.4" --dir ./runs/ --n 10  --gamma 2.6 --proportion_clinical .25  --cases_csv /Users/plibin/projects/spatial-pandemic-py/data/2009-pandemic-cases/cases-2009-extracted-from-kubiak-2012.csv -o gb_2009_bestfit.pdf
\begin{figure}
  \centering
   \includegraphics[width=.6\textwidth]{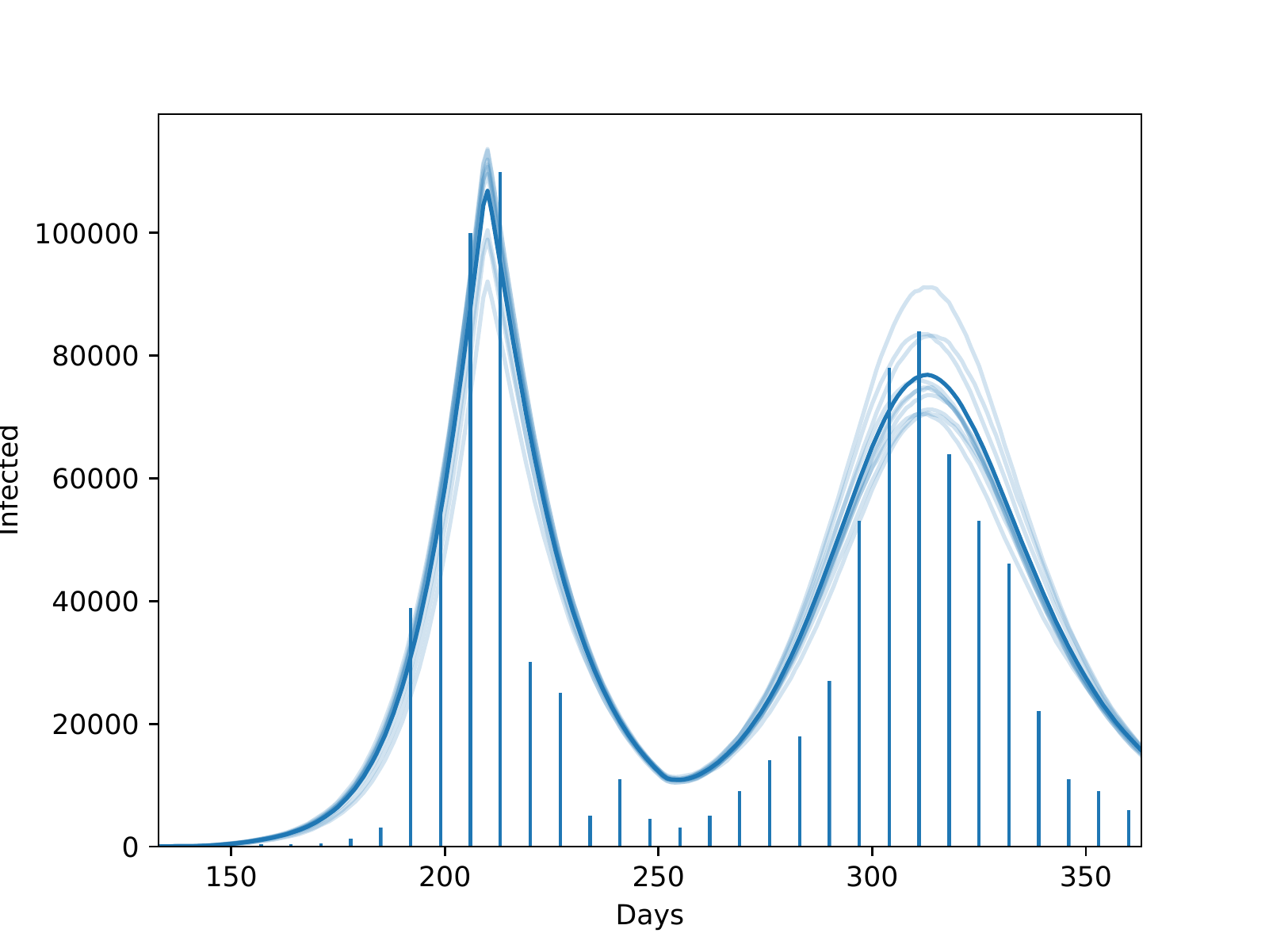}

  \caption{We show that our model (blue epidemic curves) is able to match the trends observed in the British pandemic of 2009 (the vertical bars represent the number of infected individuals that was recorded during the epidemic).  We show 10 stochastic trajectories.}
      \label{fig:gb_2009_bestfit}
\end{figure}

\section{Learning environment}
\label{sec:learning_environment}
In order to apply reinforcement learning, we construct an MDP based on the epidemiological model that we introduced in the previous section.
This epidemiological model consists of patches that correspond to administrative regions.

We have an agent for each patch that we attempt to control, and for each agent we have an action space $\mdpactionspace = \{\text{open}, \text{close}\}$ that allows us to open and close schools for one week.
Each agent has a predefined budget $\schoolclosurebudget$ of school closure actions it can execute.
Once this budget is depleted, executing a close action will default to executing an open action.
We refer to the remaining budget at time $t$ as $\schoolclosurebudget^{(t)}$.
In the epidemiological model, when schools are closed we use a contact matrix that is representative for school holidays and when schools are open we use a contact matrix that is representative for school term (details in Section~\ref{sec:intra-patch-seir}).

For each patch, we consider a state space that combines the state of the SEIR model and the remaining budget of school closures $\schoolclosurebudget_p^{(t)}$.
For the SEIR model, we have 16 state variables (i.e., $\realnumbers^{16}$), as we have an SEIR model (4 state variables) for each of the four age groups.
The remaining school closure budget is encoded as an integer, resulting in a combined state space of 17 variables.
We refer to the state space of one patch $p$, that thus combines the epidemiological states and the budget, as $\mdpstatespace_p$.
The state space $\mdpstatespace$ of the MDP corresponds to the aggregation of the state space of each patch that we attempt to control:
\begin{equation}
\bigtimes_{p \in \patches^\text{c}} \mdpstatespace_p,
\label{eq:model_state}
\end{equation}
where $\patches^c \subseteq \patches$ is the set of patches that we control.

The transition probability function $\mdptransition(\mdpstate' \mid \mdpstate,\mdpaction)$ stochastically determines the state of the epidemic in the next week, taking into account the school closure actions that were chosen, using the epidemiological dynamics as defined in the previous section.

To reduce the attack rate, we consider an immediate reward function that quantifies the negative loss in susceptibles over one simulated week:
\begin{equation}
\mdprewardfn_{\text{AR}}(\mdpstate,\mdpaction,\mdpstate') = -(S(\mdpstate)-S(\mdpstate')),
\end{equation}
where $S(.)$ is the function that determines the total number of susceptible individuals given the state of the epidemiological model.

For PPO, we have both a policy and value network. 
The policy network accepts the state of the epidemiological model as input (details in Section~\ref{sec:learning_environment}) and the output of the network contains 1 unit, which is passed through a sigmoid activation function.
This output thus represents the probability of keeping the schools open in the district.
Every hidden layer in the PPO network uses the hyperbolic tangent activation function.
The value network has the same architecture as the policy network, with the exception that the output is not passed through an activation function.
We will refer to this setting throughout this work as the single-district PPO agent.

PPO's hyper-parameters are tuned (hyper-parameter values in Supplementary Information) on a single-district (i.e., the Greenwich district) learning environment with $R_0=1.8$.
To this end, we performed a hyper-parameter sweep using Latin hypercube sampling ($n=1000$) \cite{stein1987large}.

We conduct two kinds of experiments: in the context of a single district and in the context of the Great Britain model that combines all \nrgbdistricts{} districts.
We consider two values for the reproductive number, i.e., $R_0=\{1.8,2.4\}$, to investigate the effect of distinct reproductive numbers. 
$R_0=1.8$ represents an epidemic with moderate transmission potential \cite{ferguson2006strategies} and $R_0=2.4$ represents an epidemic with high transmission potential \cite{longini2005containing}.
We investigate the effect of different school closure budgets, i.e., $\schoolclosurebudget=\{2,4,6\}$ weeks.
The epidemic is simulated for a fixed number of weeks, chosen beforehand, to ensure that there is enough time for the epidemic to fade out after its peak.
Following \cite{baguelin2010vaccination}, we use a latent period of one day ($\latencyrate=\frac{1}{1}$) and an infectious period of 1.8 days ($\recoveryrate=\frac{1}{1.8}$).
Given the contact matrix $M_{ij}(t)$, the latency rate $\latencyrate$, the recovery rate $\recoveryrate$, we can compute $\probofinf$ for an $R_0$, as specified in the Supplementary Information.

\section{PPO versus ground truth}
\label{sec:composition}
We now establish a ground truth for different population compositions, i.e., the proportion of the different age groups in a population.
We will use this ground truth to empirically validate that PPO converges to the appropriate policy.

To establish this ground truth\footnote{Note that this is a proxy to the ground truth, as we use a deterministic version of the model.}, first consider that when we deal with a single district, we can approach the `average' behaviour of the model by removing the stochastic terms from the differential equations.
Hence, for a particular parameter configuration (i.e., district, $R_0$, $\recoveryrate$, $\latencyrate$), the model will always produce the same epidemic curve.
This means that the state space of this deterministic epidemic model directly corresponds to the time of the epidemic.
For an epidemic that spans $\epiweeks$ weeks, we can formulate a school closure policy as a binary number with $\epiweeks$ digits, where the digit at position $i$ signifies whether schools should be open (1) or closed (0) during the $i$-th week.
For short-lived epidemics, such as influenza epidemics, we can enumerate these policies and evaluate them in our model (i.e., using exhaustive policy search). Note that, in the epidemiological models that we consider, the epidemic spans no more than 25 weeks, and thus exhaustive search is possible.

In this analysis, we consider different values for the basic reproductive number $R_0$ and the population composition of the district, both parameters that induce a significant change of the epidemic model’s dynamics.
To this end, we select 10 districts that are representative of the population heterogeneity in Great Britain: one district that is representative for the average of this census distribution and a set of nine districts that is representative for the diversity in this census distribution. Details on this selection procedure can be found in the Supplementary Information.

To evaluate PPO with respect to the ground truth, we repeat the experiment for which we established a ground truth (i.e., $R_0 \in \{1.8,2.4\}$, 10 districts and $\schoolclosurebudget \in \{2,4,6\}$) and learn a policy using PPO, in the stochastic epidemic model.
For each experimental setting (i.e., the combination of a district, an $R_0$ value, and a school closure budget $\schoolclosurebudget$), we run PPO 5 times (5 trials), to asses the variance of the learning performance. 
Each PPO trial is run for $10^4$ episodes of $43$ weeks.
We show the learning curves, i.e., total reward at the end of the episode, for the district that is representative for the average of the census distribution (i.e., the Barnsley district in England), with $R_0=2.4$ in Figure~\ref{fig:ppo_learning_curves}, for the other settings we report similar learning curves in the Supplementary Information.

%cd ~/research/spatial-pandemic-experiments/rl-chapter/spatial/census-experiment
%./compare_ppo_ground_truth.sh ar
\begin{figure}
\includegraphics[width=.5\linewidth]{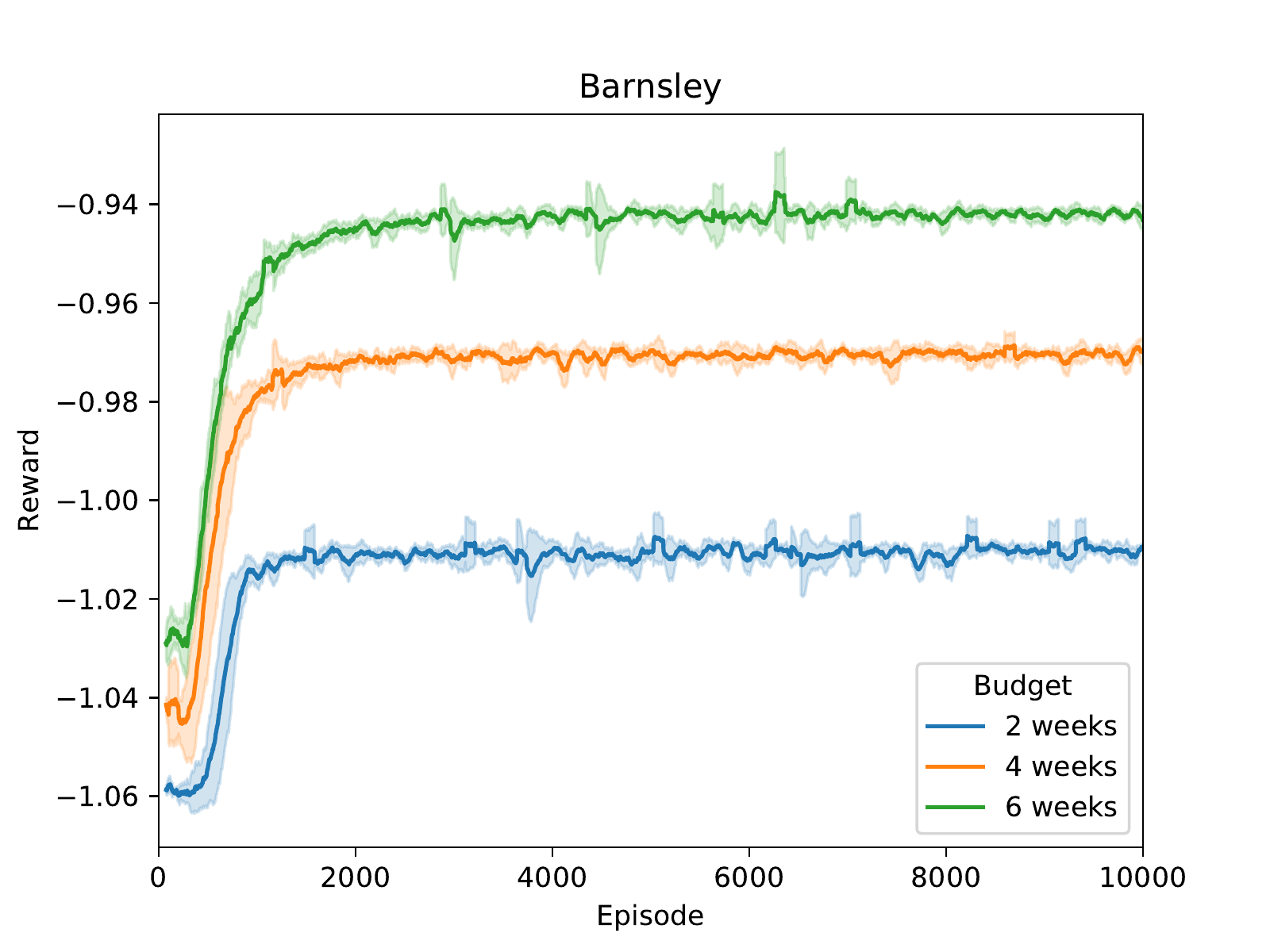}
\includegraphics[width=.5\linewidth]{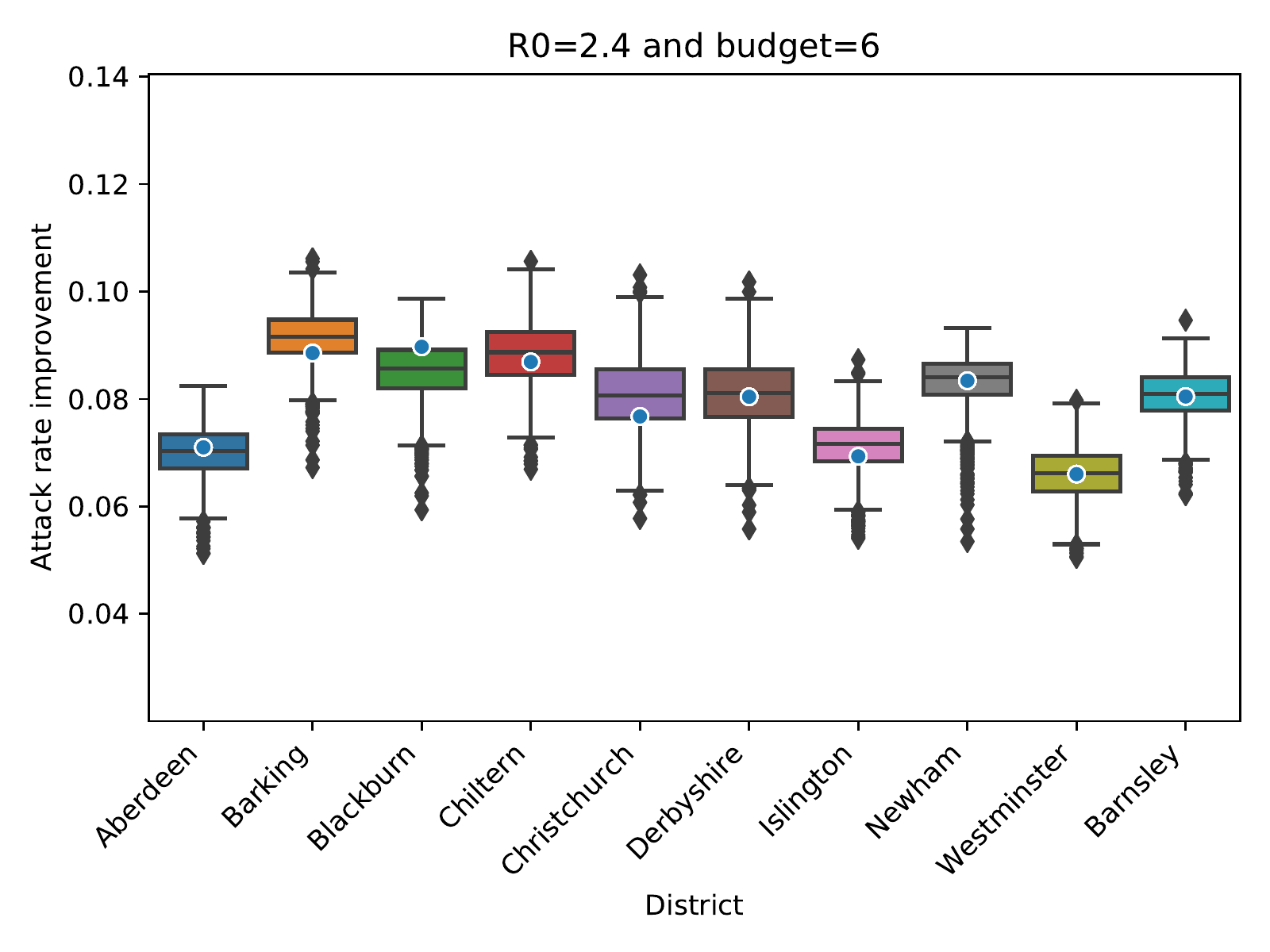}
\caption{[Left panel] PPO learning curves for the Barnsley district with $R_0=2.4$ for the three school closure budgets $\schoolclosurebudget=\{2,4,6\}$. [Right panel] We compare the PPO results to the ground truth for $R_0=2.4$ and $\schoolclosurebudget=6$. Per district, we show a box plot that denotes the outcome distribution that was obtained by simulating the policy learned by PPO 1000 times. On top of this box plot, we show the ground truth, as a blue dot.}
\label{fig:ppo_learning_curves}
\end{figure}

To compare each of the learned policies to its ground truth (one for each district), we take the learned policy and apply it 1000 times in the stochastic model, which results in a distribution over model outcomes (i.e., attack rate improvement: the difference between the attack rate produced by the model and the baseline when no schools are closed).
We then compare this distribution to the attack rate improvement that was recorded for the ground truth.
We show these results, for the setting with a school closure budget of 6 weeks and $R_0=2.4$, in Figure~\ref{fig:ppo_learning_curves}, and for the other settings in Supplementary Information.
These results show that PPO learns a policy that matches the ground truth for all districts and combinations of $R_0$ and $\schoolclosurebudget$.

Note that for these experiments, we use the same hyper-parameters for PPO that were introduced in Section~\ref{sec:learning_environment}.
This demonstrates that, for different values of $R_0$ and for different census compositions (which induce a significant change in dynamics in the epidemic model) these hyper-parameters work well.
This indicates that these hyper-parameters are adequate to be used for different variations of the model.

In this section, we compare to the ground truth (that has been found through an exhaustive policy search) to a policy learned by PPO, a deep reinforcement learning algorithm.
This allows us to empirically validate that PPO converges to the optimal policy.
This experimental validation is important, as it demonstrates the potential of deep reinforcement learning algorithms to learn policies in the context of complex epidemiological models.
This indicates that it is possible to learn in even more complex stochastic models with large action spaces, for which it is impossible to compute the ground truth.
In Section~\ref{sec:marl}, we investigate such a setting, where we aim to learn a joint policy for a set of districts, using deep multi-agent reinforcement learning.

\section{Multi-district reinforcement learning}
\label{sec:marl}
To investigate the collaborative nature of school closure policies, we apply deep multi-agent reinforcement learning algorithms.
In our model, we have \nrgbdistricts{} agents, one for each district, as agents represent the district for which they can control school closure.
As the current state-of-the-art of deep multi-agent reinforcement learning algorithms is limited to deal with about $10$ agents \cite{hernandez2019survey}, we thus need to partition our model into smaller groups of agents, such that deep multi-agent reinforcement learning algorithms become feasible.
To this end, we analyse the mobility matrix $\mobilitymatrix$ to detect  clusters of districts that represent closely connected communities (details in Supplementary Material).
Through this analysis we identify a community with 11 districts, to which we will refer as the Cornwall-Devon community, as it is comprised of the Cornwall and Devon regions.

We now examine whether there is an advantage to consider the collaboration between districts when designing school closure policies.
We conduct an experiment in our epidemiological model with \nrgbdistricts{} districts, and attempt to learn a joint policy to control the districts in the Cornwall-Devon community.
To this end, we assign an agent to each of the 11 districts of the Cornwall-Devon community, and use a reinforcement learning approach to learn a joint policy.
We compare this joint policy to a non-collaborative policy (i.e., aggregated independent learners).

We refer to the state space of one patch $p$ as $\mdpstatespace_p$, as detailed in Section~\ref{sec:learning_environment}.
The state space $\mdpstatespace$ of the MDP corresponds to the aggregation of the state space of the set of patches $\patches^\text{c}$ that we attempt to control.
In this experiment, $\patches^\text{c}$ corresponds to the 11 districts in the Cornwall-Devon community.
 
In order to learn a joint policy, we need to consider an action space that combines the actions for each district $p \in \patches^\text{c}$ that we attempt to control. 
%WRONG:As each agent has an action space $\mdpactionspace = \{\text{open}, \text{close}\}$, we have a joint action space $2^{\mdpactionspace}$, that is the power set of $\mdpactionspace$.
This results in a joint action space with a size that is exponential with respect to the number of agents.
To approach this problem, we use a PPO super-agent that controls multiple districts simultaneously, to learn a joint policy. To this end, we use a custom policy network that gets as input the combined model state of each district $p \in \patches^\text{c}$ (Equation~\ref{eq:model_state}), and as a result, the input layer has $17 \cdot |\patches^\text{c}|$ input units.
In contrast to the single-district PPO, that was introduced in Section~\ref{sec:learning_environment}, the output layer of the policy network of this agent has a unit for each district that we attempt to control.
Again, each output unit is passed through a sigmoid activation function, and hence corresponds to the probability of closing the schools in the associated district.
Similar to the single-district PPO, each hidden layer uses the hyperbolic tangent activation function.
The value network has the same architecture for the input layers and hidden layers, but only has a single output unit that represents the value for the given state.
We will refer to this agent as \emph{multi-district PPO}.
%DOLATER: show a visual representation of the network here?

\begin{figure}
  \centering
   \includegraphics[width=.4\textwidth]{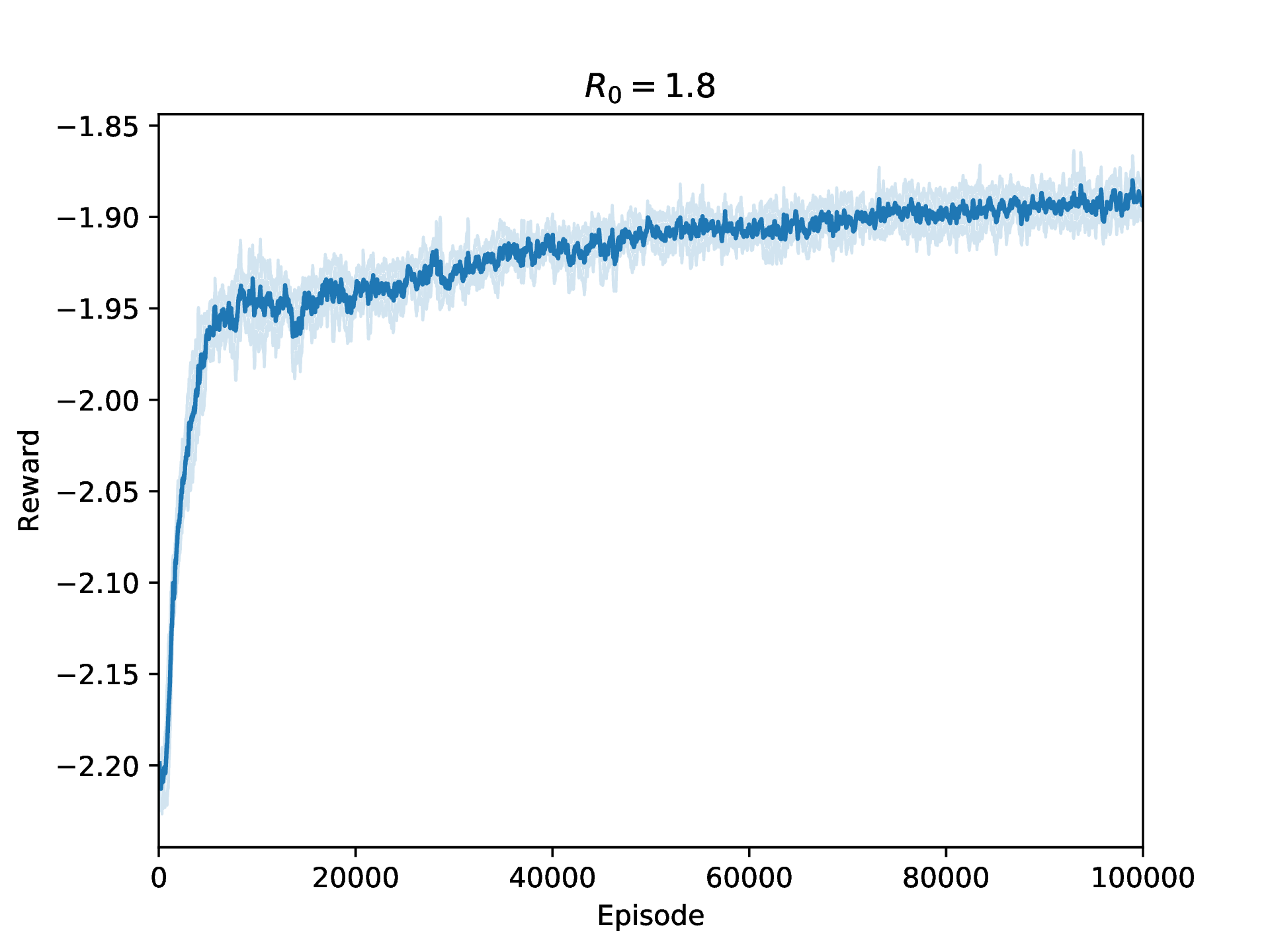}
   \includegraphics[width=.4\textwidth]{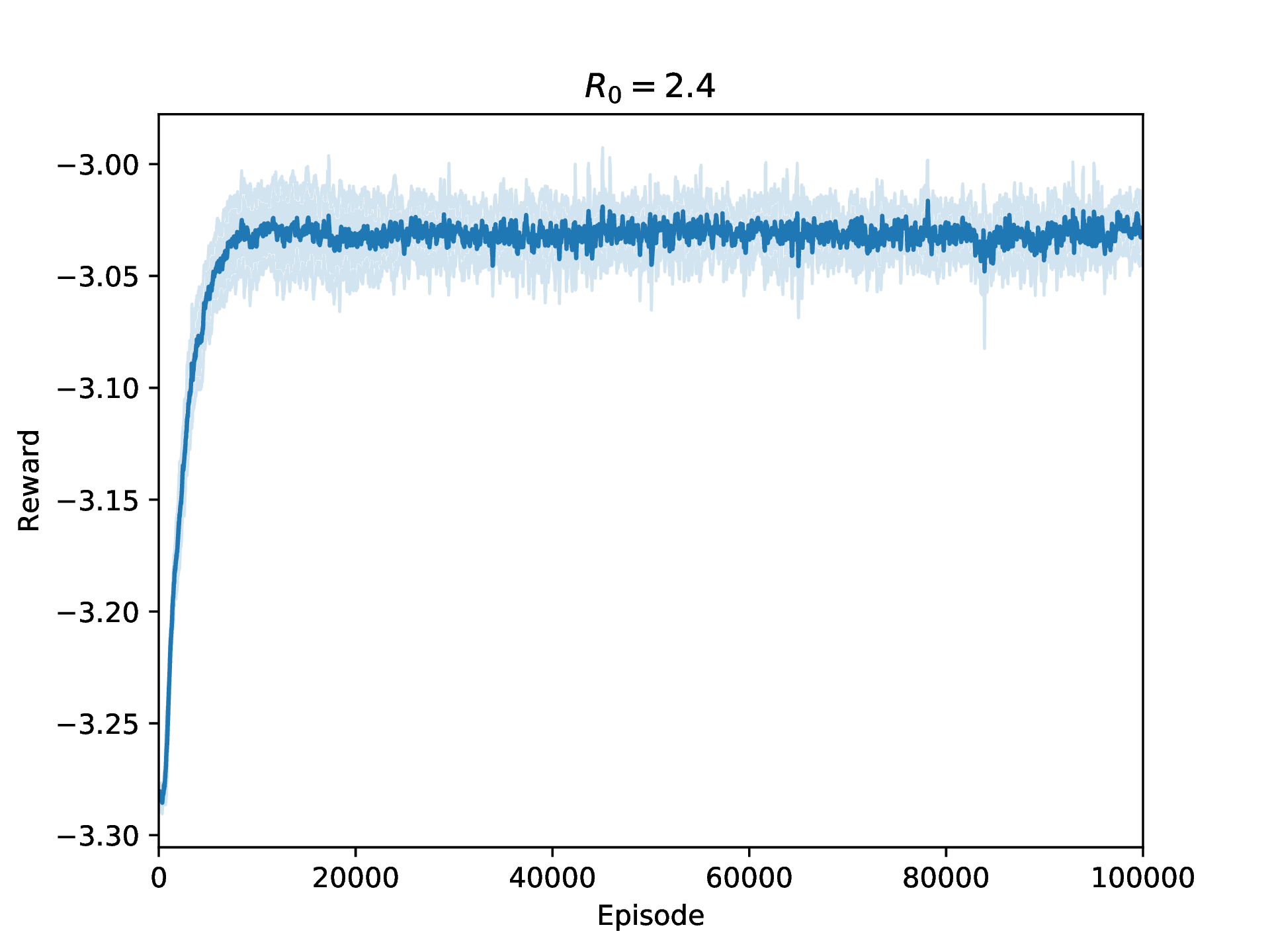}
      \caption{We show the reward curves for multi-district PPO for $R_0=1.8$ (top panel) and $R_0=2.4$ (bottom panel). The reward curves are visualized using a rolling window of 100 steps. The shaded area shows the standard deviation of the reward signal, over 5 multi-district PPO runs.}
      \label{fig:marl_ppo_learning_curves}
\end{figure}

We conduct experiments for $R_0=1.8$ (i.e., moderate transmission potential) and $R_0=2.4$ (i.e., high transmission potential), and we consider a school closure budget of 6 weeks, i.e., $\schoolclosurebudget=6$.
We run multi-district PPO 5 times, to assess the variance of the learning signal, for $10^5$ episodes of $43$ weeks, and we show the learning curves in Figure~\ref{fig:marl_ppo_learning_curves}. 
These learning curves demonstrate a stable and steady learning process. 
For $R_0=1.8$ the reward curve is still increasing, while for $R_0=2.4$ the reward curve indicates that the learning process has converged.

%cd ~/research/spatial-pandemic-experiments/spatial/ma-cornwall
% run the experiments with PBS: multi_ppo.pbs cornwall_ar.csv
% run postprocess.sh
% run plot.sh

\begin{figure}
  \centering
   \includegraphics[width=.4\textwidth]{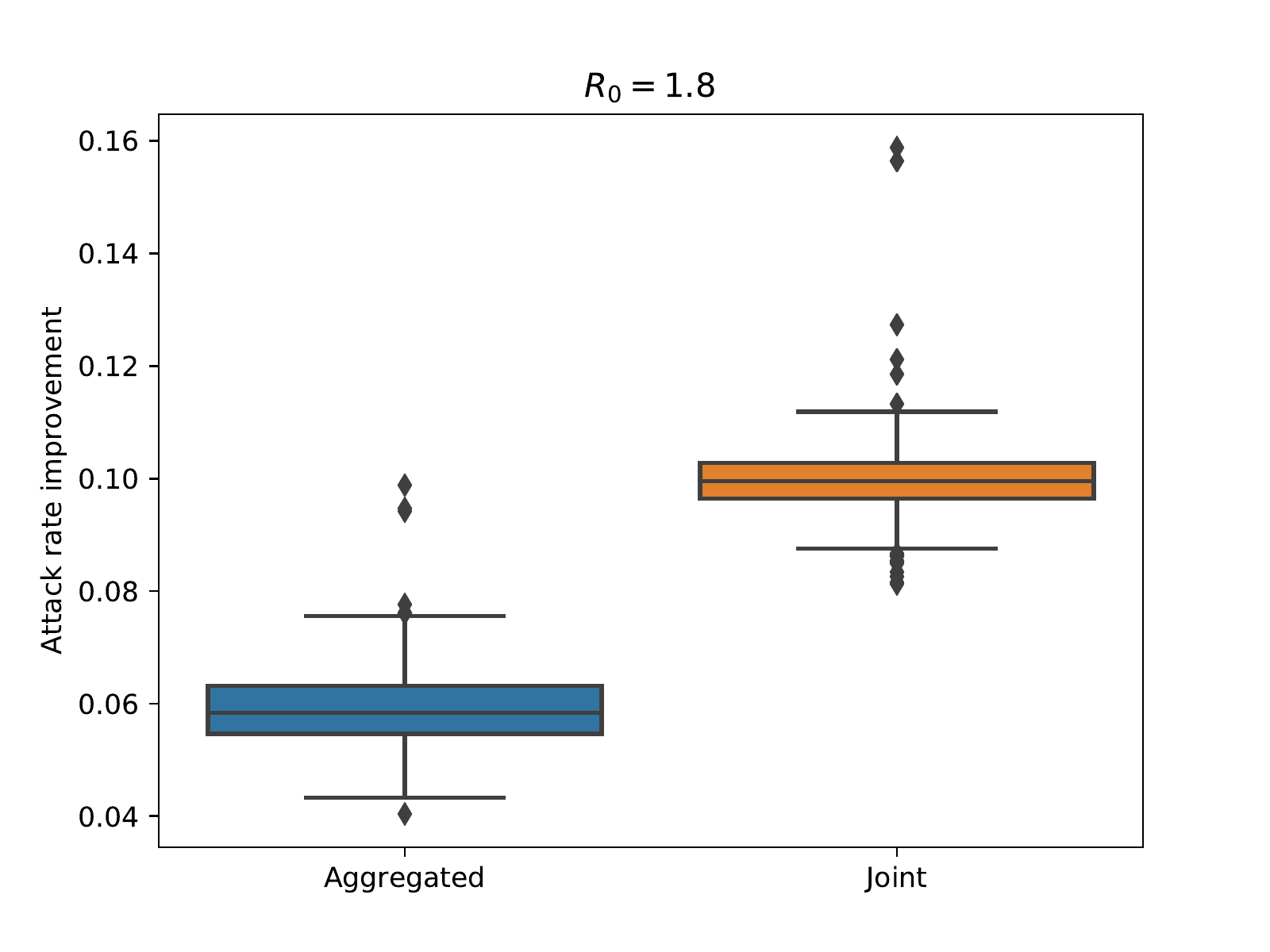}
   \includegraphics[width=.4\textwidth]{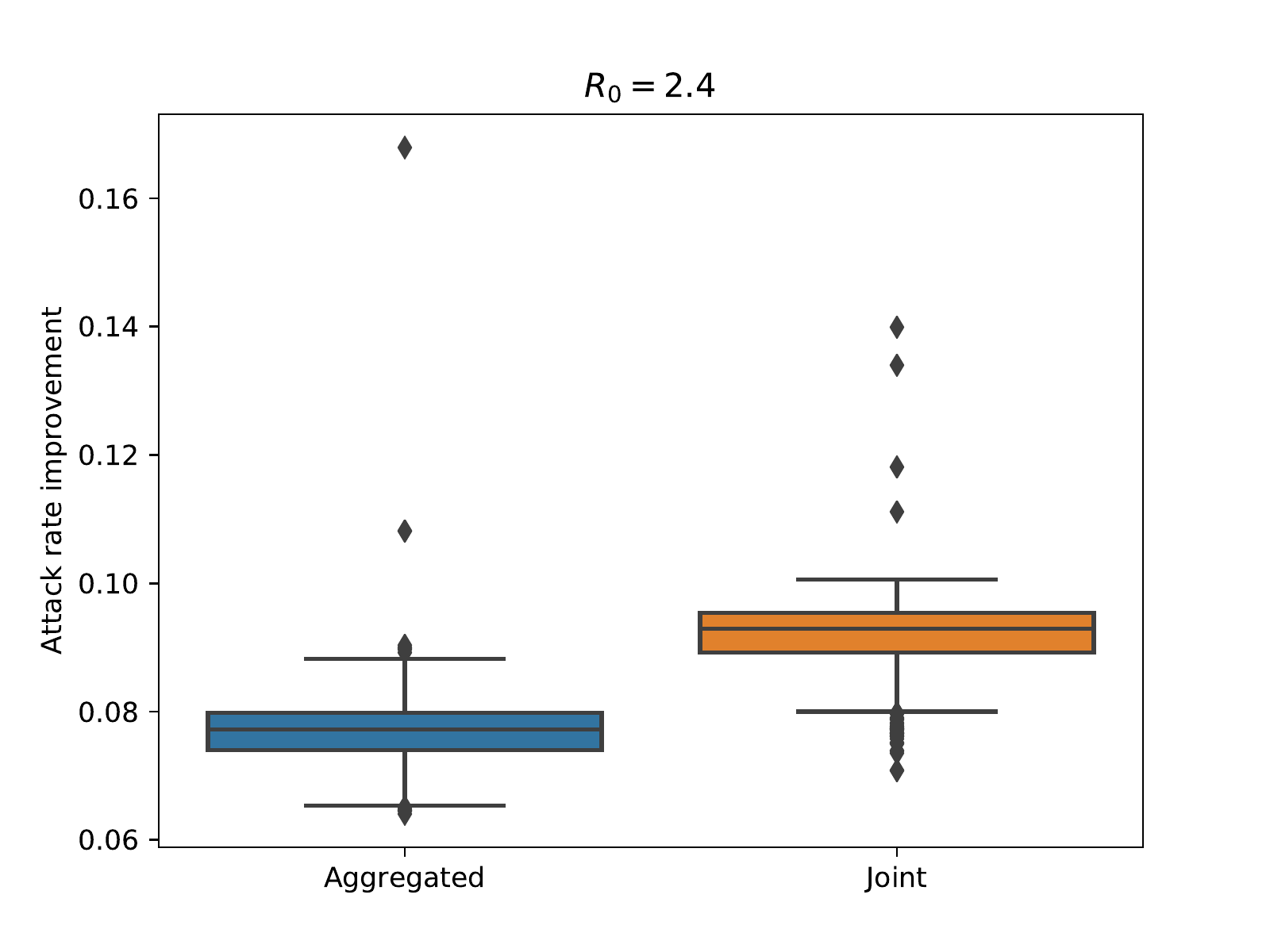}
      \caption{We compare the simulation results of the aggregated policy (blue) and the joint policy (orange) for $R_0 =1.8$ (top panel) and $R_0=2.4$ (bottom panel). For both distributions (i.e., aggregated versus joint) , we show a box plot that denotes the outcome distribution that was obtained by simulating the respective policy 1000 times.}
      \label{fig:joint_vs_aggregate}
\end{figure}

To investigate whether these \emph{joint policies} provide a collaborative advantage, we compare it to the aggregation of single district policies, to which we will refer as the \emph{aggregated policy}.
To construct this aggregated policy, we learn a distinct school closure policy for each of the 11 districts in the Cornwall-Devon community, using PPO, following the same procedure as in Section~\ref{sec:composition}.
To evaluate this aggregated policy, we execute the distinct policies simultaneously.
For the districts that are not controlled (both for the joint and aggregated policy) we keep the schools open for all time steps.
For both $R_0=1.8$ and $R_0=2.4$, respectively, we simulate the joint and the aggregated policy 1000 times, and we show the attack rate improvement distribution in Figure~\ref{fig:joint_vs_aggregate}.
These results corroborate that there is a collaborative advantage when devising school closures policies, for both $R_0=1.8$ and $R_0=2.4$.
However, the improvement is most significant for $R_0=1.8$.
We conjecture that this difference is due to the fact that there is less flexibility when the transmission potential of the epidemic is higher, since there is less time to act.
Although, we observe an improvement when a joint policy is learned, it remains challenging to interpret deep multi-agent policies, and we discuss in Section~\ref{sec:discussion} possible directions for future work with respect to multi-agent reinforcement learning.

%cd ~/research/spatial-pandemic-experiments/spatial/ma-cornwall
% python single_joint_boxplots.py --single_path /Volumes/Seagate/rl/ma-cornwall/single_learned/ --joint_path /Volumes/Seagate/rl/ma-cornwall/joint_learned/ --output .

In this analysis, where we have a limited number of actions per agent, the use of multi-district PPO proved to be successful.
However, the use of more advanced multi-agent reinforcement learning methods is warranted when a more complex action space is considered or a larger number of districts needs to be controlled.
For this reason, we also investigated the recently introduced QMIX \cite{pmlr-v80-rashid18a} algorithm, but the resulting learning curve proved to be quite unstable (shown in Supplementary Information).

We conducted our experiments in the setting of school closures, and our findings are of direct relevance with respect to the mitigation of pandemic influenza.
Furthermore, our novel approach to investigate the collaborative nature of prevention strategies as a multi-agent reinforcement learning problem, can be applied to other epidemiological settings, such as for example the ongoing COVID-19 pandemic.

\section{Discussion}
\label{sec:discussion}
We demonstrate the potential of deep reinforcement learning in the context of epidemiological decision making by conducting experiments that show that PPO converges to the optimal policy.
Next, we investigate and show that there is a collaborative advantage when devising school closures policies, by formulating this hypothesis as a multi-agent problem.

The work conducted in this manuscript indicates that there is the potential to use reinforcement learning in the context of complex stochastic epidemiological models.
For future work, it would be interesting to investigate how well these algorithms scale to even larger state and/or action spaces.
In this regard, the use of attention-based multi-agent reinforcement learning algorithms could be explored \cite{liu2019multiagent}.

Another important concern is to scale these reinforcement learning methods to individual-based epidemiological models, as such models can be easily configured to approach a variety of research scenarios, i.a., vaccine allocation, telecommuting, antiviral drug allocation.
However, the computational burden that is associated with individual-based models complicates the use of reinforcement learning methods \cite{yu2018towards}.
To this end, it would be interesting to devise methods to automatically learn a surrogate model from the individual-based model, such that the reinforcement learning agent can learn in this computationally leaner surrogate model.

While we show that deep reinforcement learning algorithms can be used to learn optimal mitigation strategies, the interpretation of such policies remains challenging \cite{gunning2019darpa}.
This is especially the case for the multi-district setting we considered, where state and time do not match, and the infection onset of the patches is highly stochastic.
To this end, further research into explainable reinforcement learning, both in a single-agent and multi-agent setting, is warranted. 

\section*{Acknowledgments}
Pieter Libin and Timothy Verstraeten were supported by a PhD grant of the FWO (Fonds Wetenschappelijk Onderzoek - Vlaanderen).
We thank Jelena Grujic (Vrije Universteit Brussel) for her comments on the network analysis.

%%%%%%%%%% Merge with supplemental materials %%%%%%%%%%
\pagebreak
%\widetext
\begin{center}
\textbf{\large Supplementary Information}
\end{center}
%%%%%%%%%% Merge with supplemental materials %%%%%%%%%%
%%%%%%%%%% Prefix a "S" to all equations, figures, tables and reset the counter %%%%%%%%%%
\setcounter{equation}{0}
\setcounter{figure}{0}
\setcounter{table}{0}
\setcounter{page}{1}
\setcounter{section}{0}
\makeatletter
\renewcommand{\theequation}{S\arabic{equation}}
\renewcommand{\thesection}{S\arabic{section}}
\renewcommand{\thefigure}{S\arabic{figure}}
%%%%%%%%%% Prefix a "S" to all equations, figures, tables and reset the counter %%%%%%%%%%

\section{Census data}
\label{sec:census}
Each compartment model is representative of one of the
districts defined in the main manuscript, and as such the compartment model is parametrised with the census data of the respective district, i.e., population counts stratified by age groups.
We use the 2011 United Kingdom census data made available by NOMIS\footnote{\url{https://www.nomisweb.co.uk}}.
This dataset contains census data for all of the considered districts for the following age groups: 0-4, 5-7, 8-9, 10-14, 15, 16-17, 18-19, 20-24, 25-29, 30-44, 45-59,60-64,65-74,75-84,85-89, 90-90+.

To be compatible with our model, we need to map this census data to the age structure imposed by the Eames contact matrix: i.e., 0-4 years (children), 5-18 years  (adolescents), 19-64 years (adults), 65-90+ years (elderly).
To define this mapping, we will refer to the number of individuals with the symbol $N$, subscripted with the dataset type (i.e., NOMIS or Eames) and the age group.

For the age group 0-4 and 65-90+ we have a direct mapping:
\begin{equation}
\begin{split}
N_{\text{Eames},\text{Children}} = N_{\text{NOMIS},\text{0-4}} \\
N_{\text{Eames},\text{Elderly}} = N_{\text{NOMIS},\text{65-90+}}
\end{split}
\end{equation}

However, as for the contact matrix the adolescents and adults are split between age 18 and 19, and for the census data these 2 age groups are aggregated, we need to make a custom mapping.
We will aggregate all shared age groups and divide the common age group in two:
\begin{equation}
\begin{split}
N_{\text{Eames},\text{Adolescents}} = N_{\text{NOMIS},\text{5-17}} + \ceil[\bigg]{\frac{N_{\text{NOMIS},\text{18-19}}}{2}}\\
N_{\text{Eames},\text{Adults}} = \ceil[\bigg]{\frac{N_{\text{NOMIS},\text{18-19}}}{2}} + N_{\text{NOMIS},\text{20-64}} \\
\end{split}
\end{equation}

\begin{figure}
  \centering
   \includegraphics[width=0.7\textwidth]{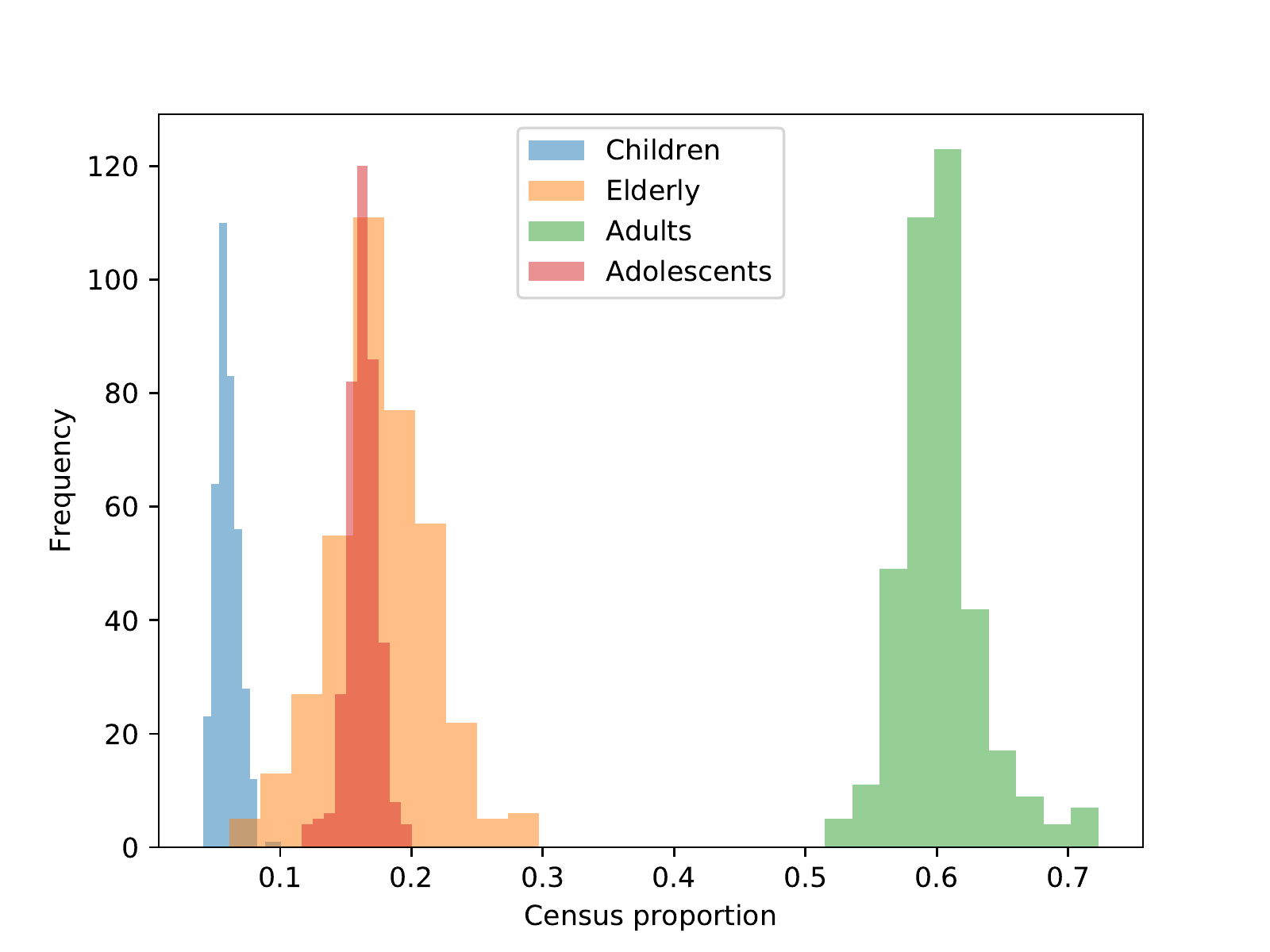}
     \caption{Histograms of the census proportions in the districts of Great Britain, according to Eames' age structure.}
      \label{fig:gb_eames_agestructure}
\end{figure}

%In figure~\ref{fig:gb_eames_agestructure} we observe that
When restructuring the census data according to the Eames age groups, we observe clear trends over the districts with respect to the proportion of children, adolescents, adults and elderly, as shown in Figure~\ref{fig:gb_eames_agestructure}.
However, the histograms in Figure~\ref{fig:gb_eames_agestructure} only show the marginalized distribution per age group.
To reason about the distribution over all age groups, consider that we have a proportion of each of the age groups, and we thus can represent this data as a positive simplex \cite{aitchison1983principal}, as defined in Definition~\ref{eq:positive_simplex}.

\begin{definition}[Unit simplex]
A unit simplex \cite{aitchison1997one}, with $D$ components, corresponds to the set:
\begin{equation}
\simplexset^D = \{\langle x_1,...,x_D\rangle \mid \forall x_i:x_i>0, \sum_{i=1}^{D}x_i = 1\}.
\end{equation}
\label{eq:positive_simplex}
\end{definition}

\begin{figure}
  \centering
   \zoombarycentric{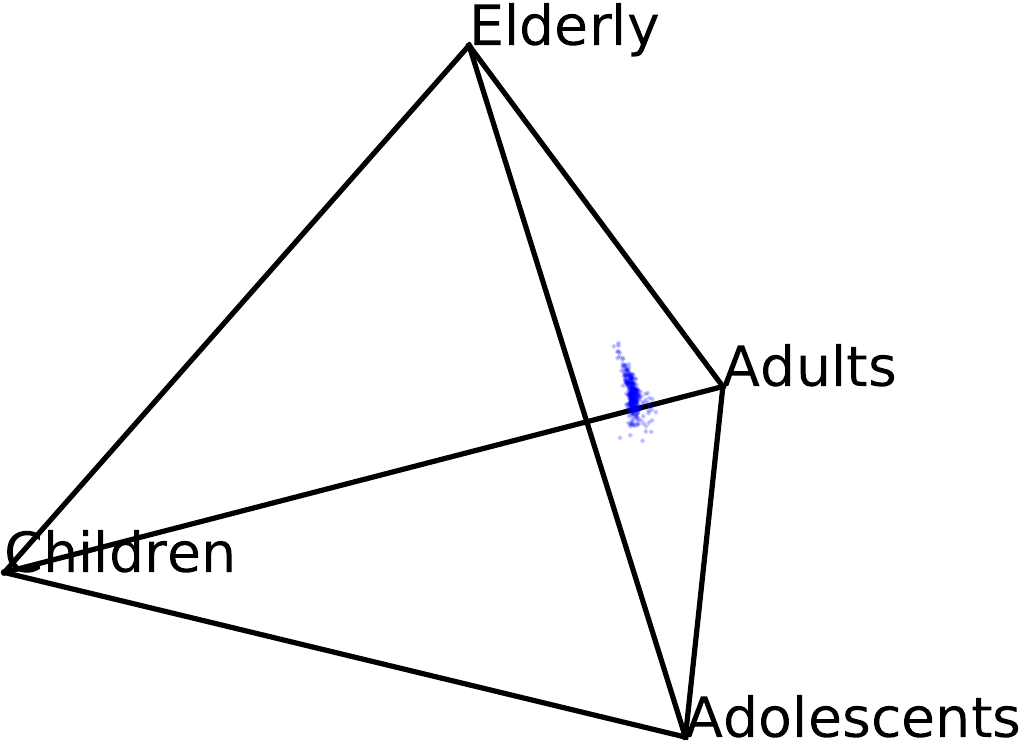}
     \caption{Barycentric projection of the census proportions in the districts of Great Britain, according to Eames' age structure. Each scatter point corresponds to one district, and each axis corresponds to the proportion of the age groups it connects. The left panel shows the original census pyramid, and the right panel zooms in on the point cloud.}
      \label{fig:gb_eames_agestructure_simplex}
\end{figure}

This representation enables us to reason about this data in a statistical framework, and to visualize the four-dimensional data in a three-dimensional space by using the Barycentric coordinate system, as shown in Figure~\ref{fig:gb_eames_agestructure_simplex}.
Figure~\ref{fig:gb_eames_agestructure_simplex} shows that the census distribution exhibits a dense region with a limited number of outliers.

%cd ~/projects/spatial-pandemic-py/src
%created with UK_Eames_AgeStructure.py
%python UK_Eames_AgeStructure.py --grouped-census-fn ~/projects/uk-districts/census/2011/gb.census.eames.csv --out-fn gb_eames_agestructure.pdf

%cd ~/research/phd-thesis/latex/figures/spatial-rl/compositional-census
%python plot_census.py --eames_census_fn ~/projects/uk-districts/census/2011/gb.census.eames.csv --output gb_census_eames_compositional.pdf
%pdfcrop gb_census_eames_compositional.pdf

Note that we use the 2011 census dataset, rather than the more recent 2018 census dataset, to be fully compatible with the mobility dataset used to inform our between-patch transition model (see Section \ref{sec:between_patch_model}).

For each district, the base contact matrix is corrected to make it reciprocal, using that district's census data.

\section{Parametrising the model with $R_0$}
%DOLATER: restructure this section (see written text)
\label{subsec:r0}
In influenza modelling literature, it is common to parametrise the model in terms of a specific $R_0$ value. 
We will now introduce an equation that enables us to compute the transmission probability $\probofinf$ for a given $R_0$ value.

To this end, we need to determine the next-generation matrix, which summarizes the number of secondary infections between age groups \cite{vynnycky2010introduction}, and determine the spectral radius of this matrix (see Definition~\ref{def:spectral_radius}).

\begin{definition}[Spectral radius]
The spectral radius of a matrix $L$, $\spectralradius(L)$, is the dominant eigenvalue of that matrix $L$.
\label{def:spectral_radius}
\end{definition}

As the contact matrix is a square matrix with positive real entries, according to the Perron-Frobenius theorem, this eigenvalue exists and is unique.
Note that, in a graph, the spectral radius is a measure of the graph's connectivity \cite{lewis2011network}.
As our context matrix $M$ can be seen as a graph that represents how strongly the different age groups are connected, this notion of connectivity applies here as well.

Following \cite{diekmann2009construction} and \cite{eames2012measured}, we construct the next-generation matrix for our SEIR model:
\begin{equation}
K = \frac{\probofinf M}{\recoveryrate}
\end{equation}
Given $K$, we can compute $R_0$ as:
\begin{equation}
\label{eq:r_0}
R_0 = \frac{\probofinf}{\recoveryrate} \spectralradius(M).
\end{equation}

Using equation~\ref{eq:r_0}, we can now compute the transmission risk $\probofinf$ for a given $R_0$, $\recoveryrate$ and contact matrix $M$.

\begin{figure}
  \centering
  \begin{minipage}[t]{0.9\textwidth}
   \includegraphics[width=.45\textwidth]{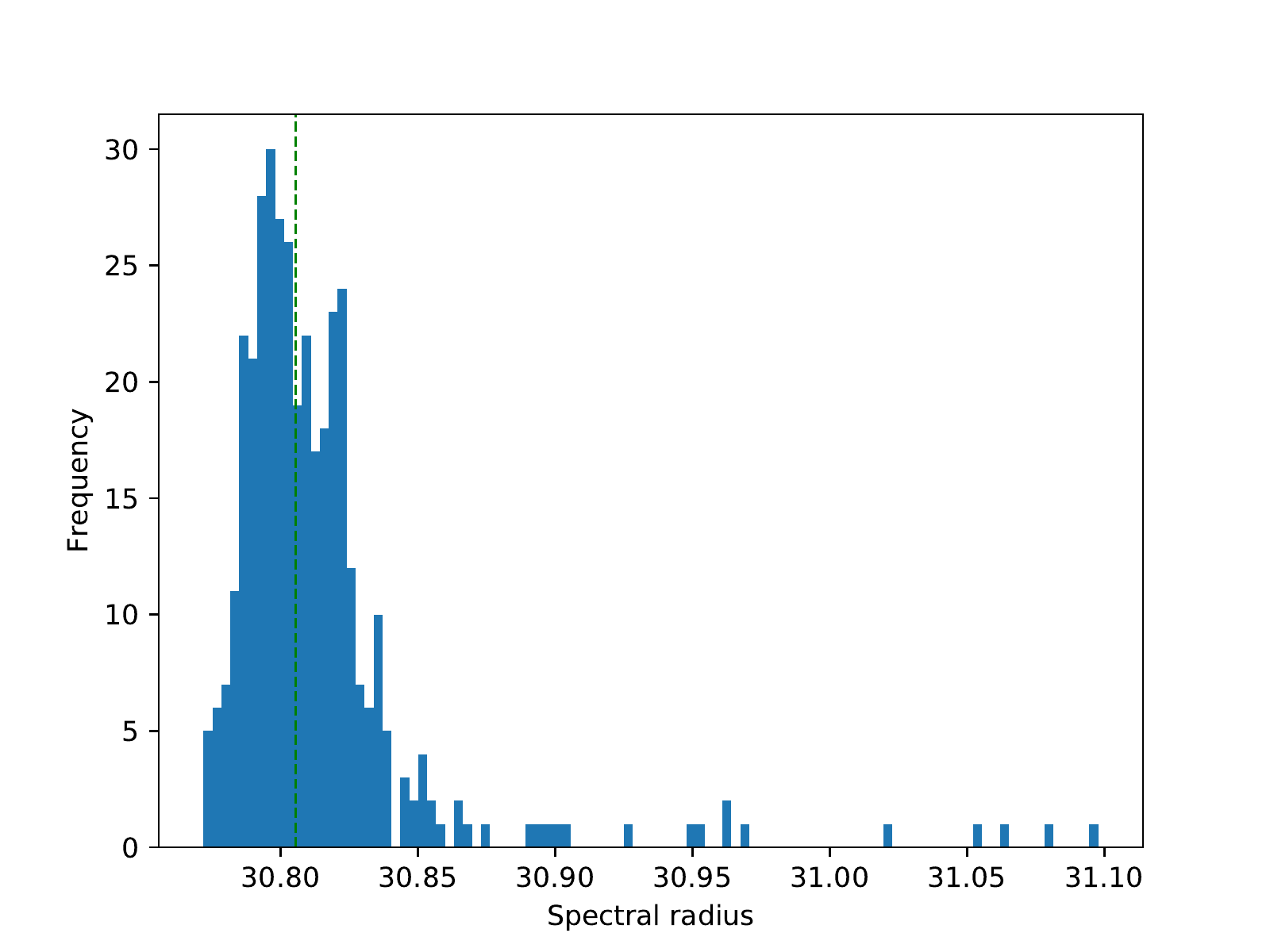}
   \includegraphics[width=.45\textwidth]{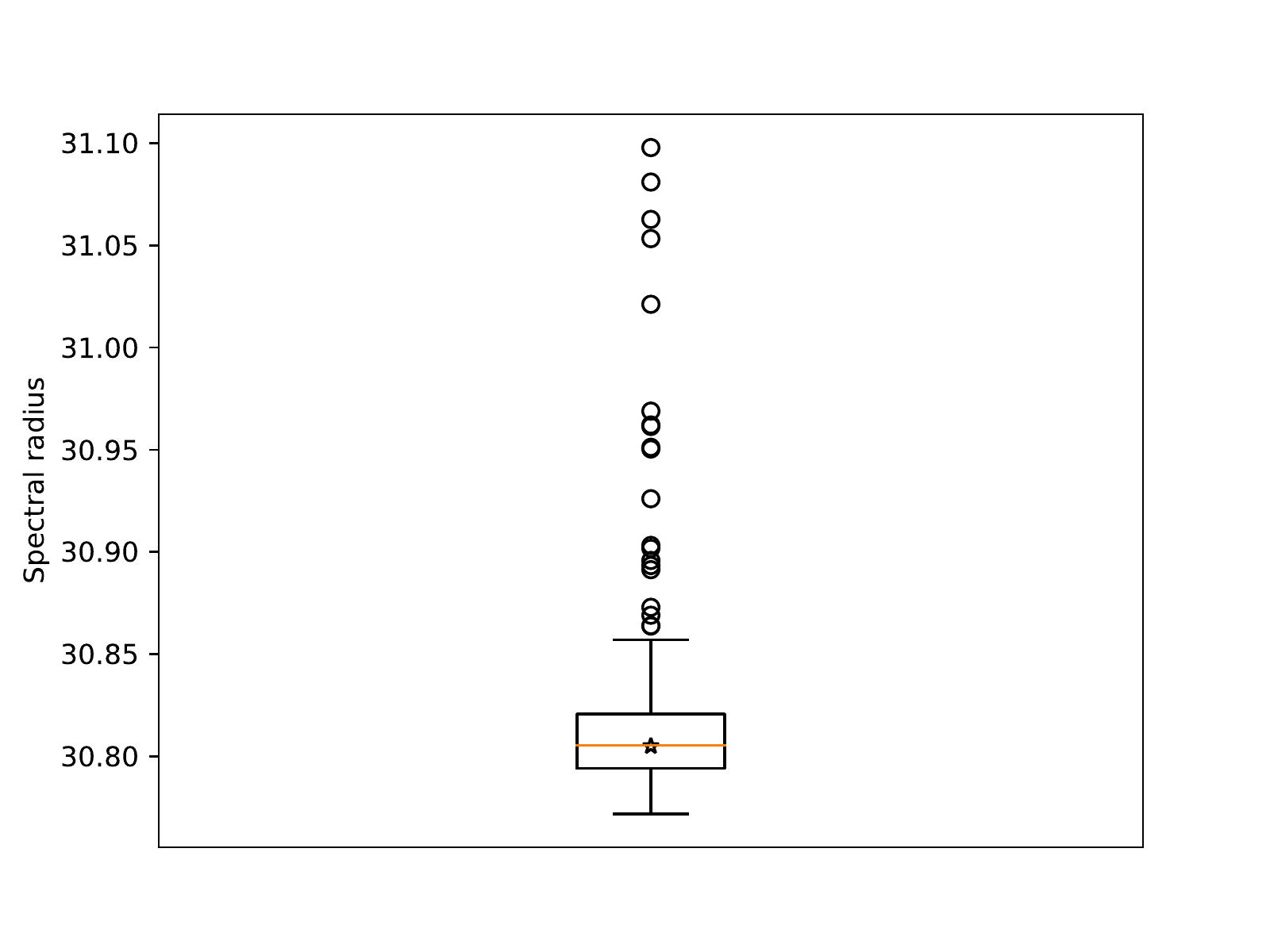}
   \end{minipage}
      \caption{Both figures display the distribution of the contact matrix' spectral radius for the different districts. To demonstrate the shape of this distribution, the left panel shows a histogram, annotated with a dotted green line that represents the median. To demonstrate the distribution with respect to its quartiles and outliers, the right panel shows a box plot of the distribution, annotated with an orange line that represents the median and a yellow star that shows the spectral radius for Great Britain.}
      \label{fig:spectral_radius}
\end{figure}

Note that for each district, we have a contact matrix that is corrected for reciprocity by using that district's census data.
Therefore, we have a distribution over $\spectralradius(M_d)$, where $M_d$ is a contact matrix for district $d$. 
We would expect that this distribution is centred around $\spectralradius(M_{GB})$, where $M_{GB}$ the contact matrix that is corrected for reciprocity using the census data representative for Great-Britain in its entirety (i.e., an aggregation of all the districts).
This is confirmed in Figure~\ref{fig:spectral_radius}, which shows that the median of the distribution over $\spectralradius(M_d)$ coincides with $\spectralradius(M_{GB})$.
Furthermore, note that the contact matrix denotes the average frequency of contacts that an individual in age group $i$ has with an individual in age group $j$.
Figure~\ref{fig:spectral_radius} thus shows a limited variance ($\sigma^2=0.001$).
% with boxplot whiskers which deviate no more than $0.029$ from the median, and the most extreme outlier deviating $0.245$ from the median.

%cd ~/projects/spatial-pandemic-py/src
%python UK_Eames_Eigen.py --grouped-census-fn ~/projects/uk-districts/census/2011/gb.census.eames.csv --contact-matrix-fn ~/projects/spatial-pandemic-py/data/contacts/eames/raw/conversational_school.csv --plot-type histogram
%python UK_Eames_Eigen.py --grouped-census-fn ~/projects/uk-districts/census/2011/gb.census.eames.csv --contact-matrix-fn ~/projects/spatial-pandemic-py/data/contacts/eames/raw/conversational_school.csv --plot-type boxplot

\section{Between-patch model}
\label{sec:between_patch_model}
Our model, that is comprised of a set of connected SEIR patches, is inspired by the recent BBC pandemic model \cite{klepac2018contagion}.
The BBC pandemic model was in its turn motivated by the model presented in \cite{gog2014spatial}.

At each time step, our model decides whether a patch $p$ becomes infected.
This is modulated by the patch's force of infection, which combines the potential of the infected patches in the system, weighted by a mobility model:
\begin{equation}
\label{eq:between_force_of_infection}
\forceofinfbetween_{p}(t) = \sum_{p' \in \patches} \mobilitymatrix_{p'p} \cdot \probofinf \cdot  \left(S^{\text{A}}_{p}(t)\right)^{\susceptiblecontribution} \cdot \infectedpotential_{p'}(t),
\end{equation}
where $\patches$ is the set of patches in the model, $\mobilitymatrix_{p'p}$ is the mobility flux between patch $p'$ and $p$, $\probofinf$ is the probability of transmission on a  contact, $S^{\text{A}}_{p}(t)$ is the susceptible population of adults in patch $p$ at time $t$ and its contribution is modulated by parameter $\susceptiblecontribution$, and $\infectedpotential_{p'}(t)$ is the infectious potential of patch $p'$ at time $t$.
We define this infectious potential as,
\begin{equation}
\infectedpotential_{p'}(t) = I^{\text{A}}_{p'}(t) \cdot M_{\text{A}\text{A}},
\end{equation}
where $I^{\text{A}}_{p'}(t)$ is the size at time $t$ of the infectious adult population and $M_{\text{A}\text{A}}$ is the average number of contacts between adults.

$\mobilitymatrix$ is a matrix based on the mobility dataset provided by NOMIS\footnote{We use the NOMIS WU03UK dataset that was released in 2011.}. This dataset describes the amount of commuting between the districts in Great-Britain.

In general, this between-patch model is constructed from first principles i.e., census data, a mobility model, the number of infected individuals and the transmission potential of the virus.
However, for the parameter $\susceptiblecontribution$ that modulates the contribution of the susceptibles in the receptive patch, while it is commonly used in literature \cite{gog2014spatial, eggo2010spatial, kissler2019geographic}, no such intuition is readily available.
Therefore, this parameter is fitted to match the properties of the epidemic that is under investigation \cite{gog2014spatial, eggo2010spatial, kissler2019geographic}.

To validate our model, we conduct two experiments.
Firstly, we compare our model to the original compartment model and perform a sensitivity analysis with respect to parameter $\susceptiblecontribution$.
Secondly, we show that our model fits the recent influenza pandemic of 2009, by choosing an appropriate value for all model parameters.

Given this time-dependent force of infection, we model the event that a patch becomes infected with a non-homogeneous Poisson process \cite{wang2018characterizing,tomba2008simple,barthelemy2010fluctuation}.
A Poisson process can be used to model the occurrence of events with a given intensity (see Definition~\ref{def:HPP}), and non-homogeneous Poisson processes generalize this concept to time-dependent intensities  (see Definition~\ref{def:NHPP}).
As the process' intensity depends on how the model (i.e., the set of all patches) evolves, we cannot sample the time at which a patch becomes infected a priori.
Therefore, we determine this time of infection using the time scale transformation algorithm \cite{cinlar2013introduction}.
Firstly, we explain the generic time scale transformation algorithm (Section~\ref{sec:time_scale_transformation}). Secondly, we adjust the algorithm to our setting (Section~\ref{sec:time_scale_transformation_epi}).

\subsection{Time scale transformation algorithm}
\label{sec:time_scale_transformation}
The time scale transformation algorithm enables us to determine the time at which an event, modelled by a non-homogeneous Poisson process, will take place \cite{cinlar2013introduction}.

%It does this by determining a random threshold on the cumulative intensity of the process.
%The event will be triggered when this intensity threshold is surpassed.

We will start by formally defining the homogeneous and non-homogeneous Poisson process.
A Poisson process is a counting arrival process, defined on a sample space $\Omega$ with probability measure $P$.

\begin{definition}[Arrival process]
An arrival process is a stochastic process \cite{cinlar2013introduction},
\begin{equation}
\arrivalprocess = \{ \arrivalprocess_{t}; t \geq 0 \},
\end{equation}
such that for any $\omega \in \Omega$, the mapping $t \rightarrow \arrivalprocess_t(\omega)$, has $\arrivalprocess_0 = 0$, is non-decreasing, increases only by integer jumps and is right continuous.
\end{definition}

\begin{definition}[Homogeneous Poisson process]
\label{def:HPP}
A \textit{homogeneous} Poisson process is an arrival process \cite{cinlar2013introduction,bertsekas2002introduction},
\begin{equation}
\poissonprocess = \{ \poissonprocess_{t}; t \geq 0 \},
\end{equation}
for which these axioms hold:
\begin{enumerate}
\item for almost all $\omega \in \Omega$, $t \rightarrow \poissonprocess_t(\omega)$ jumps in steps of size 1
\item the number of arrivals within any interval $[t..t+s]$, is independent of the history of arrivals prior to $t$
\item the process is time-homogeneous
\end{enumerate}
\end{definition}

%(DOLATER: add figure like in "Cinlar's" Stochastic processes, Fig 4.1.1, but with multiple samples (in different colors?))

From this definition, we can show that for each homogeneous Poisson process $\poissonprocess$:
\begin{equation}
\forall t \geq 0: P(\poissonprocess_t = k) = \frac{e^{-\ppintensity t}(\ppintensity t)^k}{k!},
\end{equation}
for some constant $\ppintensity \geq 0$, where $\ppintensity$ signifies the intensity (i.e., rate) of the process.

The concept of a Poisson process can be generalized to a \textit{non-homogeneous} Poisson process by removing the time-homogeneity requirement:
\begin{definition}[Non-homogeneous Poisson process]
\label{def:NHPP}
A \textit{non-homogeneous} Poisson process is an arrival process \cite{cinlar2013introduction},
\begin{equation}
\poissonprocess^{\ppintensity(t)} = \{ \poissonprocess^{\ppintensity(t)}_{t}; t \geq 0 \},
\end{equation}
for which these axioms hold:
\begin{enumerate}
\item for almost all $\omega \in \Omega$, $t \rightarrow \poissonprocess_t(\omega)$ jumps in steps of size 1
\item the number of arrivals within any interval $[t,t+s]$, is independent of the history of arrivals prior to $t$
\end{enumerate}
$\poissonprocess^{\ppintensity(t)}_{t}$ has a time-dependent rate that is specified by the intensity function $\ppintensity(t)$, where $\ppintensity(t) \geq 0$.
\end{definition}

We define the process' cumulative intensity function:
\begin{definition}[Cumulative intensity function]
\label{def:cum_intensity_function}
A non-homogeneous Poisson process $\poissonprocess^{\ppintensity(t)}$ with intensity function $\ppintensity(t)$ has a cumulative intensity function:
\begin{equation}
\ppcumintensity(t) = \int_{0}^{t} \ppintensity(s) ds
\end{equation}
\end{definition}

Furthermore, we can show that \cite{osaki2012applied}:
\begin{equation}
\E[\poissonprocess^{\ppintensity(t)}_{t+h} - \poissonprocess^{\ppintensity(t)}_{t}] = \int_{t}^{t+h} \ppintensity(s) ds,
\end{equation}
and thus we have that $\ppcumintensity(t)$ is the expectation function of $\poissonprocess^{\ppintensity(t)}_{t}$:
\begin{equation}
\ppcumintensity(t) = \E[\poissonprocess^{\ppintensity(t)}_{t}]
\end{equation}

From Definition~\ref{def:cum_intensity_function}, it is clear that $\ppcumintensity(t)$ will be a non-decreasing function and at least right-continuous.

\begin{figure}
\centering
   \includegraphics[width=0.7\textwidth]{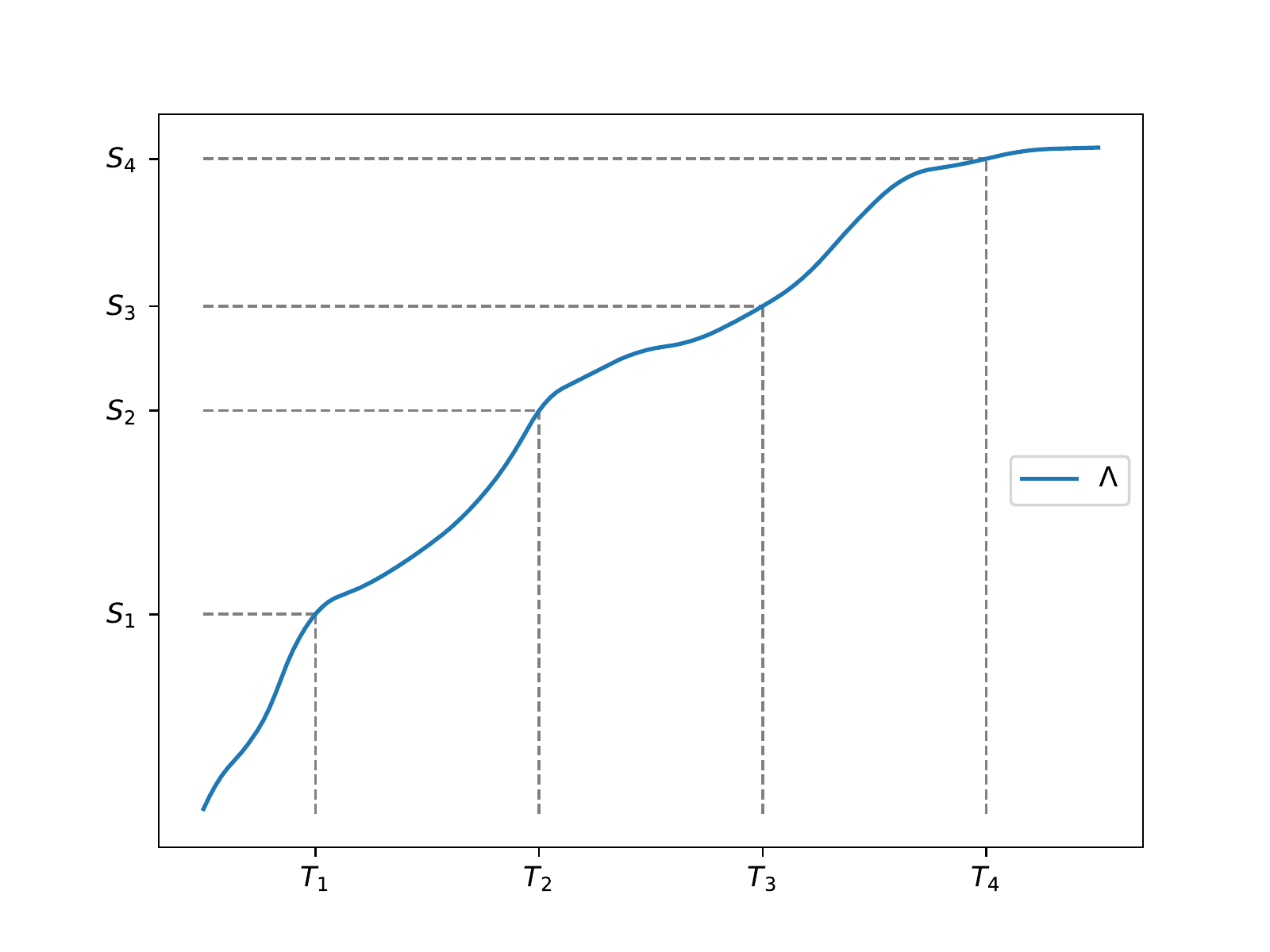}

  \caption{A visual example of Theorem~\ref{mapping_stationary_poisson_process}: $T_1,T_2,...$ form a non-homogeneous Poisson process with expectation function $\ppcumintensity$ if and only if $S_1,S_2,...$ form a homogeneous Poisson process with rate $1$.}
      \label{fig:mapping_theorem}
\end{figure}

The crucial theorem that underlies the time scale transformation algorithm denotes that the arrival times in a non-homogeneous Poisson process can be mapped to a homogeneous Poisson process with rate $1$ (Theorem~\ref{mapping_stationary_poisson_process}).
We present an example that demonstrates this theorem in Figure~\ref{fig:mapping_theorem}.

\begin{theorem}[Mapping non-homogeneous Poisson processes]
\label{mapping_stationary_poisson_process}
Let $\ppcumintensity$ be a continuous non-decreasing cumulative intensity function. Then,
\begin{equation}
T_1,T_2,...
\end{equation}
are the arrival times in a non-homogeneous Poisson process if and only if
\begin{equation}
\ppcumintensity(T_1),\ppcumintensity(T_2),...
\end{equation}
are the arrival times in a homogeneous Poisson process with rate $1$ \cite{cinlar2013introduction}.
\end{theorem}

%(DOLATER: to understand this theorem, can we give a visual explanation of the proof?)

%cd ~/research/phd-thesis/latex/figures/spatial
%python mapping_theorem.py --output mapping_theorem.pdf

The time scale transformation algorithm uses the relation in Theorem~\ref{mapping_stationary_poisson_process} to transform a homogeneous Poisson process with $\ppintensity=1$ into a non-homogeneous Poisson process with expectation function $\ppcumintensity$. The homogeneous process is formed by sampling from an exponential probability distribution with $\ppintensity=1$.
To make this transformation possible, a time inverse function of $\ppcumintensity$ is required:

\begin{definition}[Time inverse of $\ppcumintensity(t)$]
\label{def:time_inverse}
The time inverse $\timeinverse$ of an expectation function $\ppcumintensity(t)$ for a non-homogeneous Poisson process $\poissonprocess^{\ppintensity(t)}_{t}$:
\begin{equation}
\timeinverse(s) = \inf\{t: \ppcumintensity(t) > s \}
\end{equation}
\end{definition}

In Algorithm~\ref{alg:time_scale_transformation}, we formalize this procedure.
At each step $i$, we obtain a sample $X_i$ from an exponential probability distribution with rate $\ppintensity=1$, which is added to the set of samples $\mathcal{X}_i$.
The sum of the elements in $\mathcal{X}_i$ represents the $i^{\text{th}}$ arrival in the homogeneous Poisson process, and is transformed into the $i^{\text{th}}$ arrival in the non-homogeneous Poisson process using the inverse time function $\timeinverse(.)$.

\begin{algorithm}[!h]
%\SetKwInput{KwData}{Given}
%\KwData{$\timeinverse(.)$}
$\mathcal{X}_0 = \varnothing$\\
\For{ $i = 1, \ \dots$}{
$X_i \sim \mathcal{E}\text{xp}(\ppintensity=1)$\\
$\mathcal{X}_i = \mathcal{X}_{i-1} \cup  \{X_i\}$\\
$t_i = \timeinverse\left(\sum\limits_{x \in \mathcal{X}_i} x\right)$
}
\caption{Time scale transformation algorithm}
\label{alg:time_scale_transformation}
\end{algorithm}

\subsection{Time scale transformation algorithm to model the infection of patches}
\label{sec:time_scale_transformation_epi}
In order to use the time scale transformation algorithm (Algorithm~\ref{alg:time_scale_transformation}) in our epidemiological model, note that
the patches' internal state is updated in a discrete number of time steps.
We determine a patch's intensity $\forceofinf_p(t)$ (Equation~\ref{eq:between_force_of_infection}) at the end of each day.
This results in a sequence of intensities between which we linearly interpolate to obtain a piecewise linear intensity function:
\begin{equation}
\label{eq:interpolated_intensity_function}
\ppintensity(t) = \text{line}(t, \lceil t \rceil - 1, \lceil t \rceil , \forceofinf_p(t)),
\end{equation}
where
\begin{equation}
\text{line}(x, x_1, x_2, f) = f(x_1) + \frac{f(x_2) - f(x_1)}{x_2-x_1}(x-x_1)
\end{equation}
interpolates linearly between $(x_1, f(x_1))$ and $(x_2, f(x_2))$.

This piecewise linear intensity function $\ppintensity(t)$ is continuous and thus its cumulative counterpart $\ppcumintensity(t)$ is continuous as well.
Furthermore, as $\forceofinf_p(t) \geq 0$ for all $t \geq 0$, $\ppcumintensity(t)$ is non-decreasing.

%cd ~/research/phd-thesis/latex/figures/spatial
%python intensity_function_example.py --output intensity_function_example.pdf
\begin{figure}
  \centering
  \includegraphics[width=0.7\textwidth]{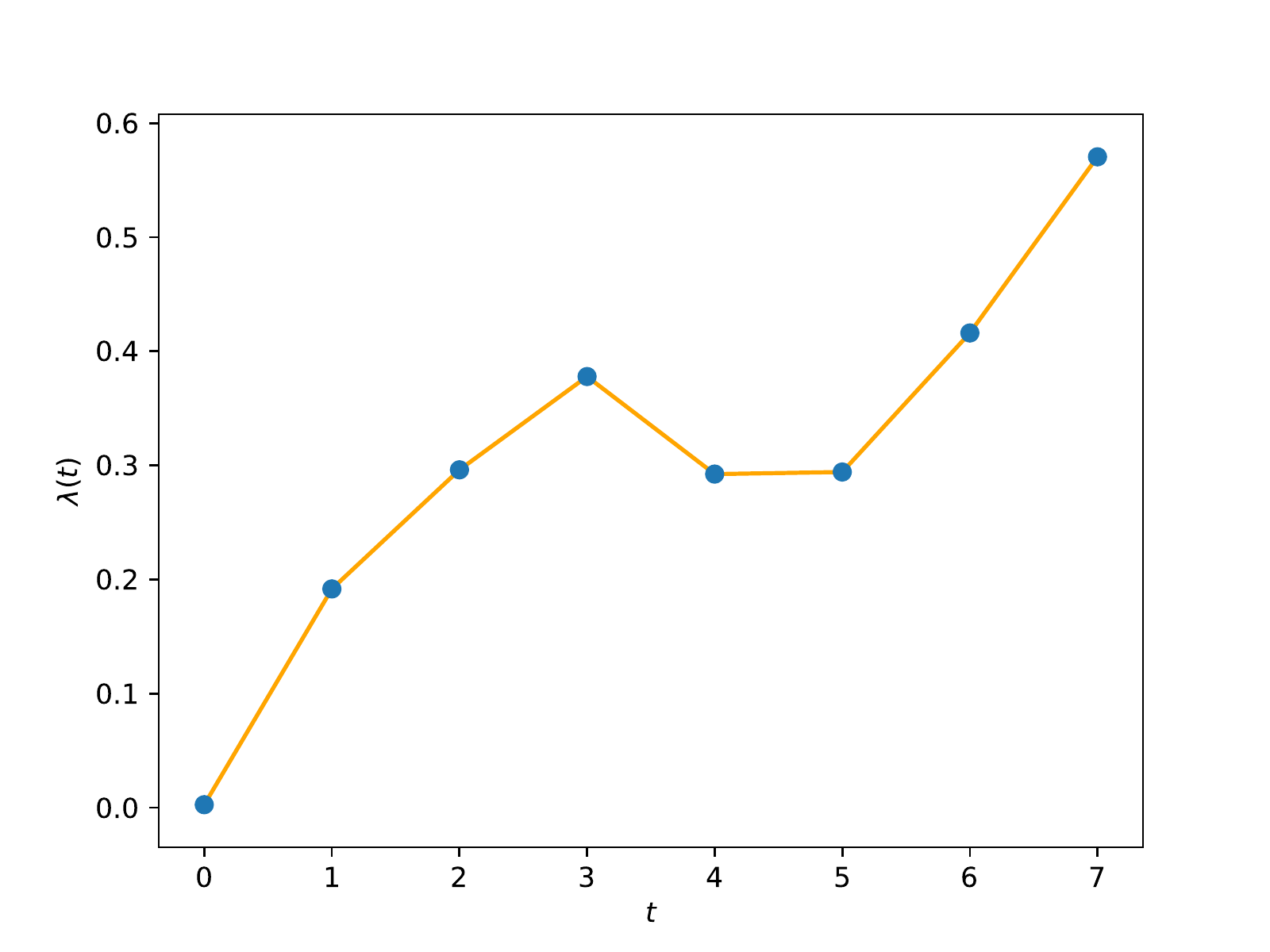}
    \caption{An example of a piecewise linear intensity function for a patch in our model (see Equation~\ref{eq:interpolated_intensity_function}). The blue scatter points represent the evaluation of $\forceofinf(t)$ (Equation~\ref{eq:between_force_of_infection}) at discrete time steps (i.e., the end of each day). The orange connecting lines represent the linear interpolation between $\forceofinf(i-1)$ and $\forceofinf(i)$.}
     \label{fig:intensity_function_example}
\end{figure}

As $\forceofinf_p(t)$ depends on the simulation state at time $t$, it is clear that we cannot evaluate this function beyond the current simulator time step.
However, the definition of the time inverse $\timeinverse$ (Definition~\ref{def:time_inverse}) shows that we can use the arrival time in the homogeneous Poisson process as a threshold for the arrival time in the non-homogeneous Poisson process.
We formalize this threshold-based time scale transformation algorithm in Algorithm~\ref{alg:time_scale_transformation_threshold}.
Note that this algorithm approximates the original algorithm as we check whether the threshold is surpassed at discrete time steps.

\begin{algorithm}[!h]
$X \sim \mathcal{E}\text{xp}(\ppintensity=1)$\\
\For{ $t = 1, \ \dots$}{
\If{$\ppcumintensity(t) \geq X$} {
Trigger event\\
$X^{(t)} \sim \mathcal{E}\text{xp}(\ppintensity=1)$\\
$X = X + X^{(t)}$\\
}
}
\caption{Time scale transformation algorithm using discrete time steps}
\label{alg:time_scale_transformation_threshold}
\end{algorithm}

\section{Model validation}
\label{sec:si_spatial_model_validation}
Our objective is to construct a model that is representative for contemporary Great Britain with respect to population census and mobility trends.
This model is to be used to study school closure intervention strategies for future influenza pandemics.
While in many studies \cite{kissler2019geographic,gog2014spatial,eggo2010spatial}, a model is created specifically to fit one epidemic case, we aim for a model that is robust with respect to different epidemic parameters, most importantly $R_0$, the basic reproduction number.

To validate our model according to these goals, we conduct two experiments.
In the first experiment, we compare our patch model to a SEIR compartment model that uses the same contact matrix and age structure.
While we do not expect our model to behave exactly like the compartment model, as the patches and the mobility network that connects them induces a different dynamic, we do expect to see similar trends with respect to the epidemic curve and peak day.
In the second experiment we show that our model is able to reproduce the trends that were observed during the 2009 influenza pandemic, commonly known as the swine-origin influenza pandemic, that originated in Mexico.
The 2009 influenza pandemic in Great Britain is an interesting case to validate our model for three main reasons.
Firstly, the pandemic occurred quite recently and thus our model's census and mobility scheme should be a good fit, as both the datasets on which we base our census and mobility model were released in 2011.
Secondly, due to the time when the virus entered Great Britain, the summer holiday started 11 weeks after the emergence of the epidemic.
The timing of the holidays had a severe impact on the progress of the epidemic and resulted in a epidemic curve with two peaks.
This characteristic epidemic curve enables us to demonstrate the predictive power of our age-dependent contact model with support for school closures.
Thirdly, the number of symptomatic cases that occurred in Great Britain during the 2009 pandemic was recorded meticulously and is publicly available \cite{kubiak20122009}.

\subsection{Comparison to the Eames SEIR compartment model}
In this experiment, we compare our patch model to a simple SEIR model that encompasses the same age structure and contact matrix \cite{eames2012measured}, to which we will refer as Eames-SEIR from this point onwards.
We consider a stochastic implementation of the Eames-SEIR.
This experimental setting will be central to the reinforcement learning experiments, related to finding optimal school closure policies, that we present in the main manuscript.

Following \cite{eames2012measured} and \cite{baguelin2010vaccination}, we use a latent period of one day ($\latencyrate=\frac{1}{1}$) and an infectious period of 1.8 days ($\recoveryrate=\frac{1}{1.8}$).
We perform our experiment for a set of $R_0$ values within the range of $1.4$ to $2.4$, in steps of 0.2. This range is considered representative for the epidemic potential of influenza pandemics \cite{Basta2009,Medlock2009}.

\begin{figure}
  \centering
   \includegraphics[width=0.7\textwidth]{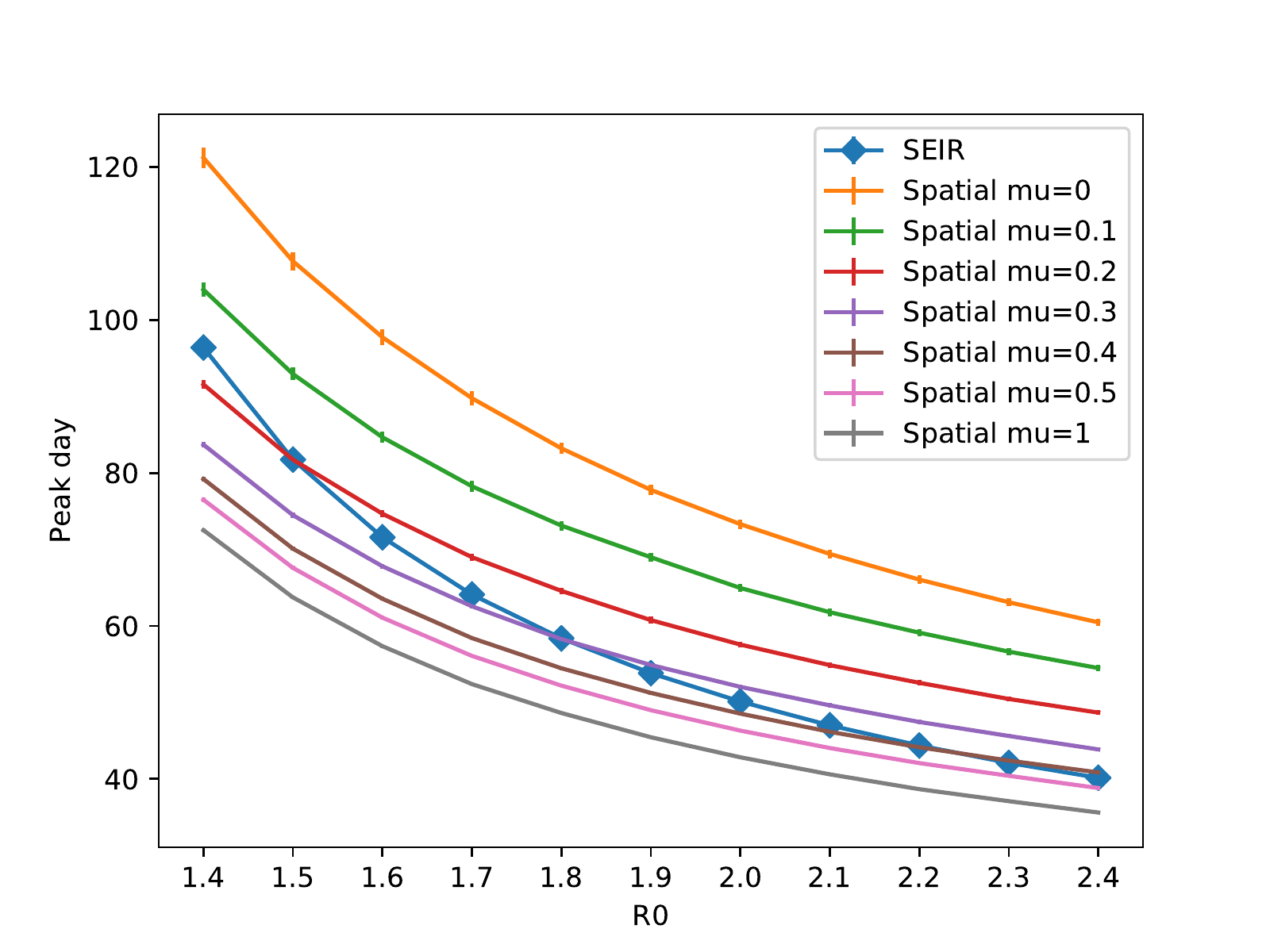}
     \caption{Time of peak day (y-axis) for $R_0 \in \{1.4,1.6,1.8,2.0,2.2,2.4\}$ (x-axis). A curve is shown for different values of $\susceptiblecontribution$ (plain curve) and the peak days as produced by the SEIR-Eames model (curve with diamond scatter points). For each $R_0$, 100 stochastic trajectories were sampled and the bound signifies the 95\% confidence interval of the sample. }
      \label{fig:mu_sa}
\end{figure}

Furthermore, we need to choose a value for the parameter in the between-patch model, i.e., $\susceptiblecontribution$, that modulates the contribution of susceptible adults in the receiving patch (see Section~\ref{sec:between_patch_model}).
This parameter is typically fitted towards data, however, in this experiment and in the reinforcement learning experiments in the main manuscript, we consider a model to investigate future epidemics.
Our goal is to calibrate our model such that it produces peak days that are similar to the peak days in Eames-SEIR \cite{eames2012measured}, which is a prominent model for pandemic influenza that moreover generates peak days that are in agreement with earlier work \cite{ferguson2006strategies}.
Therefore, we investigate the effect of $\susceptiblecontribution$ in this setting, through a sensitivity analysis.
We consider $\susceptiblecontribution$ in the interval $[0,1]$, where the left end of the interval (i.e., $\susceptiblecontribution=0$) signifies that the contribution of susceptible adults is ignored and the right end of the interval (i.e., $\susceptiblecontribution=1$) signifies that the contribution of adults is not modulated.
In Figure~\ref{fig:mu_sa}, we show the results for the sensitivity analysis for $\susceptiblecontribution \in \{0,0.1,0.2,0.3,0.4,0.5,1\}$, together with the peak days for the Eames-SEIR model.
From these results, it is clear that the different values for $\susceptiblecontribution$ form a gradient within the interval $[0,1]$.

%DOLATER: now both SEIR-Eames and Eames-SEIR is used, use one term

%cd ~/research/spatial-pandemic-experiments/spatial/baseline
%./run_mus.sh
%
% (or on the vsc thinking)
% wsub -A lp_hiv_networks -batch run_mus.pbs -data run_mus.csv
%
%python plot_peak_day_sa.py --R0s "1.4,1.5,1.6,1.7,1.8,1.9,2.0,2.1,2.2,2.3,2.4" --n 100 --dir "./base/" --output mu_sa.pdf

\begin{figure}
  \centering
   \includegraphics[width=0.7\textwidth]{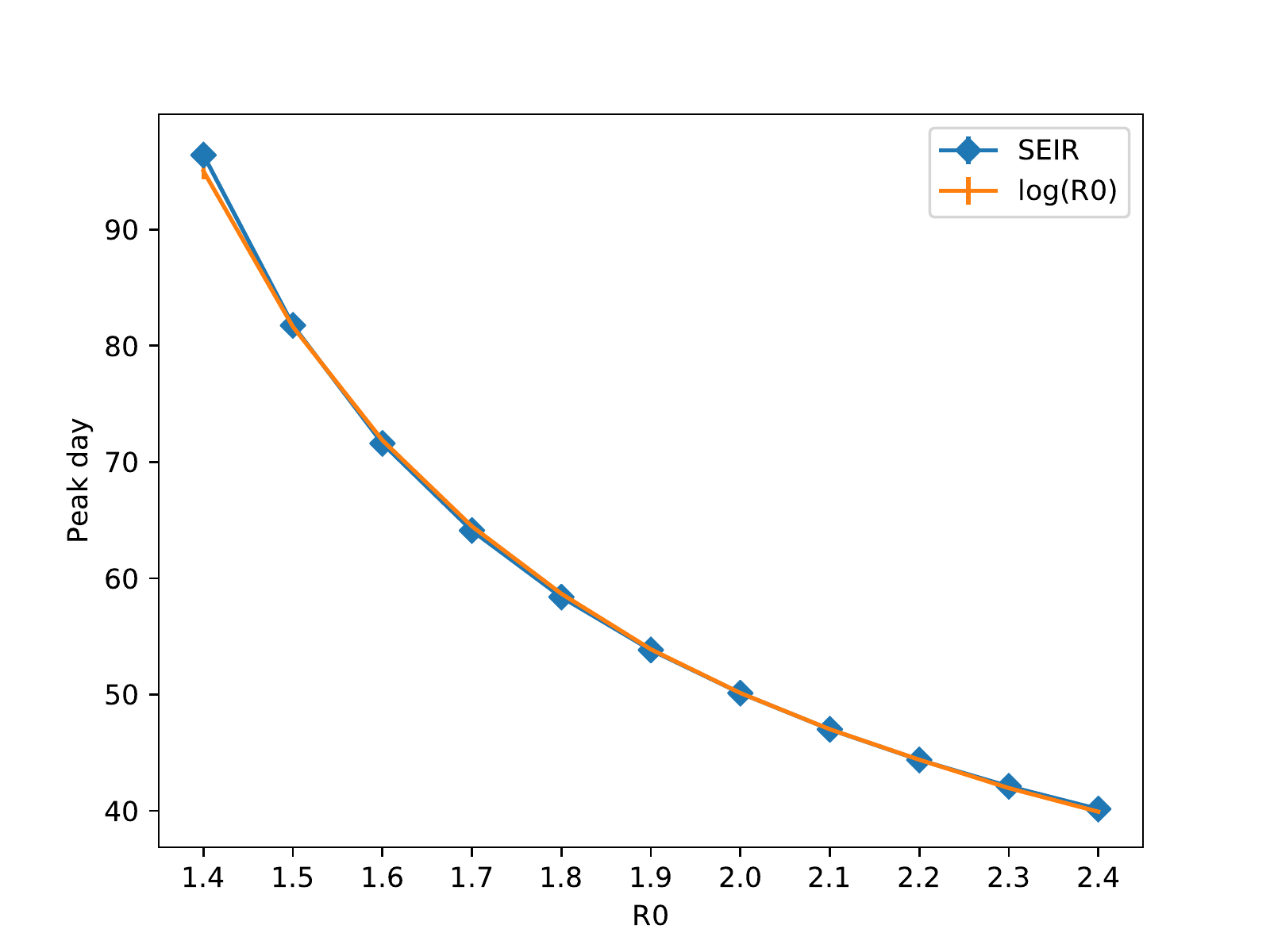}
     \caption{Number of peak days (y-axis) for $R_0 \in \{1.4,1.6,1.8,2.0,2.2,2.4\}$ (x-axis). A curve is shown for $\susceptiblecontribution=log(R_0) \cdot 0.6$ (orange curve) and the peak days as produced by the SEIR-Eames model (blue curve with diamond scatter points). For each $R_0$, 100 stochastic trajectories were sampled and the bound signifies the 95\% confidence interval of the sample. }
      \label{fig:mu_log}
\end{figure}

However, no value of $\susceptiblecontribution$ provides a good fit for all of the considered $R_0$'s, when comparing the peak days to the Eames-SEIR model.
Rather, we can discern a log-relationship between $\susceptiblecontribution$ and the best fit for the different $R_0$'s.
Based on this observation, we propose to define:
\begin{equation}
\label{eq:log_R0_mu}
\susceptiblecontribution=\log(R_0) \cdot \mathpzc{s},
\end{equation}
where $\mathpzc{s}$ is a scaling factor.
For this experimental setting, we find that $\mathpzc{s}=.6$ provides a good fit for all of the considered $R_0$'s, which we show in Figure~\ref{fig:mu_log}.

%cd ~/research/spatial-pandemic-experiments/spatial/baseline
%./run_mus.sh
%
% (or on the vsc thinking)
% wsub -A lp_hiv_networks -batch run_mus.pbs -data run_mus.csv
%
%python plot_peak_day_log.py --R0s "1.4,1.5,1.6,1.7,1.8,1.9,2.0,2.1,2.2,2.3,2.4" --n 100 --dir "./base/" --output mu_log.pdf

\begin{figure}
  \centering
  \begin{minipage}[t]{1\textwidth}
   \includegraphics[width=.5\textwidth]{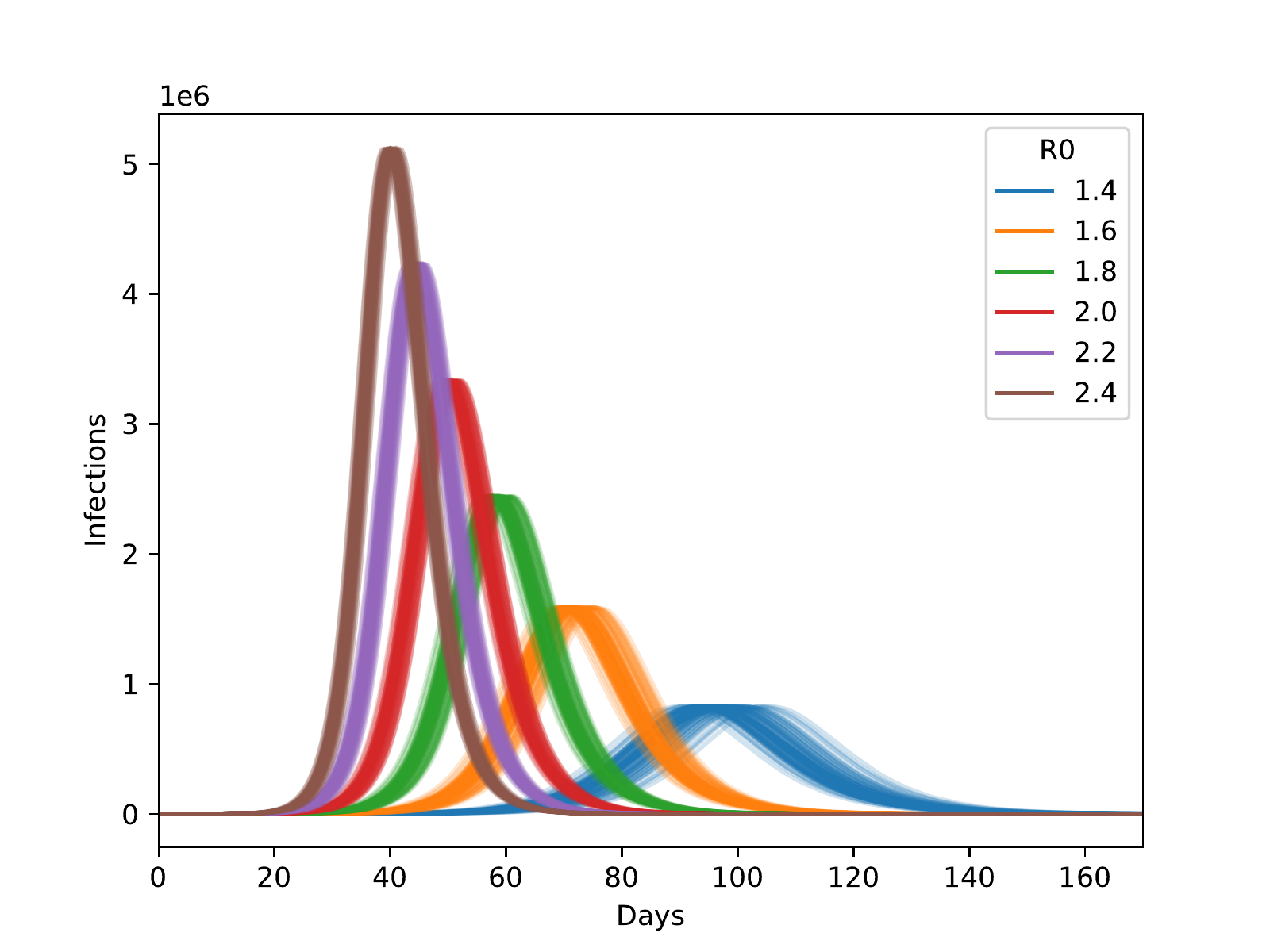}
   \includegraphics[width=.5\textwidth]{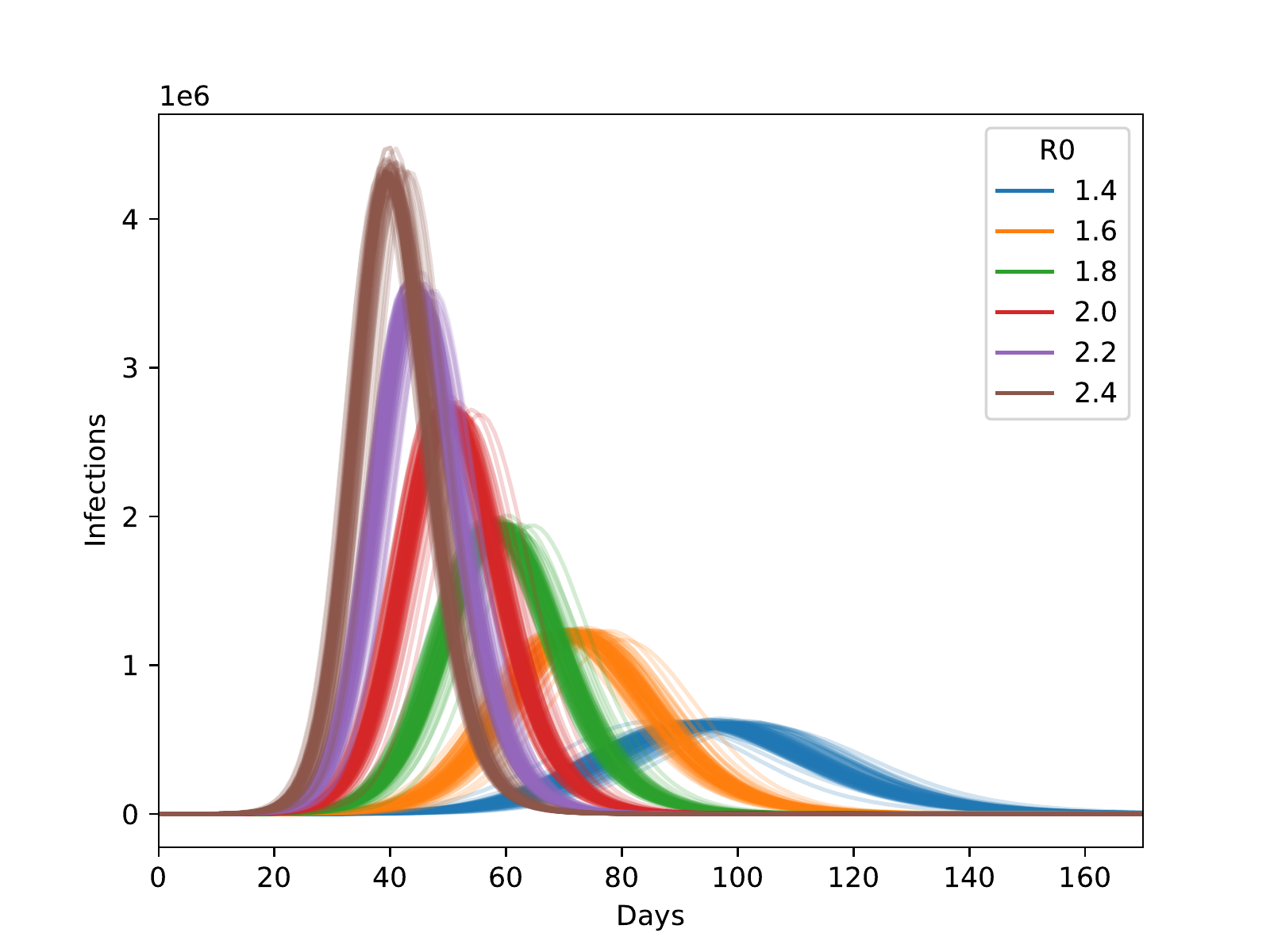}
   \end{minipage}
      \caption{Epidemic trajectories for the Eames-SEIR model (left panel) and the spatial model (right panel). One epidemic trajectory encodes the number of infections per day. Trajectory distributions are shown for $R_0 \in  \{1.4,1.6,1.8,2.0,2.2,2.4\}$, with a different colour per reproductive number. For each $R_0$, the distribution consists out of 100 trajectory samples.}
      \label{fig:first_exp_trajectories}
\end{figure}

Provided this choice of $\susceptiblecontribution$, when we compare the epidemic trajectories of our spatial model with the SEIR-Eames model in Figure~\ref{fig:first_exp_trajectories}, we observe similar trends with respect to the shape of the trajectory distributions.
The main difference is that the epidemic curves grow slower in our spatial model than in the Eames-SEIR model and also achieve a lower peak incidence.
This is expected, as we constrain mixing in our spatial model within the districts, and thus increase the resolution of our model, which has been shown to more accurately predict peak incidence \cite{mills2014spatial}.

%cd ~/research/spatial-pandemic-experiments/spatial/baseline
%./run_open.sh
%
% (or on the thinking vsc)
%wsub -A lp_hiv_networks -batch run_open.pbs -data run_open.csv
%
%python plot_simple_trajectories.py --R0s "1.4,1.6,1.8,2.0,2.2,2.4" --n 100 --dir "./base/" --school open --end 170 --output first_exp_simple_trajectories.pdf
%python plot_trajectories.py --R0s "1.4,1.6,1.8,2.0,2.2,2.4" --n 100 --dir "./base/" --school open --end 170 --output first_exp_spatial_trajectories.pdf

%cd ~/research/spatial-pandemic-experiments/spatial/baseline
%./run_open.sh
%python plot_first_day_per_district.py --R0s "1.4,1.6,1.8,2.0,2.2,2.4" --n 100 --dir "./base/" --end 70 --output first_day_per_district.pdf
\begin{figure}
  \centering
   \includegraphics[width=0.7\textwidth]{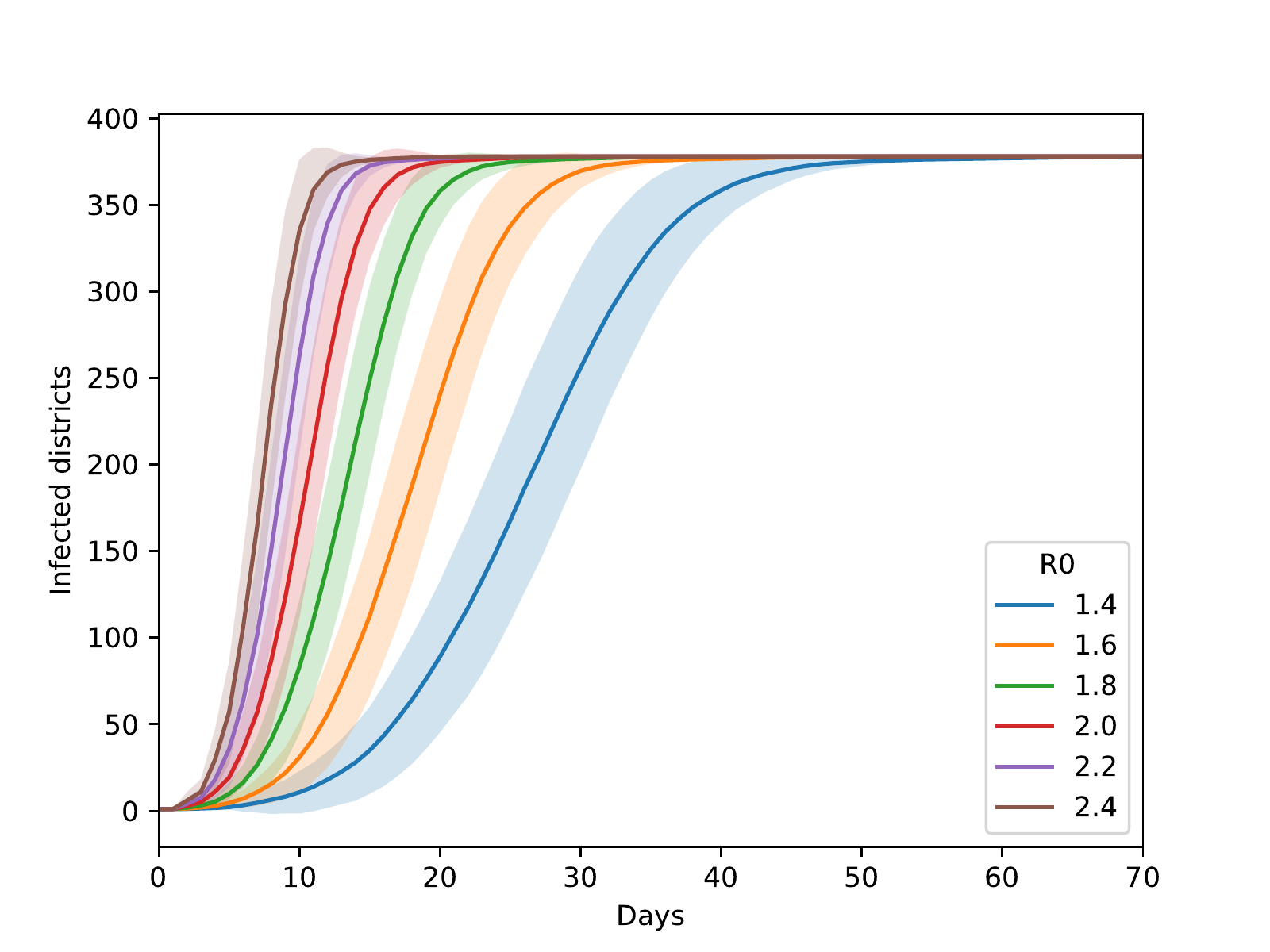}
     \caption{Number of infected districts (y-axis) per day (x-axis) for $R_0 \in \{1.4,1.6,1.8,2.0,2.2,2.4\}$. For each $R_0$, 100 stochastic trajectories were were sampled, of which the curve represents the mean, and the bound represents the standard deviation of the samples. }
      \label{fig:first_day_per_district}
\end{figure}

Furthermore, in Figure~\ref{fig:first_day_per_district}, we show the number of districts that get infected over time for different $R_0$ values.
This shows that all districts get infected, and the time it takes for all districts to reach this point depends mainly on the transmission-ability of the virus strain.
%DOLATER: The standard deviation bounds show that the amount of stochastic dispersion grows inversely proportional with the value of $R_0$
%(DOLATER: Timo: in Figure~\ref{fig:first_day_per_district}, it is hard to see the std, can you plot it separately per R0 (without the mean, just the std))

\begin{figure}
  \centering
   \includegraphics[width=0.7\textwidth]{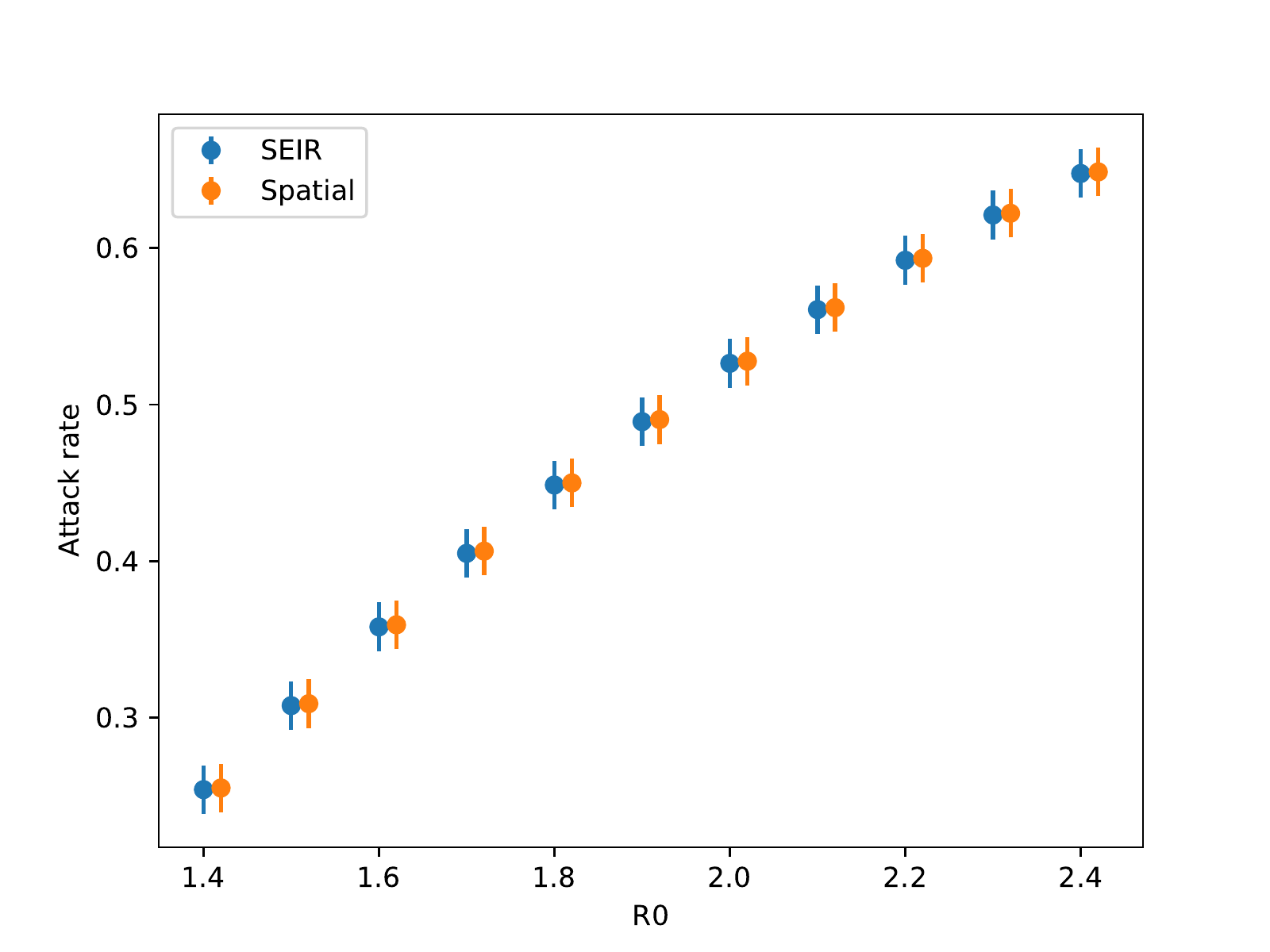}
     \caption{Attack rate (y-axis) for $R_0 \in \{1.4,1.6,1.8,2.0,2.2,2.4\}$ (x-axis). Results are shown for the Eames-SEIR model (blue scatter) and the spatial model (orange scatter). For each model, we depict the standard deviation as bars on top of the scatter points. For each $R_0$, 100 stochastic trajectories were obtained.}
      \label{fig:attack_rate_comparison}
\end{figure}

We expect the attack rate to be similar for the SEIR-Eames and spatial model. When all districts get infected, the attack rate in the spatial model is the sum of the attack rates of a set of SEIR-Eames models (i.e., one Eames-SEIR model per district).
We verified that the attack rates are indeed nearly identical, as shown in Figure~\ref{fig:attack_rate_comparison}, with little variance for either of the models.

%cd ~/research/spatial-pandemic-experiments/spatial/baseline
%./run_open.sh
%python plot_ar_simple.py --R0s "1.4,1.5,1.6,1.7,1.8,1.9,2.0,2.1,2.2,2.3,2.4" --n 100 --dir "./base/" --output attack_rate_comparison.pdf
%DOLATER, legend now says SEIR, make this Eames-SEIR

%(DOLATER: add plot with GB map where the infection is drawn on)
%(DOLATER: add plot with curve per district)

\subsection{2009 influenza pandemic in Great Britain}
\label{sec:si_gb_influenza_pandemic}
\begin{figure}

  \centering
  \includegraphics[width=0.7\textwidth]{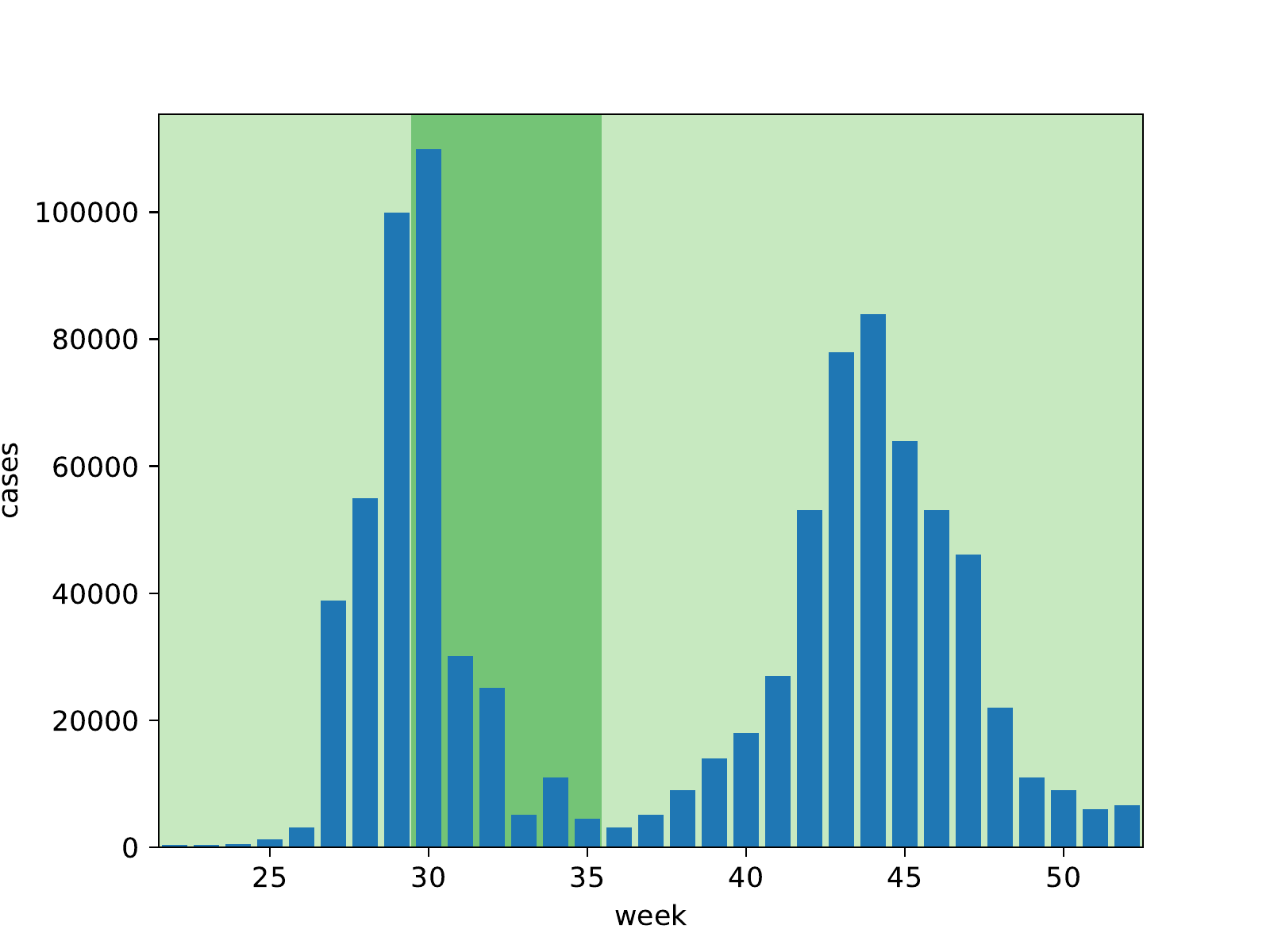}

  \caption{This figure shows the amount of cases that were recorded by the HPA on a weekly basis (blue bars). The background in this figure signifies the time of the summer holidays (dark green).}
     \label{fig:hpa_cases_and_holidays}
\end{figure}

The virus responsible for the 2009 influenza pandemic arrived in Great Britain during the first week of May 2009 (week 19).
Following this introduction, the epidemic grew for 11 weeks until the summer school holidays started, after which the epidemic showed its first peak.
After the school holidays, the epidemic was rekindled and grew to a second peak.
In Figure~\ref{fig:hpa_cases_and_holidays} we show the weekly case count, as recorded by the British Health Protection Agency (HPA) and the time at which the school holidays take place.

%cd ~/projects/spatial-pandemic-py/data/2009-pandemic-cases
%python plot.py -o hpa_cases_and_holidays.eps
%DOLATER: should the light-green in the above plot, be replaced with simple white?
%DOLATER: axes: cases -> Cases, week -> Week

To reproduce this distinctive epidemic curve, we use our original model as it was described in the main manuscript.
We consider two free parameters: the basic reproduction number and the time of the infectious period.
The general consensus is that the basic reproduction number was moderate during the 2009 influenza pandemic, with estimates ranging from $1.16$ to $2$ \cite{kubiak20122009,tuite2010estimated,de2009preliminary,fraser2009pandemic,yang2009transmissibility,nishiura2010pros,balcan2009seasonal}.
We present a detailed overview of the reported basic reproduction number estimates in Table~\ref{tab:R0_gb_2009}.
For the period of infectiousness we found estimates of $1.8$, $2.5$ and $3.38$ days \cite{eames2012measured,balcan2009seasonal, tuite2010estimated}.
We present a detailed overview of the infectious period estimates in Table~\ref{tab:inf_period_gb_2009}.

\begin{table}
\centering
\begin{tabular}{|l|l|}
\hline
\textbf{$R_0$} & \textbf{ Source}                                   \\
\hline
1.22-1.58   & \cite{fraser2009pandemic}        \\
1.3-1.7     & \cite{yang2009transmissibility}  \\
1.21-1.35   & \cite{nishiura2010pros}         \\
1.75        & \cite{balcan2009seasonal}        \\
1.87-2.07   & \cite{de2009preliminary}         \\
1.31        & \cite{tuite2010estimated}        \\
1.16-1.59   & \cite{kubiak20122009}            \\
\hline
\end{tabular}
\caption{Overview of basic reproduction numbers from literature.}
\label{tab:R0_gb_2009}
\end{table}

\begin{table}
\centering
\begin{tabular}{|l|l|}
\hline
\textbf{Infectious period} & \textbf{ Source}                                   \\
\hline
1.8   & \cite{eames2012measured}        \\
2.5    & \cite{balcan2009seasonal}  \\
3.38   & \cite{tuite2010estimated}         \\
\hline
\end{tabular}
\caption{Overview of infectious periods from literature.}
\label{tab:inf_period_gb_2009}
\end{table}

Given these prior estimates, we parametrize our model with a basic reproduction number that is in the range of $1.2$ to $2.0$ and consider a duration of infectiousness of respectively $1.8$, $2.5$ and $3.38$ days.

\begin{figure}
  \centering
  \begin{minipage}[t]{1\textwidth}
  \centering
   \includegraphics[width=.45\textwidth]{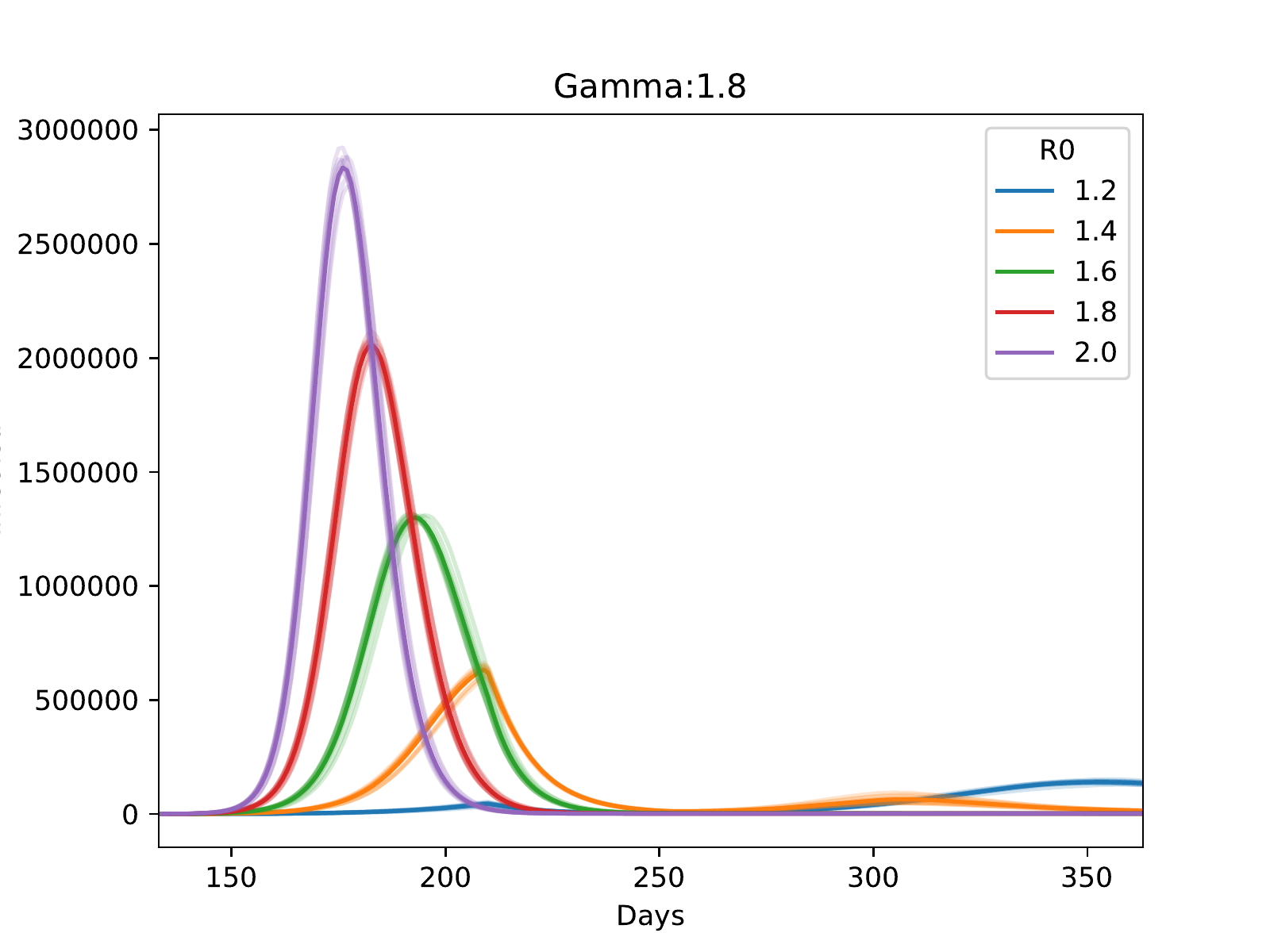}
   \end{minipage}
  \begin{minipage}[t]{1\textwidth}
  \centering
   \includegraphics[width=.45\textwidth]{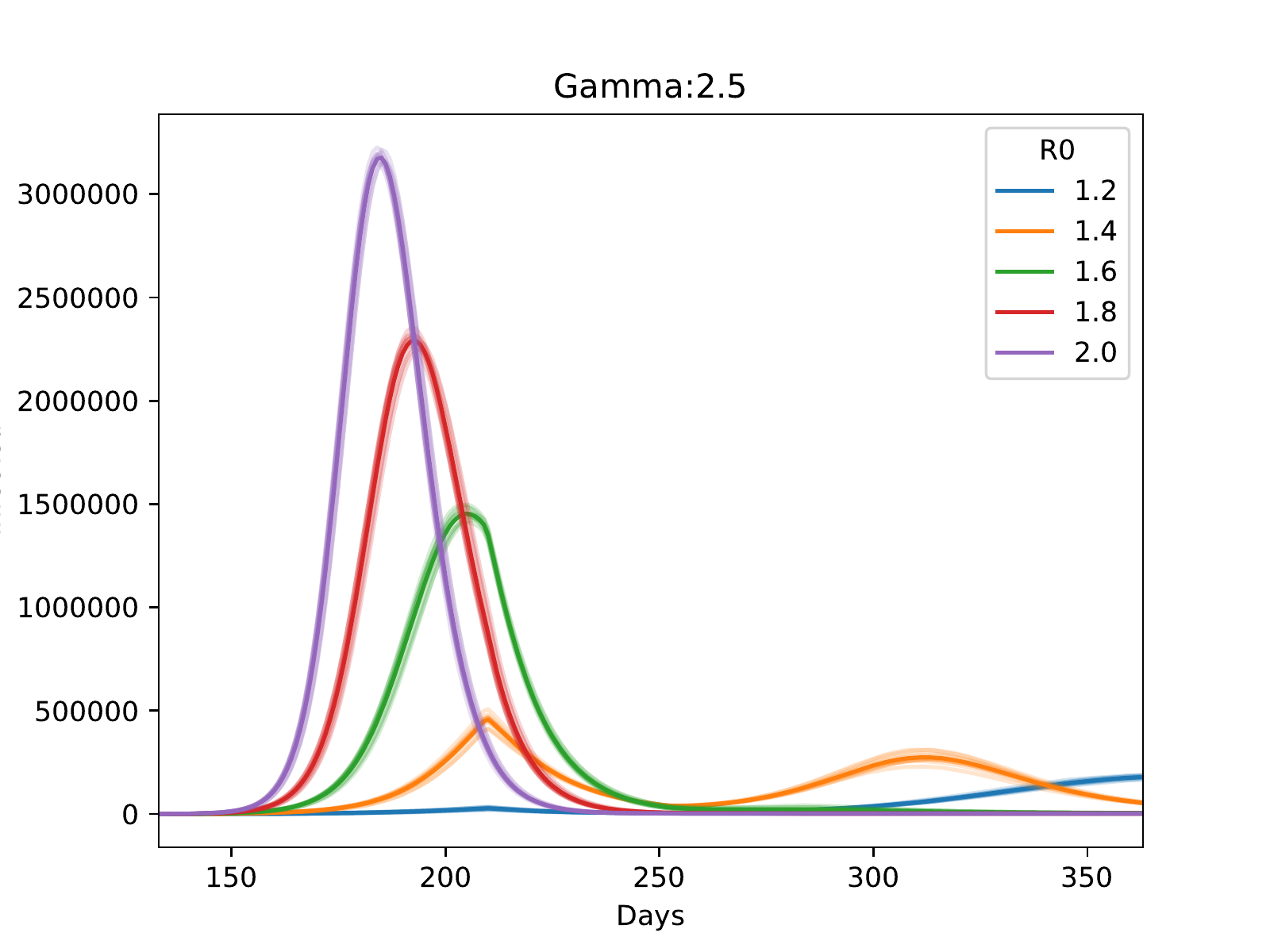}
   \includegraphics[width=.45\textwidth]{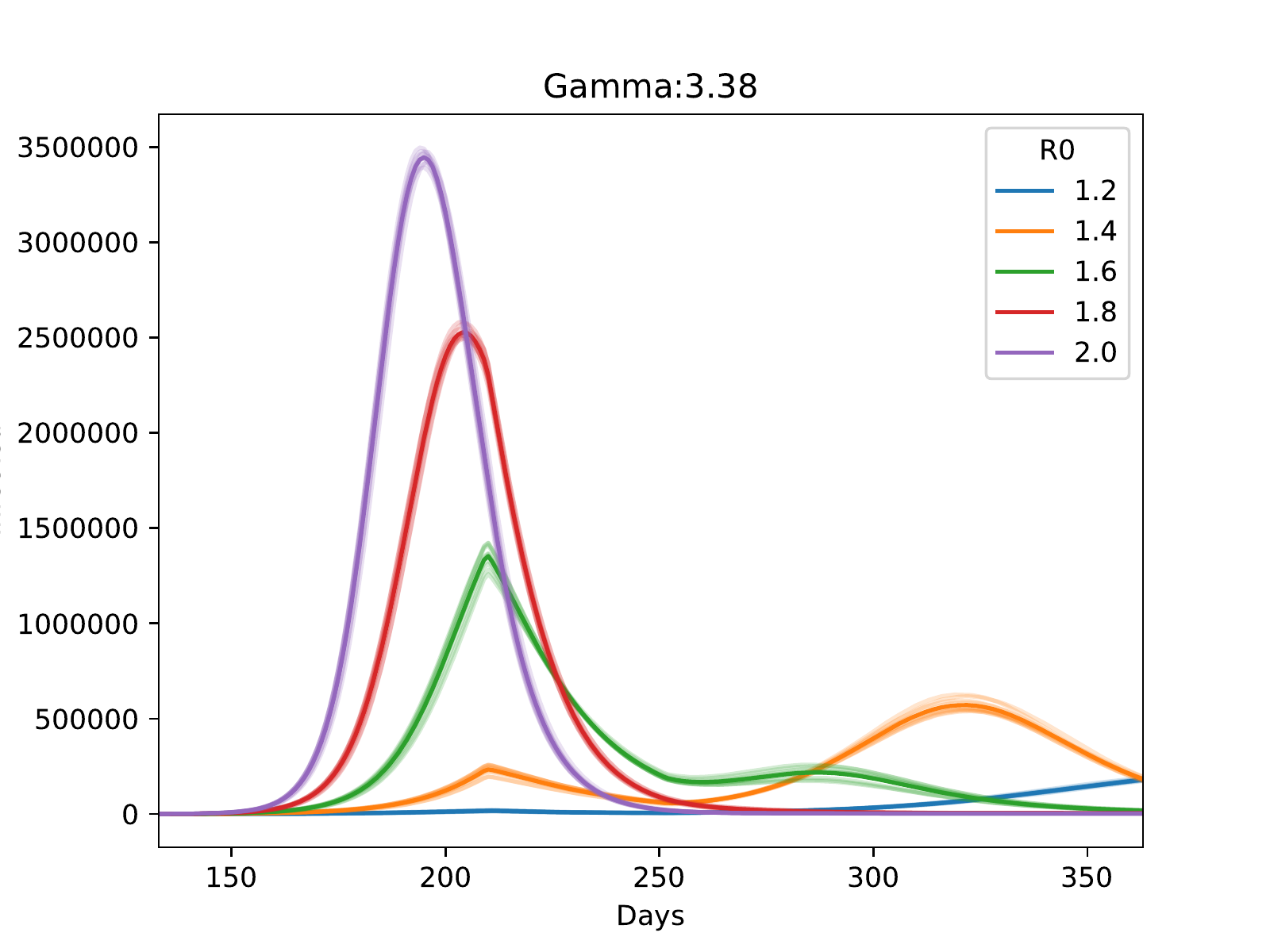}
   \end{minipage}

    \caption{We demonstrate our model for $R_0 \in \{1.2, 1.4,1.6,1.8,2.0\}$ (enumerated in the legend) and an infectious period of $1.8$ days (top panel), $2.5$ days (bottom left panel) and $3.38$ days (bottom right panel). For each parameter combination, we show a set of stochastic trajectories (light coloured lines) and the mean of these trajectories (dark coloured line). For clarity, we only show 10 stochastic trajectories in this Figure. }
      \label{fig:gb_2009_allR0}
\end{figure}

For this experiment, we found,
\begin{equation}
\susceptiblecontribution=\log(R_0) \cdot 2.74,
\end{equation}
to be a good fit for the overall comparison.
In Figure~\ref{fig:gb_2009_allR0} we show the epidemic curve for our model with respect to these parameters.
In general, the epidemic curves that result from using an infectious period of $1.8$ days are insufficient to reproduce the trends of the 2009 pandemic.
For the other infectious periods (i.e., $2.5$ and $3.38$), we show that for all but the highest reproductive numbers we observe an epidemic curve with 2 peaks.
Furthermore, we observe a deeper trough in the epidemic curve when an infectious period of $2.5$ days is chosen.

%cd /Users/plibin/research/spatial-pandemic-experiments/spatial/uk_2009
%./run_all.sh
%
%(or on the Thinking VSC)
%wsub -A lp_hiv_networks -batch run_all.pbs -data run_all.csv
%
%python plot_trajectories.py --R0s "1.2,1.4,1.6,1.8,2.0" --dir ./runs/ --n 10  --gamma 3.38 --proportion_clinical 1 -o gb_2009_allR0-gamma338.pdf
%python plot_trajectories.py --R0s "1.2,1.4,1.6,1.8,2.0" --dir ./runs/ --n 10  --gamma 2.5 --proportion_clinical 1 -o gb_2009_allR0-gamma25.pdf
%python plot_trajectories.py --R0s "1.2,1.4,1.6,1.8,2.0" --dir ./runs/ --n 10  --gamma 1.8 --proportion_clinical 1 -o gb_2009_allR0-gamma18.pdf

In Figure~\ref{fig:gb_2009_bestfit}, we show a set of model realisations in conjunction with the symptomatic case data, which shows that we were able to closely match the epidemic trends observed during the British pandemic in 2009.
This model was configured with a basic reproductive number of $1.4$ and infectious period of $2.6$.
The reproductive number is in good concordance with the general consensus that the virus responsible for the 2009 pandemic exhibited a moderate infectiousness.
While the infectious period slightly differs from the value reported by \cite{balcan2009seasonal} (i.e., $2.5$ days), it lies well within the confidence bounds reported in this study (confidence interval: 1.1-4.0 days).
Note that our model reports the number of infections, while the HPA only recorded symptomatic cases.
Therefore we scale the epidemic curve with a factor of $\frac{1}{4}$.
While atypical, this large number of asymptomatic cases produced by our model is in line with earlier serological surveys \cite{miller2010incidence} and with previous modelling studies \cite{kubiak20122009}.

%cd /Users/plibin/research/spatial-pandemic-experiments/spatial/uk_2009
%./run_all.sh
%
%(or on the Thinking VSC)
%wsub -A lp_hiv_networks -batch run_all.pbs -data run_all.csv
%
%python plot_trajectories.py --R0s "1.4" --dir ./runs/ --n 10  --gamma 2.6 --proportion_clinical .25  --cases_csv /Users/plibin/projects/spatial-pandemic-py/data/2009-pandemic-cases/cases-2009-extracted-from-kubiak-2012.csv -o gb_2009_bestfit.pdf
\begin{figure}
  \centering
   \includegraphics[width=.7\textwidth]{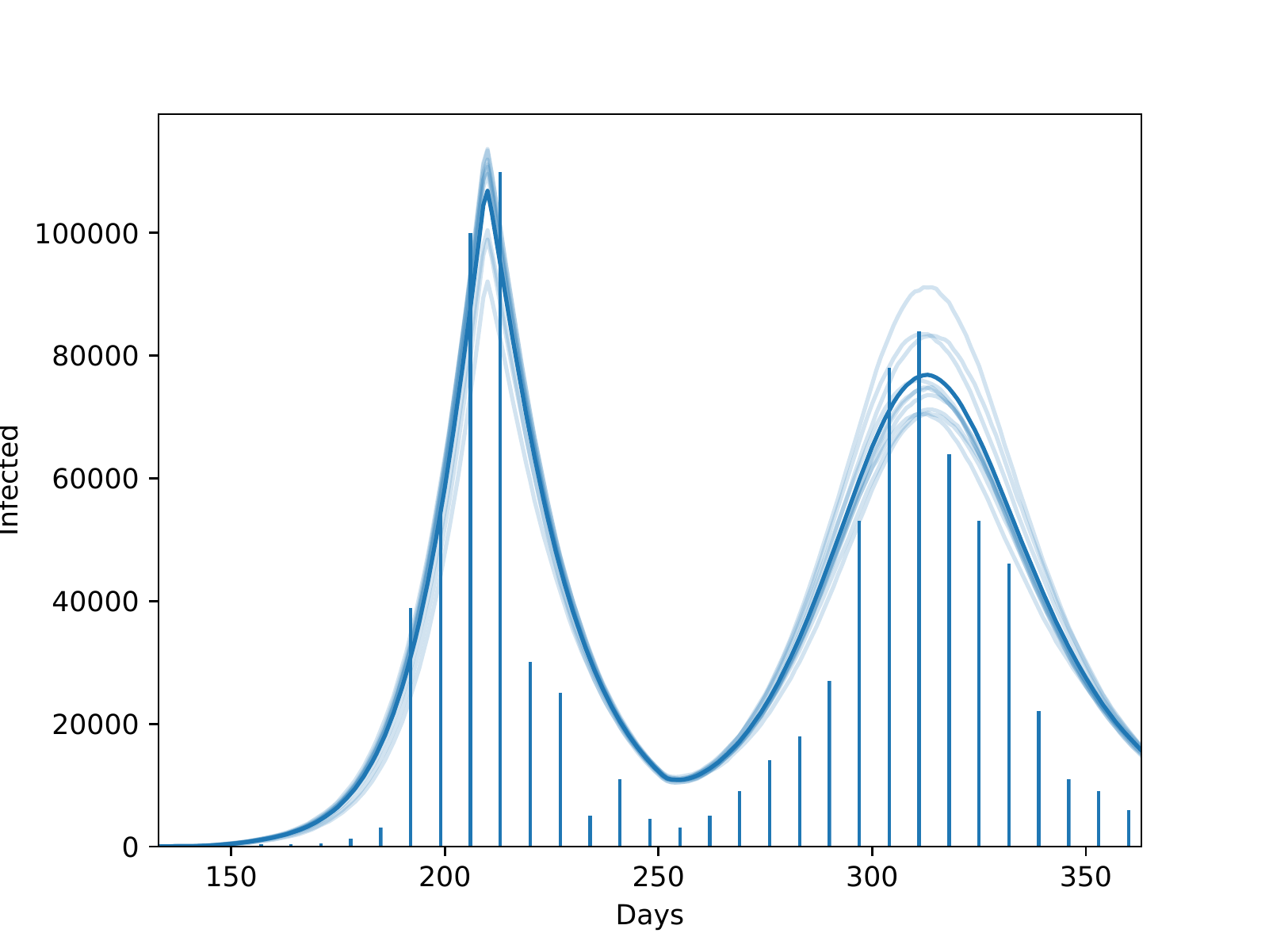}

    \caption{We show that our model, using a reproductive number of $1.4$ and a an average duration of infectiousness of $2.6$ days is able to match the trends observed in the British pandemic of 2009.  For clarity, we only show 10 stochastic trajectories in this Figure.}
      \label{fig:gb_2009_bestfit}
\end{figure}

\section{Computational complexity and performance}
\label{sec:computational_complexity}
An analysis of the computational complexity of our model needs to consider that the model incorporates two components.
On the one hand, infection in the patches is triggered via the time scale transformation algorithm (see Section~\ref{sec:time_scale_transformation_epi}).
On the other hand, once infected, each patch in the system evolves independently, and we use the Euler-Maruyama approximation method to obtain samples from the stochastic differential equation that is associated with the patch.
The time scale transformation algorithm samples a random threshold for each patch, which is compared to the force of infection of the associated patch.
This comparison occurs at each time step that the patch was not yet infected.
Computing the force of infection in Equation~\ref{eq:between_force_of_infection} considers all model patches, and thus has a worst case complexity that is linear in the number of model patches, i.e., 
\begin{equation}
\mathcal{O}(|\patches|).
\end{equation}
In worst case, at each time step $t$, if only one of the model patches is infected, and we need to compute the force of infection for each patch, which has a quadratic complexity in the number of model patches, i.e., 
\begin{equation}
\mathcal{O}(t \cdot |\patches|^2).
\label{eq:complexity_1}
\end{equation}

However, when each patch only needs to be infected once, we observe that we have a complexity in terms of infected $\patches_i$ and uninfected patches $\patches_{\neg i}$.
After all, at each time step, we only need to consider the force of infection of the uninfected patches, and this force of infection only takes into account the infected patches, i.e.,
\begin{equation}
\mathcal{O}(t \cdot |\patches_i \cdot \patches_{\neg i}|).
\label{eq:complexity_2}
\end{equation}
Since we have,
\begin{equation}
\patches = \patches_i + \patches_{\neg i},
\end{equation}
we can see that while this expression has the same worst case complexity as in Equation~\ref{eq:complexity_1}, in practice less operations will be required.

For both the complexity in Equation~\ref{eq:complexity_1} and Equation~\ref{eq:complexity_2}, it is clear that as long as the number of patches is limited, as is the case in our model, this procedure will be computationally efficient, as this can be implemented as a vector product, on vectors that all fit in memory (RAM).

When a patch is infected, at each time step it will be advanced by using a number of operations that is proportional to the number of compartments in the the age-dependent SEIR model.

This model was implemented in Python, and the performance critical sections were either implemented using NumPy when a vector representation was possible (e.g., to compute the force of infection) \cite{numpy2006}, or JIT-compiled using Numba (e.g., to evolve the  age-dependent SEIR model in a patch) \cite{lam2015numba}. 
This implementation performs well, resulting in about $2$ simulation runs per second on a MacBook Pro.

\section{Selecting districts to establish a ground truth}
\begin{figure}\centering
\zoombarycentric{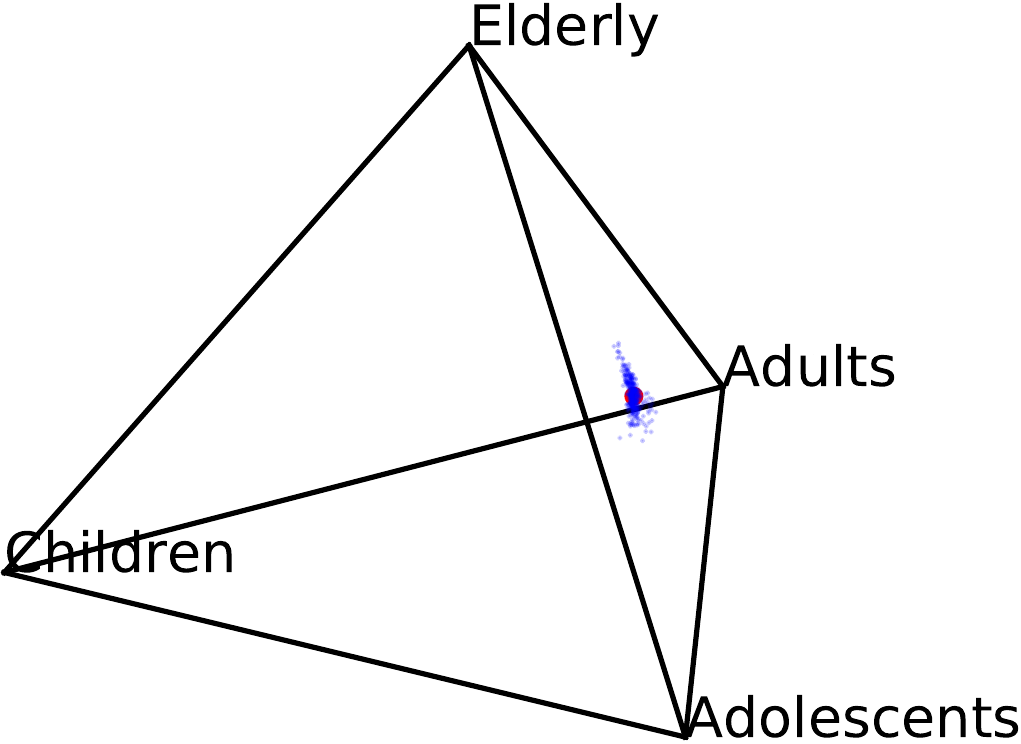}
\caption{Barycentric projection of the census proportions in the districts of Great Britain (blue scatter points), according to Eames' age structure. The geometric mean of this distribution is shown as a red scatter point. The left panel shows the original census pyramid, and the right panel zooms in on the point cloud.}
\label{fig:gb_census_eames_compositional_mean}
\end{figure}

To establish a ground truth, we select 10 districts that are representative of the population heterogeneity in Great Britain.
To this end, we remind the reader that in Section~\ref{sec:census}, we analysed the population heterogeneity by representing the population structure as a positive simplex.
We select 10 districts: one district that is representative for the average of this distribution and a set of nine districts that is representative for the diversity in this distribution.
To determine the average district, we consider the population heterogeneity distribution over all districts, and determine the Aitchison's mean (Definition~\ref{def:aitchison_mean}) of this distribution \cite{aitchison1994principles}.
We then select the district that is closest to the Aitchison's mean (Definition~\ref{def:aitchison_mean}) according to the Aitchison distance (Definition~\ref{def:aitchison_distance}), as shown in Figure \ref{fig:gb_census_eames_compositional_mean}.

\begin{definition}[Aitchison's mean]
Given a set of points from a unit simplex (Definition~\ref{eq:positive_simplex}),
\begin{equation}
P = \{p^{(i)} \mid \ p^{(i)} \in \simplexset^D \}^N_{i=1},
\end{equation}
the Aitchison's mean \cite{aitchison1997one} is:
\begin{equation}
\aitchisonmean(P) = \frac{\langle h_1,...,h_D\rangle}{\sum_{d=1}^D h_d},
\end{equation}
where,
\begin{equation}
h_d = \left( \prod_{p^{(i)} \in P} p^{(i)}_{d} \right)^{(1/N)},
\end{equation}
is the geometric mean of the $d$-th component over all simplex points in $P$.
\label{def:aitchison_mean}
\end{definition}

\begin{definition}[Aitchison distance]
Given two points from a unit simplex $p,q \in \simplexset^D$  (Definition~\ref{eq:positive_simplex}), we define the Aitchison distance function \cite{aitchison1992criteria}:
\begin{equation}
d_A(p, q) = \left[\sum_{d=1}^{D} \left( \log \frac{p_d}{h(p)} - \log \frac{q_d}{h(q)}  \right)  \right]^{1/2},
\end{equation}
where,
\begin{equation}
h(p) = \left(\prod^D_{d=1} p_d \right)^{(1/D)},
\end{equation}
denotes the geometric mean of $p$.
This distance defines a metric on the simplex sample space.
\label{def:aitchison_distance}
\end{definition}

%DOLATER: known why this distance is best suited for compositional data: In mathematical terms the distance defines a metric on the simplex sample space and has all the necessary properties of scale invariance, permutation invariance, perturbation invariance, and subcompositional dominance as set out in Aitchison (1992) required for applications in compositional data analysis.

%cd ~/research/phd-thesis/latex/figures/spatial-rl/compositional-census
%python plot_census.py --eames_census_fn ~/projects/uk-districts/census/2011/gb.census.eames.csv --show_mean --output gb_census_eames_compositional_mean.pdf
%pdfcrop gb_census_eames_compositional_mean.pdf

\begin{figure}
  \centering
   \zoombarycentric{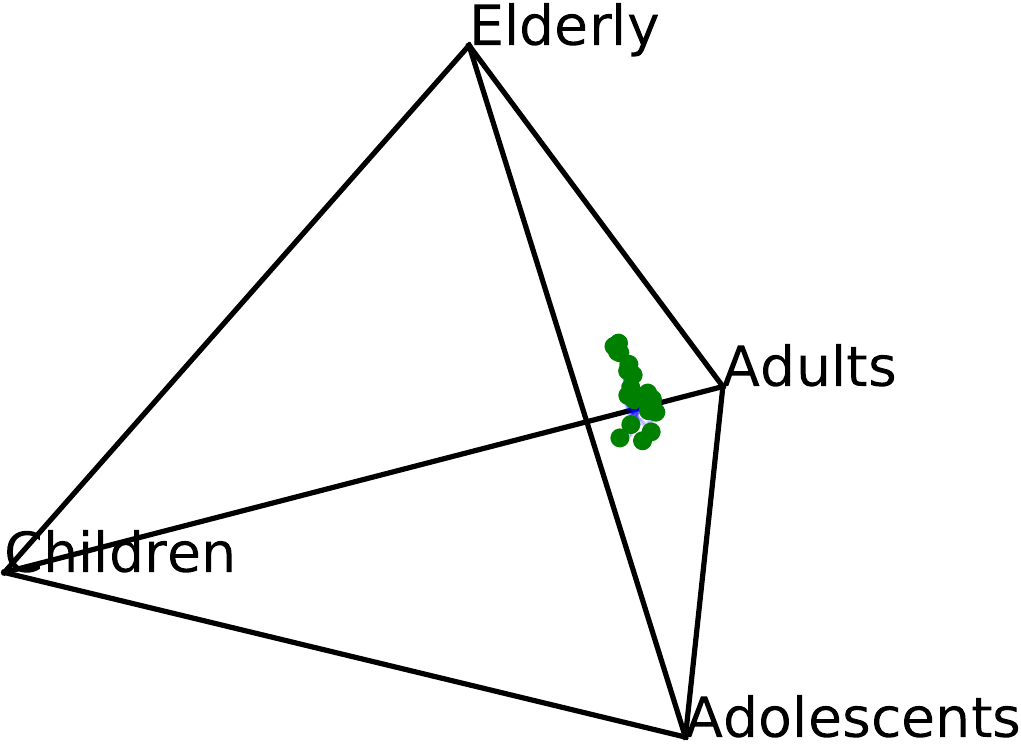}
   \caption{Barycentric projection of the census proportions in the districts of Great Britain (blue scatter points), according to Eames' age structure. The census points that are part of the convex hull are shown in green. The left panel shows the original census pyramid, and the right panel zooms in on the point cloud.}
   \label{fig:gb_census_eames_convex_hull}
\end{figure}

Next, we determine the outer extreme points, as these represent the most diverse census points.
To do this, we compute the convex hull of the point cloud (i.e., the smallest convex set of points that contains the point cloud), as shown in Figure \ref{fig:gb_census_eames_convex_hull}.

%cd ~/research/phd-thesis/latex/figures/spatial-rl/compositional-census
%python plot_census.py --eames_census_fn ~/projects/uk-districts/census/2011/gb.census.eames.csv --show_hull --output gb_census_eames_convex_hull-crop.pdf
%pdfcrop gb_census_eames_convex_hull-crop.pdf

\begin{figure}
  \centering
   \zoombarycentric{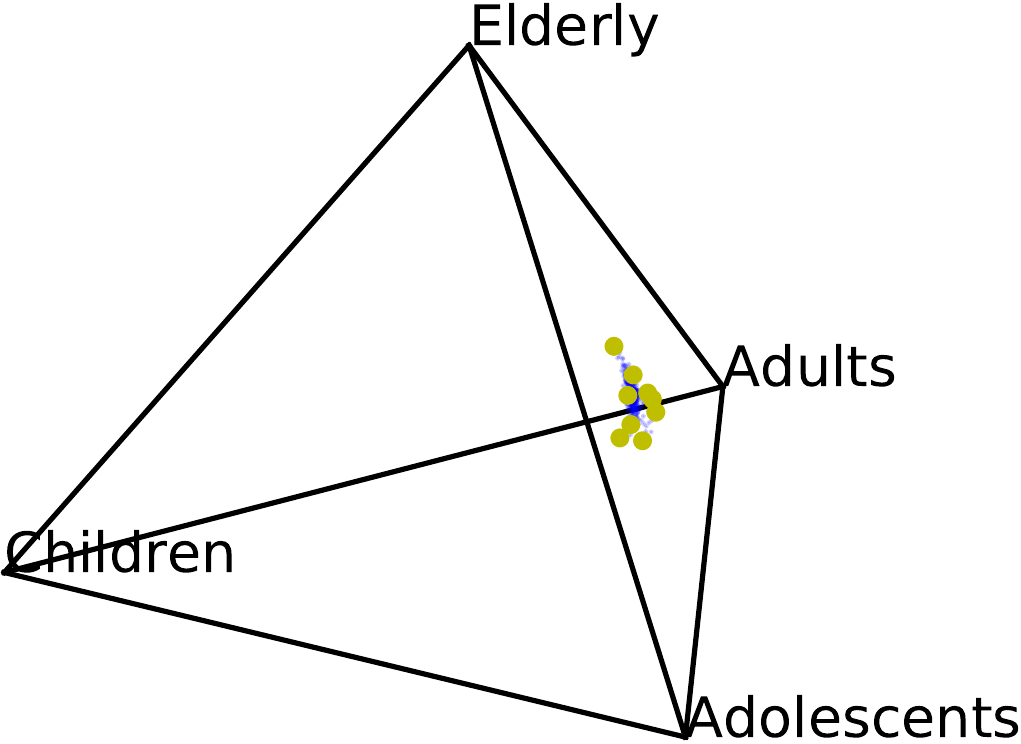}
   \caption{Barycentric projection of the census proportions in the districts of Great Britain (blue scatter points), according to Eames' age structure. The census point that were selected out of the convex hull are shown in yellow. The left panel shows the original census pyramid, and the right panel zooms in on the point cloud.}
   \label{fig:gb_census_eames_selection}
\end{figure}

We proceed by taking the points that belong to the surface of the convex hull, of which we make a sub-selection of 9 census points.
As the convex hull consists out of 21 points, we consider all k-combinations (with $k=9$) and select the set of points that maximizes the minimum Aitchison distance between the selected points, as shown in Figure~\ref{fig:gb_census_eames_selection}.

%cd ~/research/phd-thesis/latex/figures/spatial-rl/compositional-census
%python plot_census.py --eames_census_fn ~/projects/uk-districts/census/2011/gb.census.eames.csv --show_selection-k 9 --output gb_census_eames_selection-crop.pdf
%pdfcrop gb_census_eames_selection.pdf

\section{Finding communities}
\label{sec:mobility_network}
To investigate the collaborative nature of school closure policies, we apply deep multi-agent reinforcement learning algorithms.
In our model, we have \nrgbdistricts{} agents, one for each district, as agents represent the district for which they can control school closure.
As the current state-of-the-art of deep multi-agent reinforcement learning algorithms is limited to deal with about $10$ agents \cite{hernandez2019survey}, we thus need to partition our model into smaller groups of agents, such that deep multi-agent reinforcement learning algorithms become feasible.

To this end, we consider the mobility matrix $\mobilitymatrix$ and define a directed commute graph for $\mobilitymatrix_{ij} \geq 0$ (Definition~\ref{def:commute_graph}).

\begin{definition}[Commute graph]
For a commuting matrix $\mobilitymatrix$ that describes the mobility flux between a set of districts $\mathcal{D}$, we define a commute graph,
\begin{equation}
G_c = \langle V_c, A_c \rangle,
\end{equation}
where $V_c$ is the set of vertices, with a vertex for each of the districts in $\mathcal{D}$, and $A_c$ is the adjacency matrix that specifies the vertices that are connected:
\begin{equation}
(A_c)_{ij} = \begin{cases}
	1 ,& \mobilitymatrix_{ij} > 0\\
	0 ,& \mobilitymatrix_{ij} = 0 \\
\end{cases} 
\end{equation}
Each pair of connected vertices $i$ and $j$ has a weight $\mobilitymatrix_{ij}$.
\label{def:commute_graph}
\end{definition}

\begin{figure}[H]
  \centering
   \includegraphics[width=.5\linewidth]{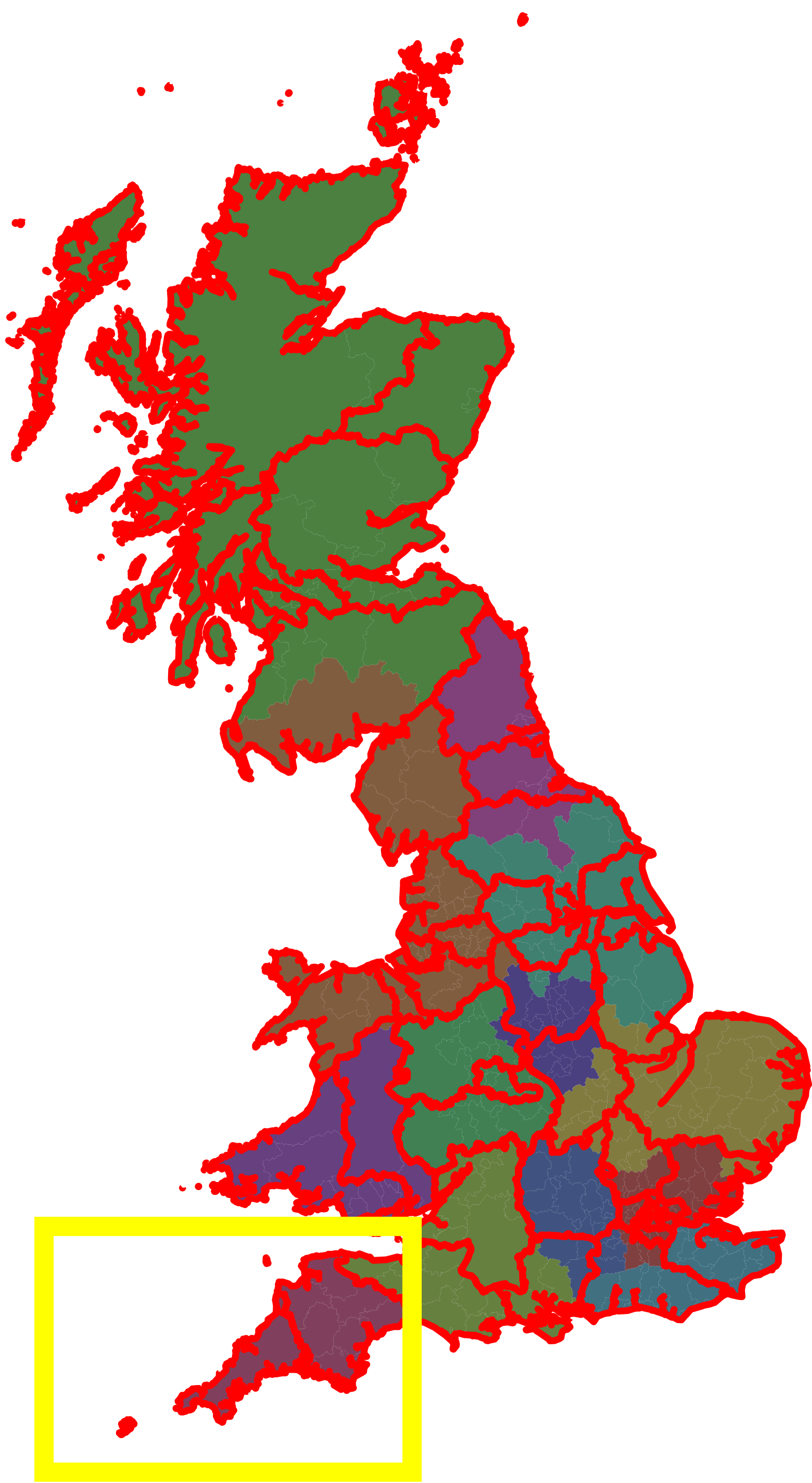}
      \caption{We show the communities, that resulted from applying the Leiden algorithm, on the map of Great Britain. We show all administrative districts colour-coded by the community they belong to and the add the borders of the NUTS-2 administrative regions on top of this map. We annotate the Cornwall-Devon community with a yellow rectangle.}
      \label{fig:partition_on_map}
\end{figure}

To detect communities in the commute graph, we used the Leiden algorithm \cite{traag2019louvain}, an algorithm that searches for communities that maximize the network modularity \cite{leicht2008community}.
We found a partition of which we demonstrated the robustness ($p$-value $\leq{0.001}$) using a bootstrapping approach presented in \cite{radivojevic2017community}.
Furthermore, by rendering this partition on top of the map of Great Britain, as is shown in Figure~\ref{fig:partition_on_map}, we show that the districts belonging to the same community are close to each other geographically, as we would expect.
Moreover, when we overlay the NUTS-2 administrative regions\footnote{NUTS (Nomenclature of Territorial Units for Statistics) is a geocode standard constructed by Eurostat to reference the subdivisions of European countries. NUTS-2 is the second level and corresponds to basic regions for the application of regional policies.} on the partitioning (Figure~\ref{fig:partition_on_map}), we observe that our partitioning scheme mostly overlaps with the NUTS-2 regions, which indicates that the Leiden algorithm produces a sensible partitioning.

%cd ~/research/spatial-pandemic-experiments/flux
%python leiden.py --method reichardt-modularity --seed 1 --n_iter 100 --output partition.txt
%python plot_partition_on_map.py --partition ./partition.txt -o partition_on_map.pdf
%python plot_partition_on_map.py --partition ./partition.txt -o partition_on_map_nuts.pdf --nuts
%pdfcrop partition_on_map.pdf
%pdfcrop partition_on_map_nuts.pdf

We conduct our multi-agent reinforcement learning experiments in the community with 11 districts, to which we will refer as the Cornwall-Devon community (see Figure~\ref{fig:partition_on_map}), as it is comprised of the Cornwall and Devon NUTS-2 regions.
%cd /Users/plibin/research/spatial-pandemic-experiments/rl-chapter/spatial/census-experiment/
%generate figures:
%./plot_ppo_reward.sh 18
%latex
%python summarize_ar_ppo_curves_latex.py --R0 1.8 --outcome ar
\section{PPO learning curves ($R_0=1.8$)}
\label{sec:apx_ppo_vs_ground_truth_learning_curves_18}
\begin{figure}[H]
	\begin{minipage}{.8\linewidth}
	\centering
	\includegraphics[width=.45\textwidth]{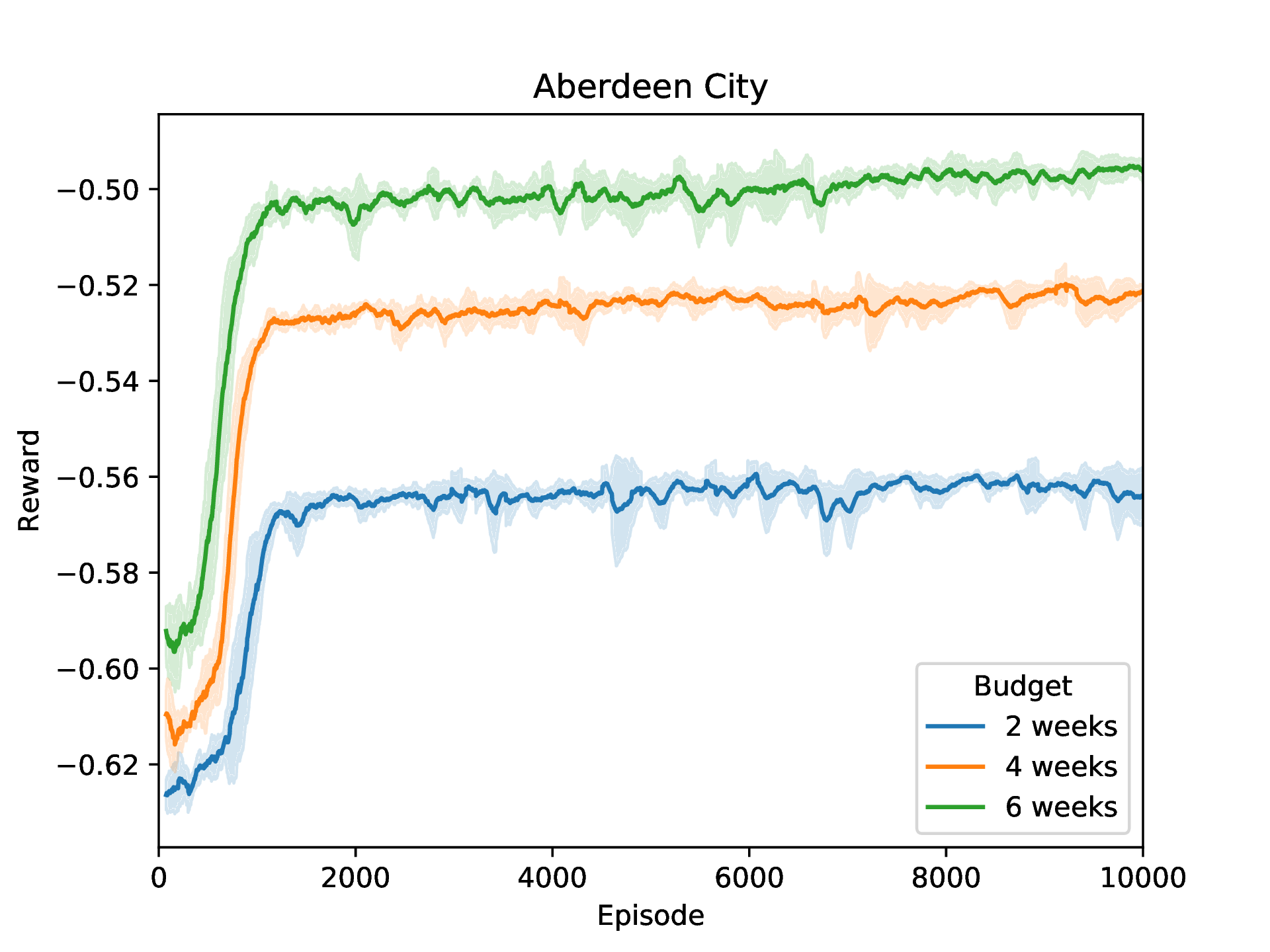}
	\includegraphics[width=.45\textwidth]{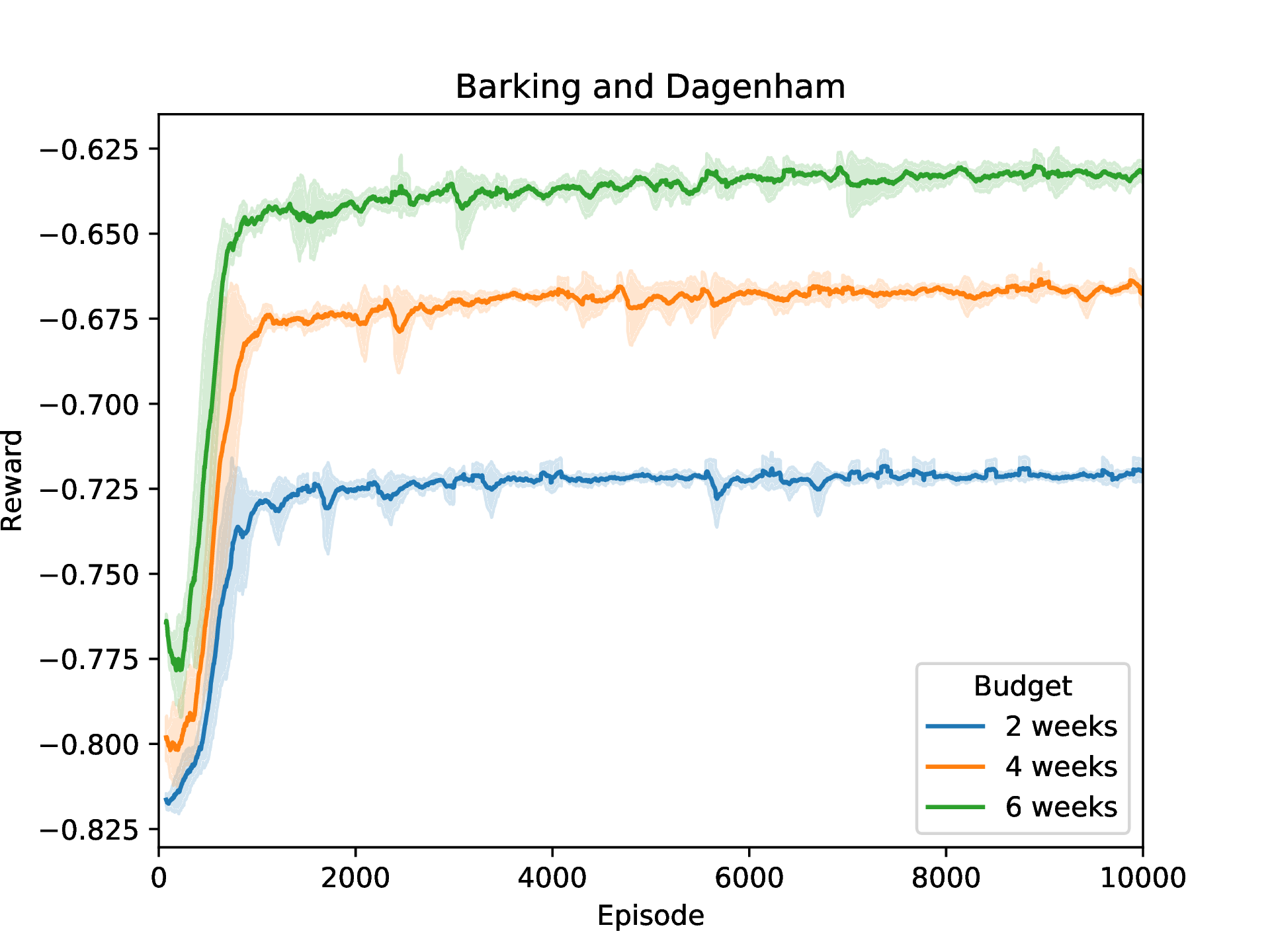}
	\end{minipage}
	\begin{minipage}{.8\linewidth}
	\centering
	\includegraphics[width=.45\textwidth]{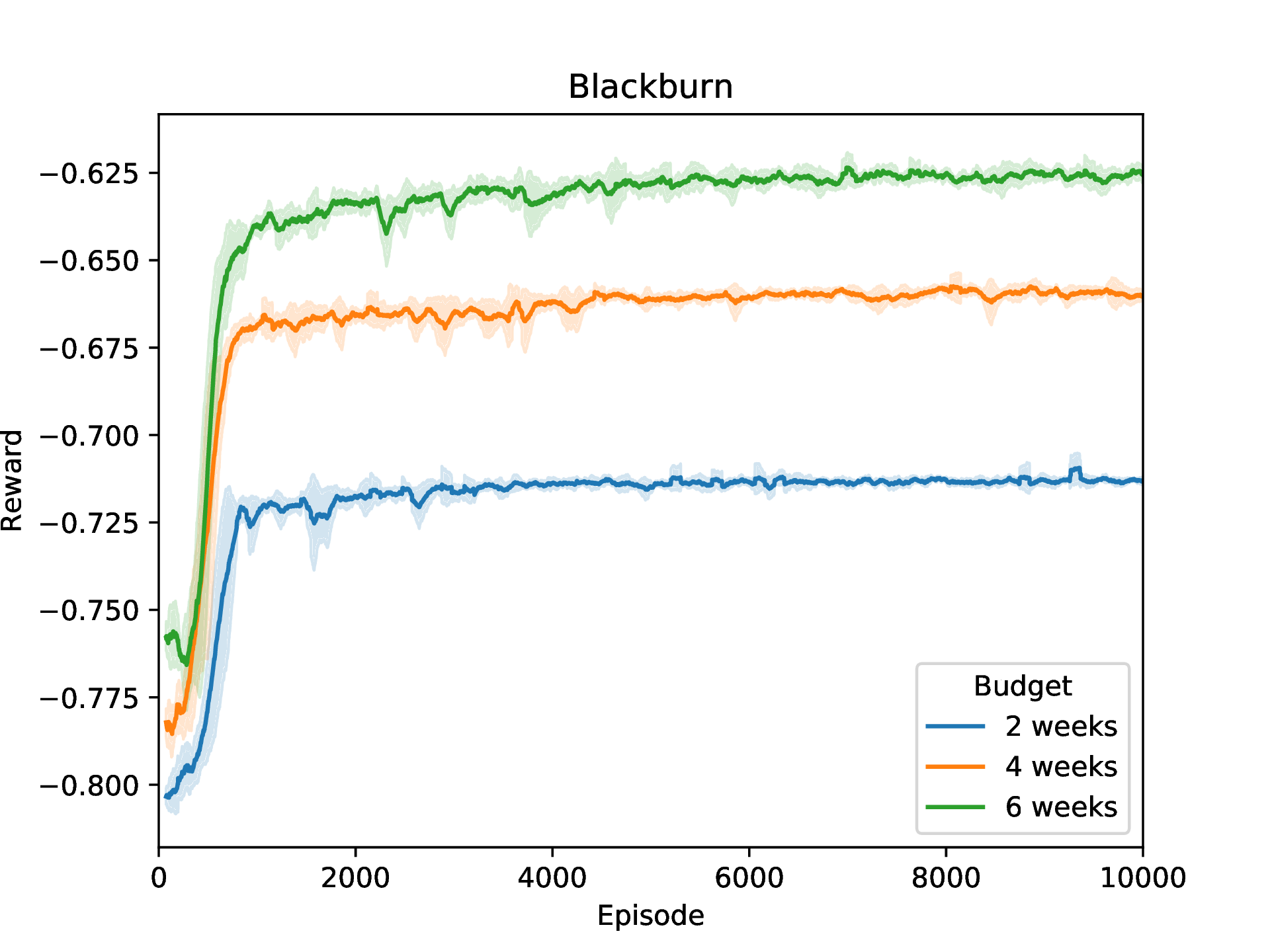}
	\includegraphics[width=.45\textwidth]{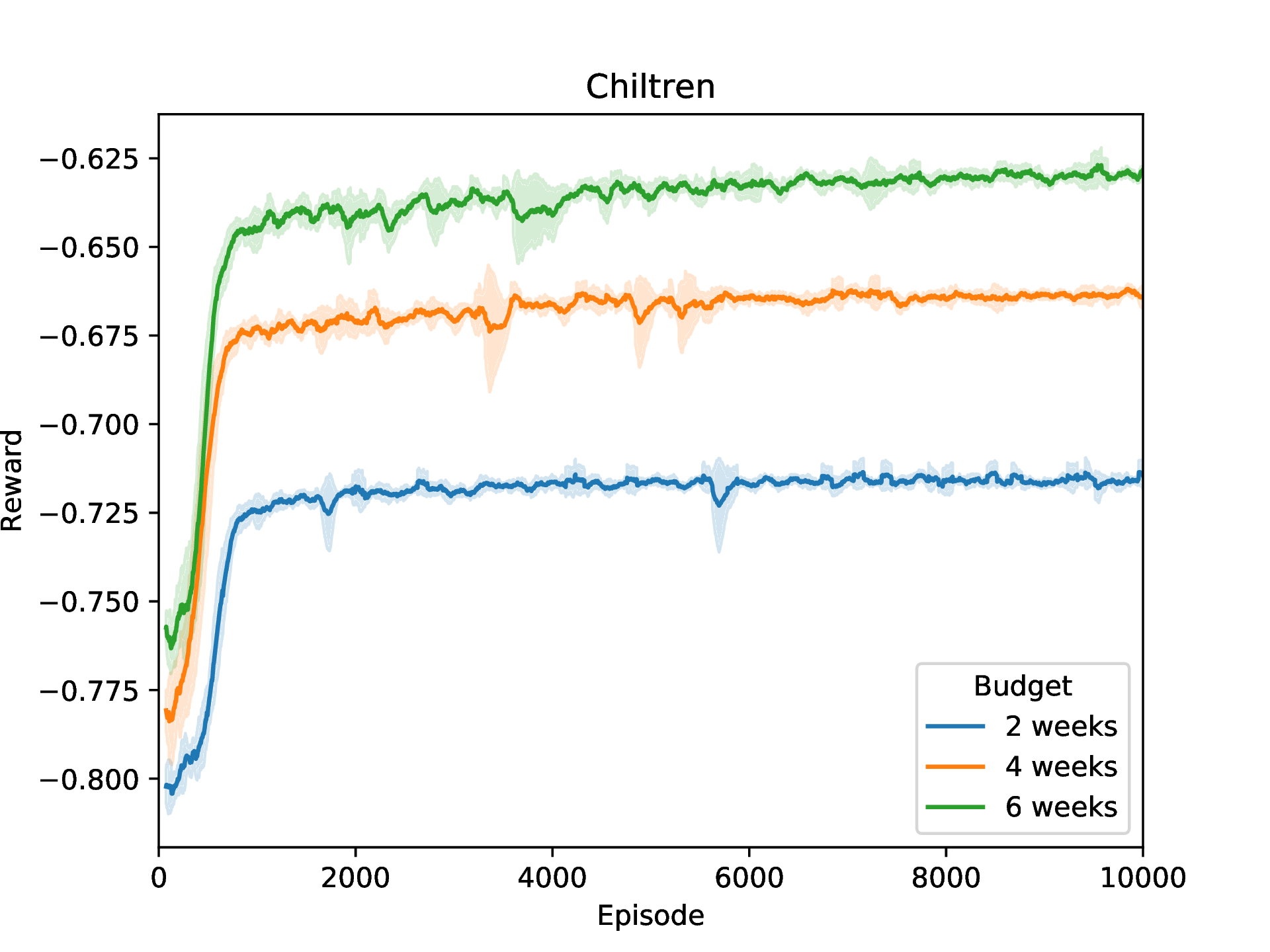}
	\end{minipage}
	\begin{minipage}{.8\linewidth}
	\centering
	\includegraphics[width=.45\textwidth]{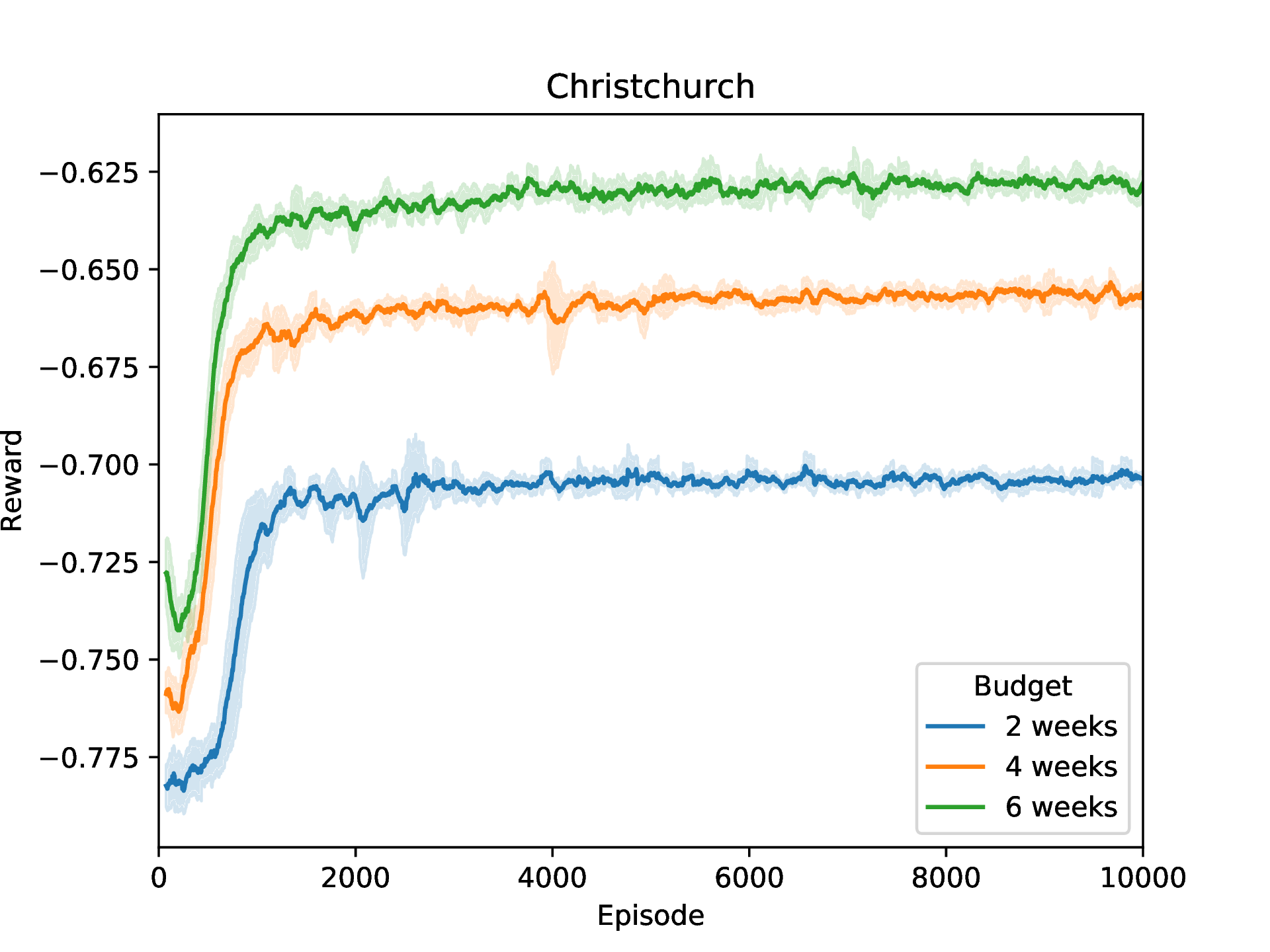}
	\includegraphics[width=.45\textwidth]{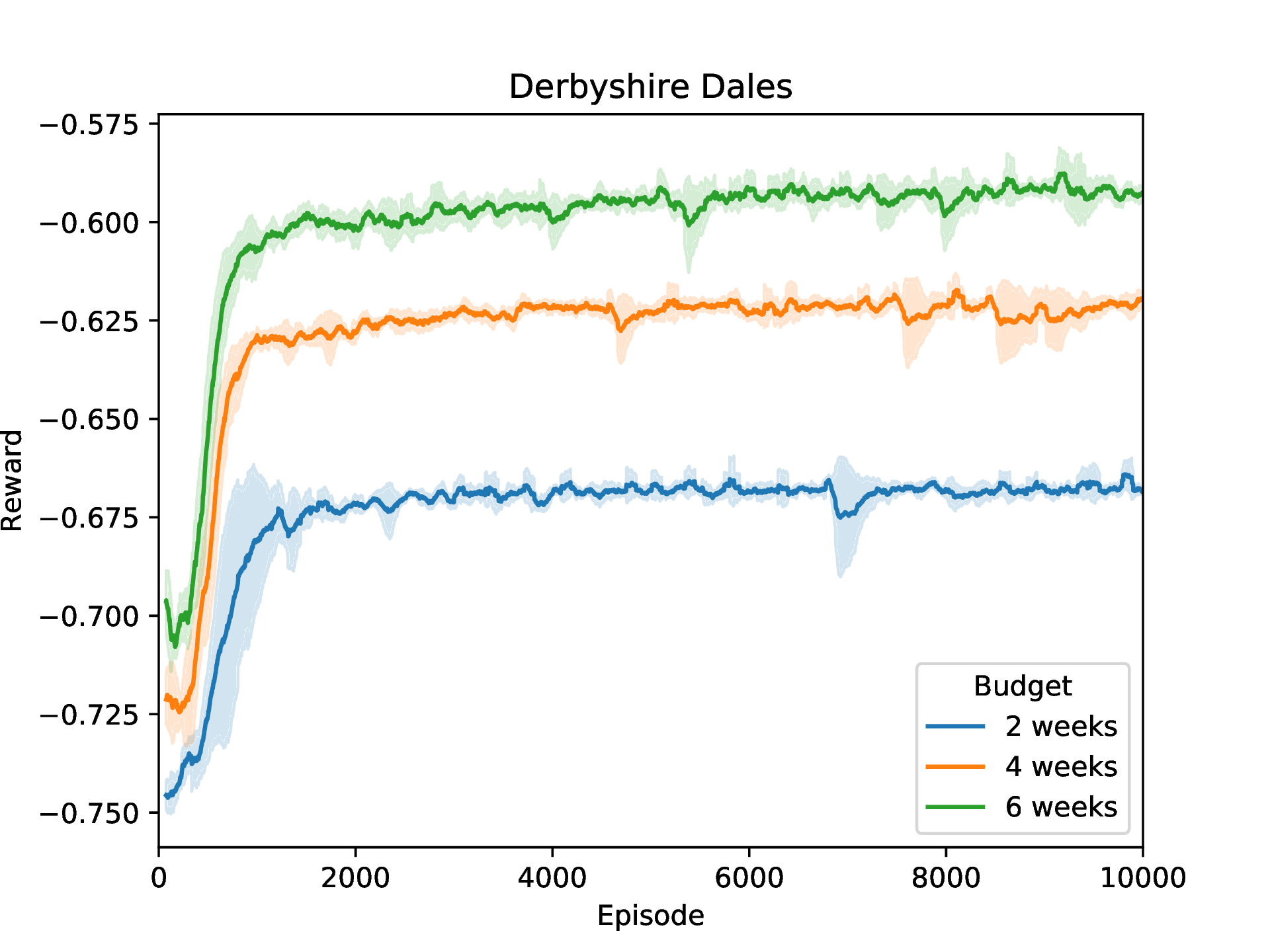}
	\end{minipage}
	\begin{minipage}{.8\linewidth}
	\centering
	\includegraphics[width=.45\textwidth]{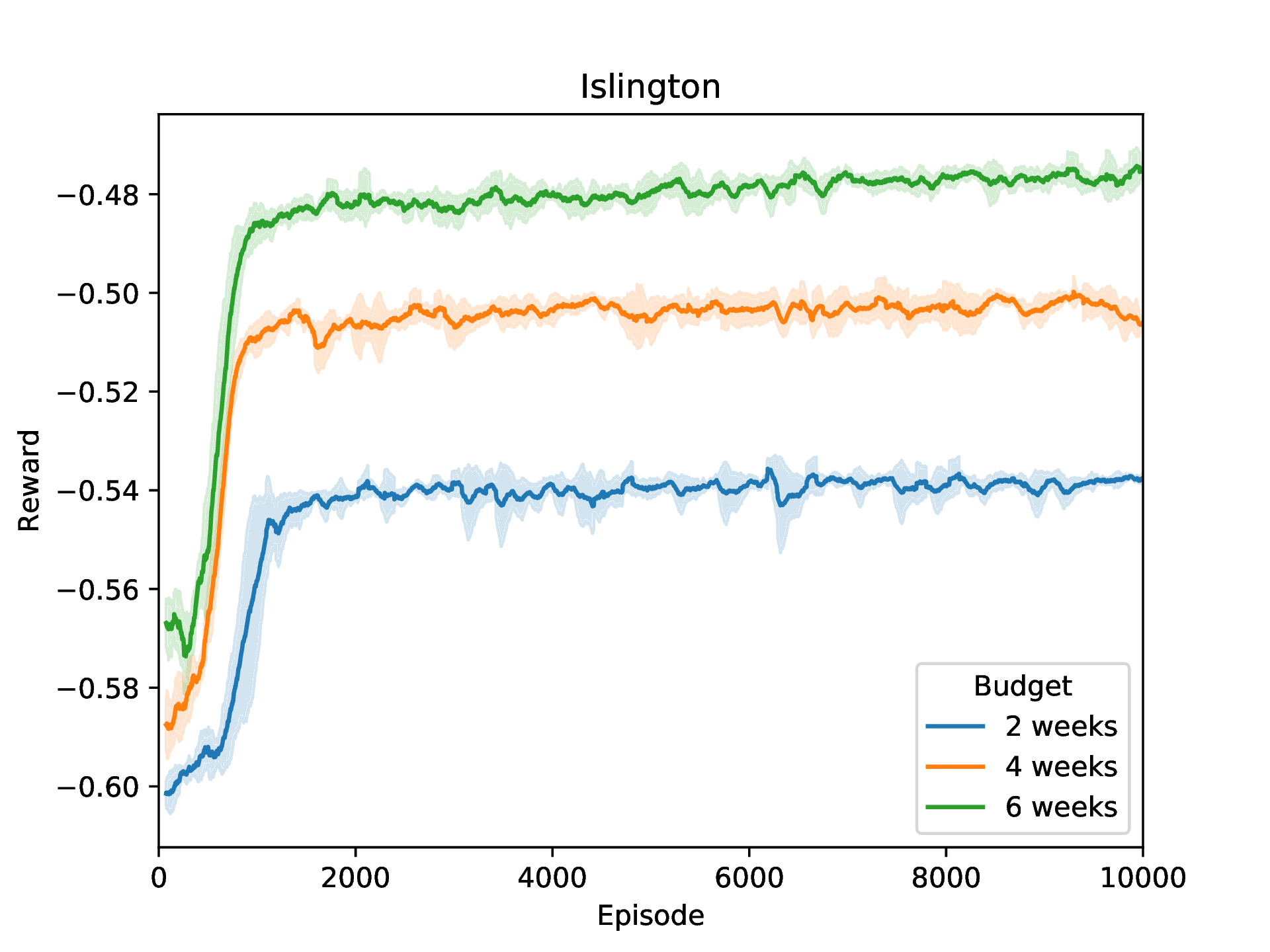}
	\includegraphics[width=.45\textwidth]{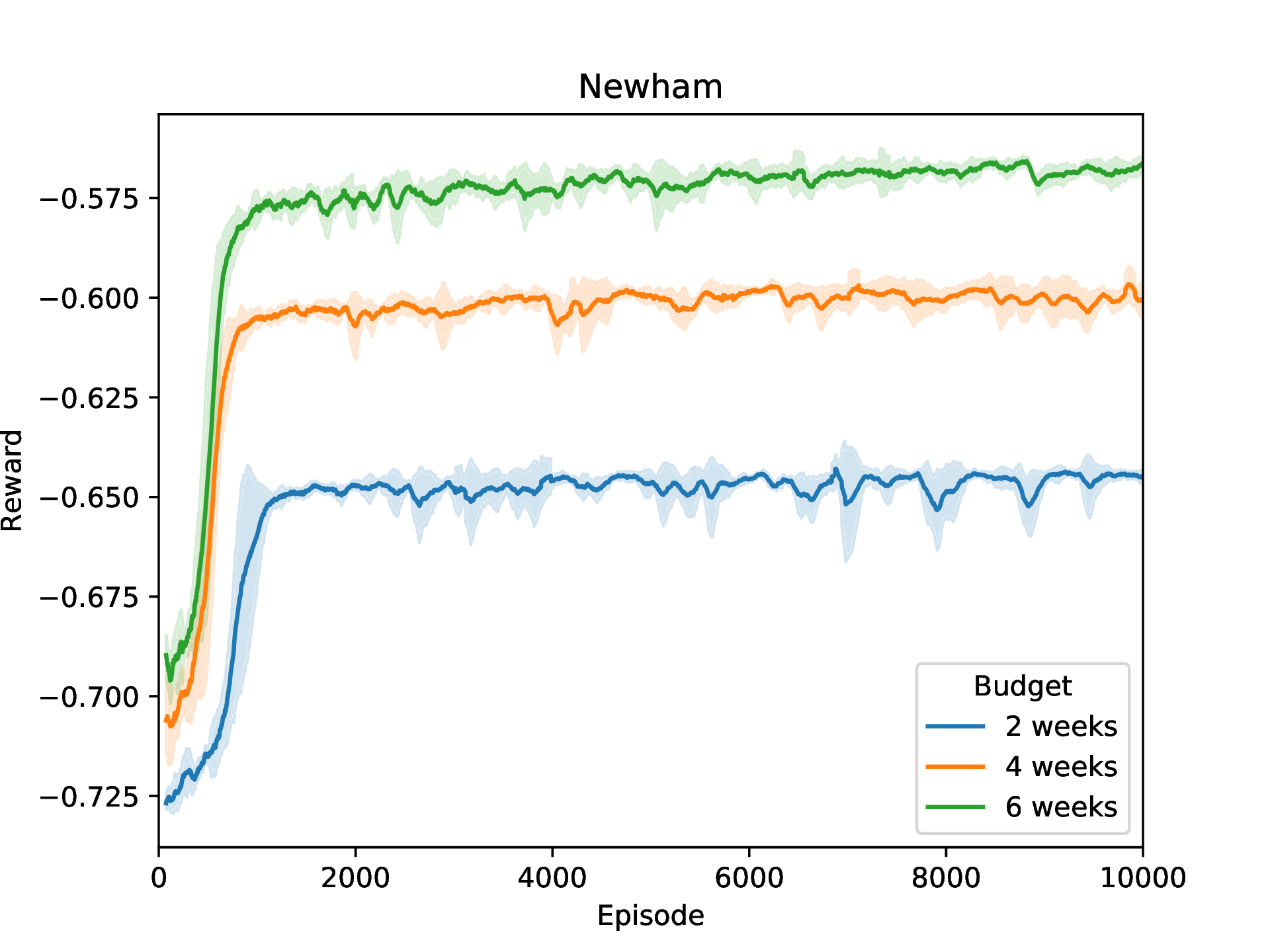}
	\end{minipage}
	\begin{minipage}{.8\linewidth}
	\centering
	\includegraphics[width=.45\textwidth]{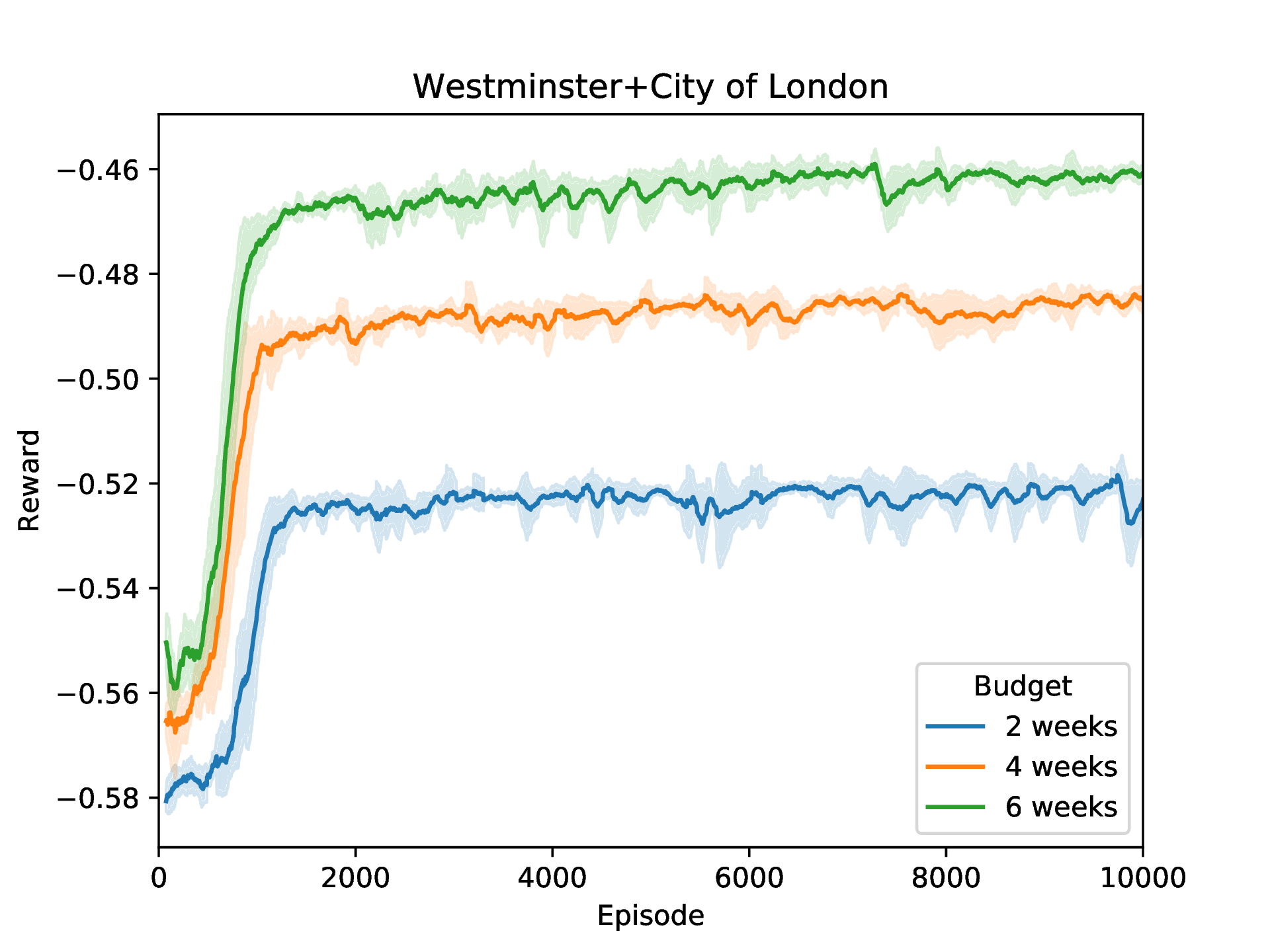}
	\includegraphics[width=.45\textwidth]{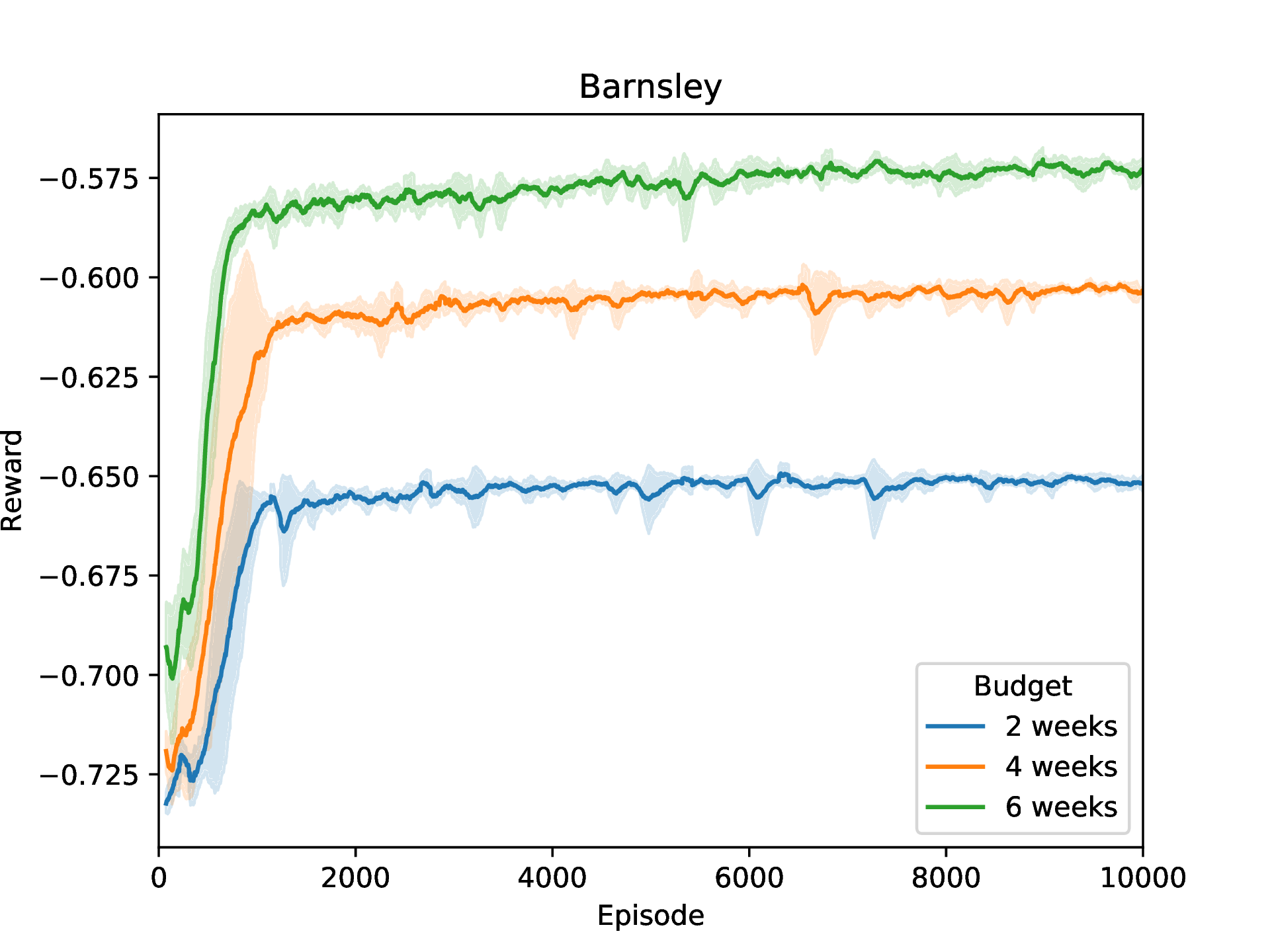}
	\end{minipage}
\caption{PPO learning curves for $R_0=1.8$.}
\end{figure}
%cd /Users/plibin/research/spatial-pandemic-experiments/rl-chapter/spatial/census-experiment/
%generate figures:
%./plot_ppo_reward.sh 24
%latex
%python summarize_ar_ppo_curves_latex.py --R0 2.4 --outcome ar
\section{PPO learning curves ($R_0=2.4$)}
\label{sec:apx_ppo_vs_ground_truth_learning_curves_24}
\begin{figure}[H]
	\begin{minipage}{0.8\textwidth}
	\centering
	\includegraphics[width=.45\textwidth]{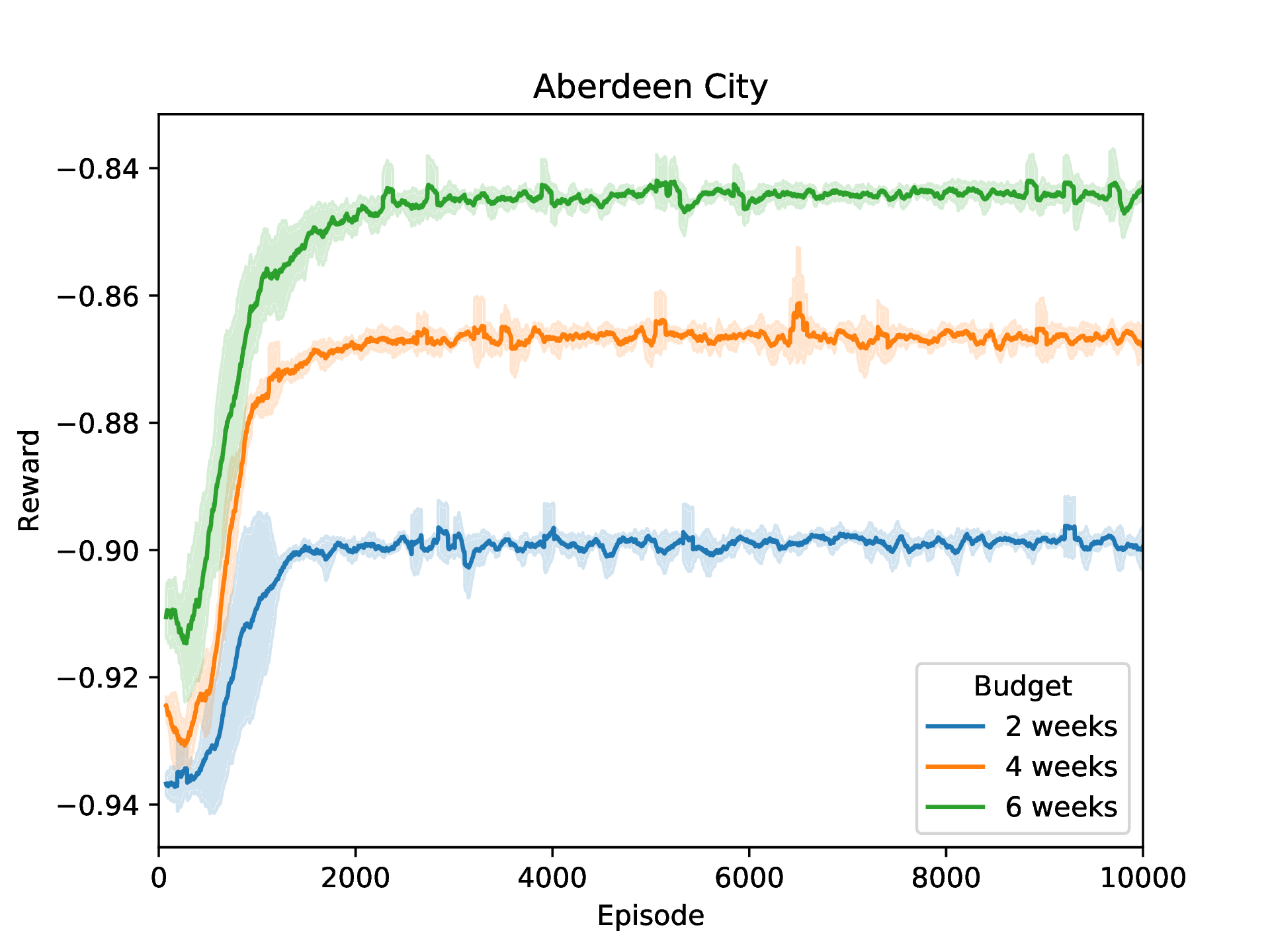}
	\includegraphics[width=.45\textwidth]{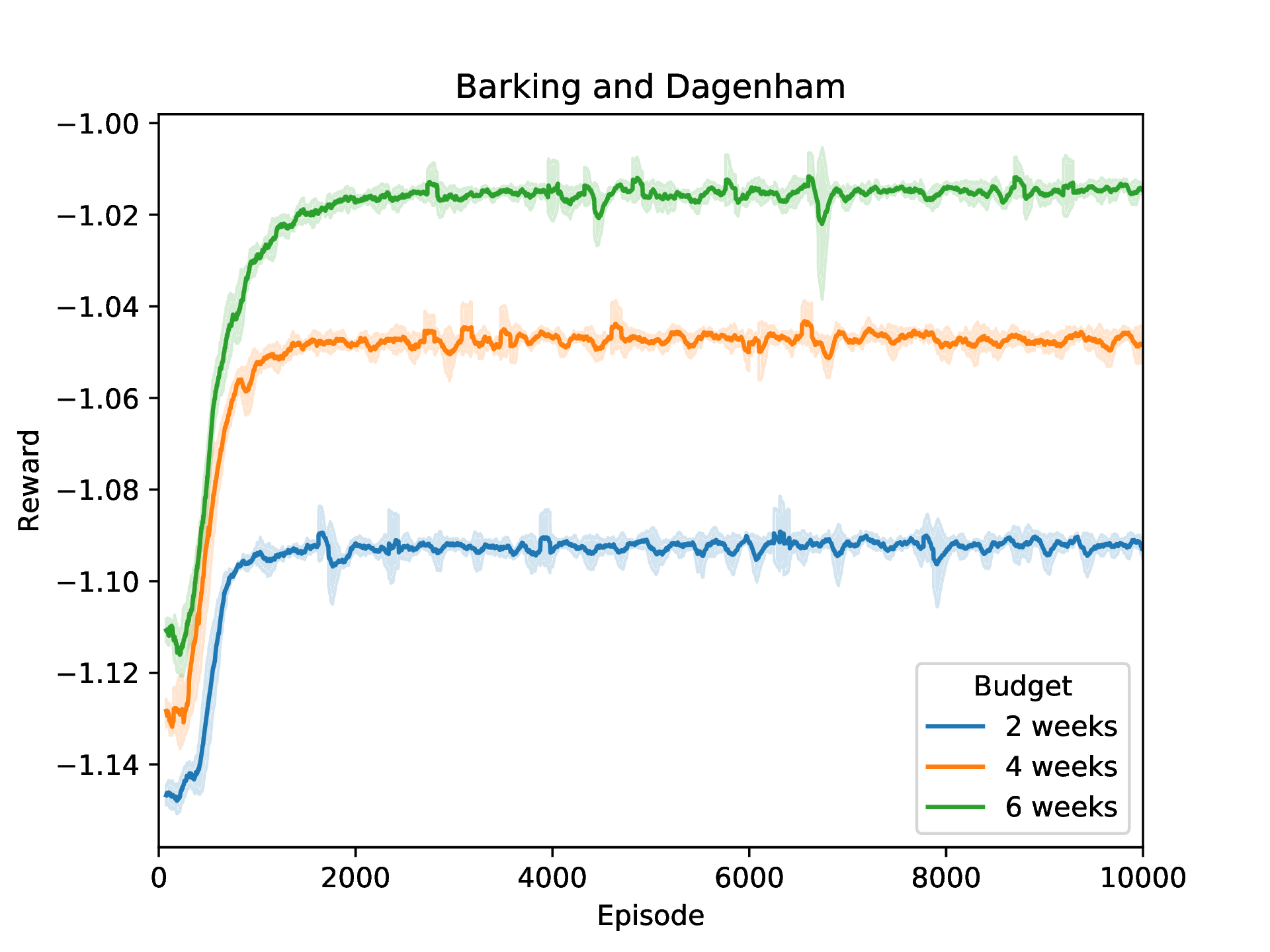}
	\end{minipage}
	\begin{minipage}{0.8\textwidth}
	\centering
	\includegraphics[width=.45\textwidth]{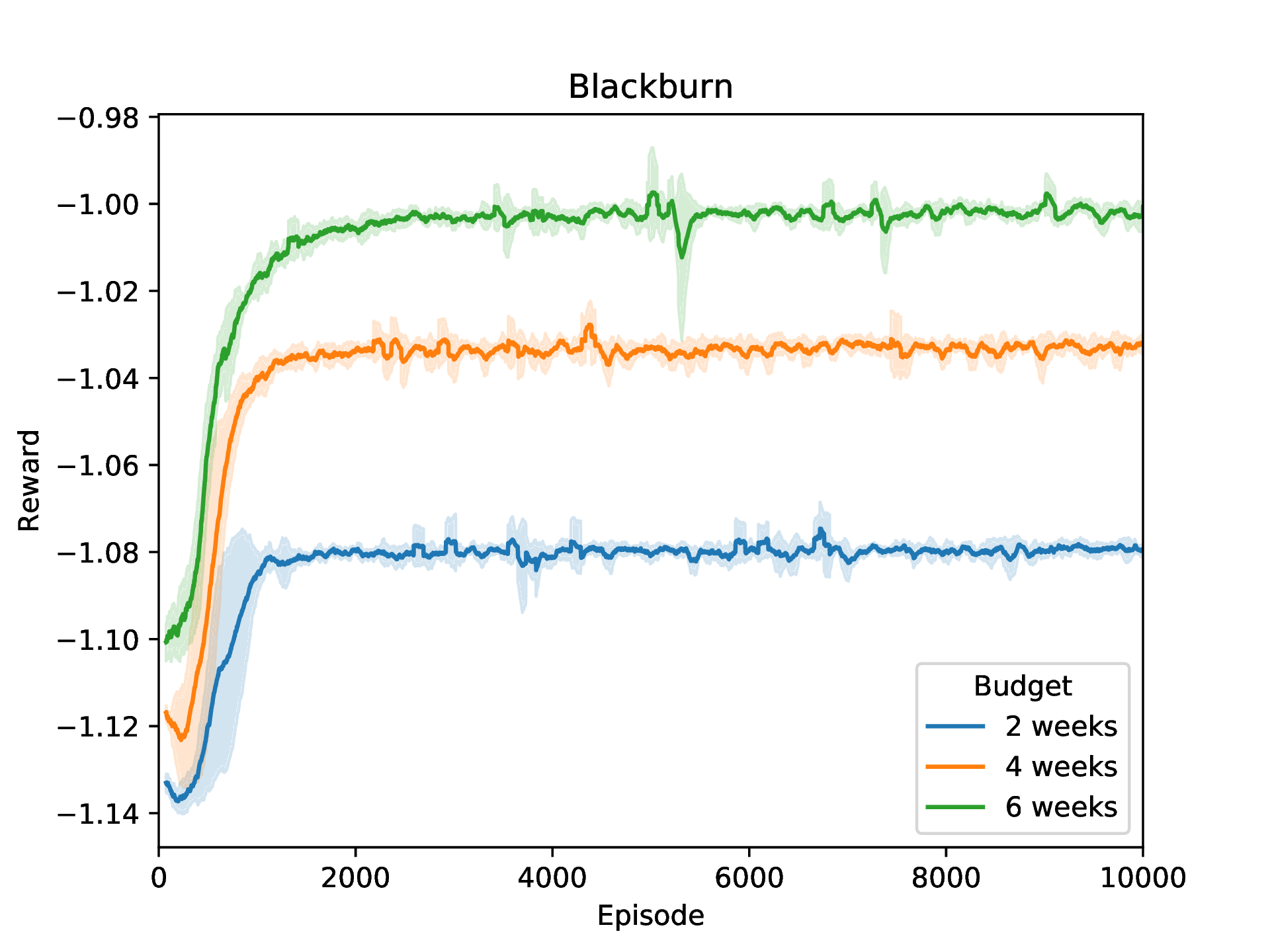}
	\includegraphics[width=.45\textwidth]{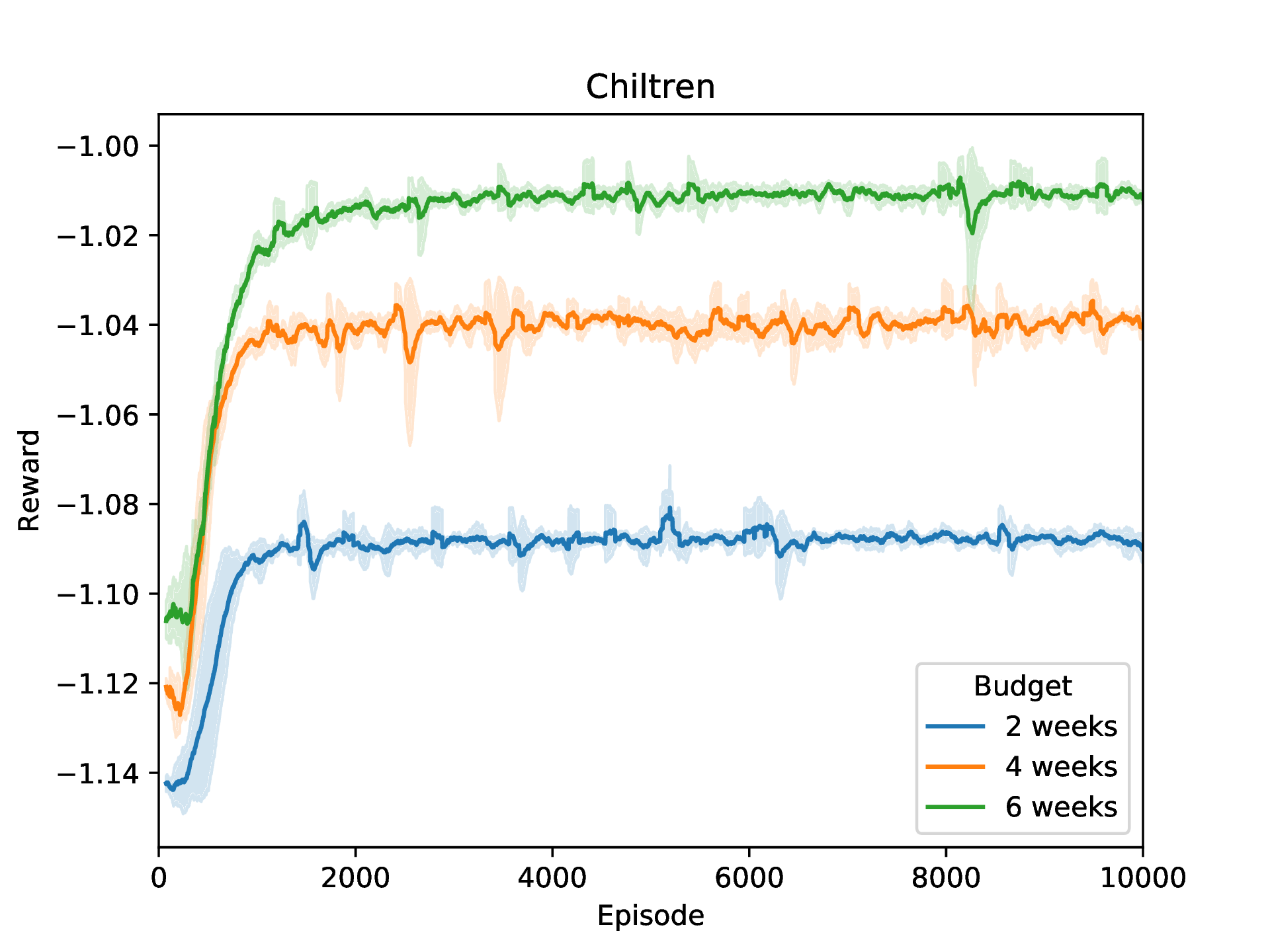}
	\end{minipage}
	\begin{minipage}{0.8\textwidth}
	\centering
	\includegraphics[width=.45\textwidth]{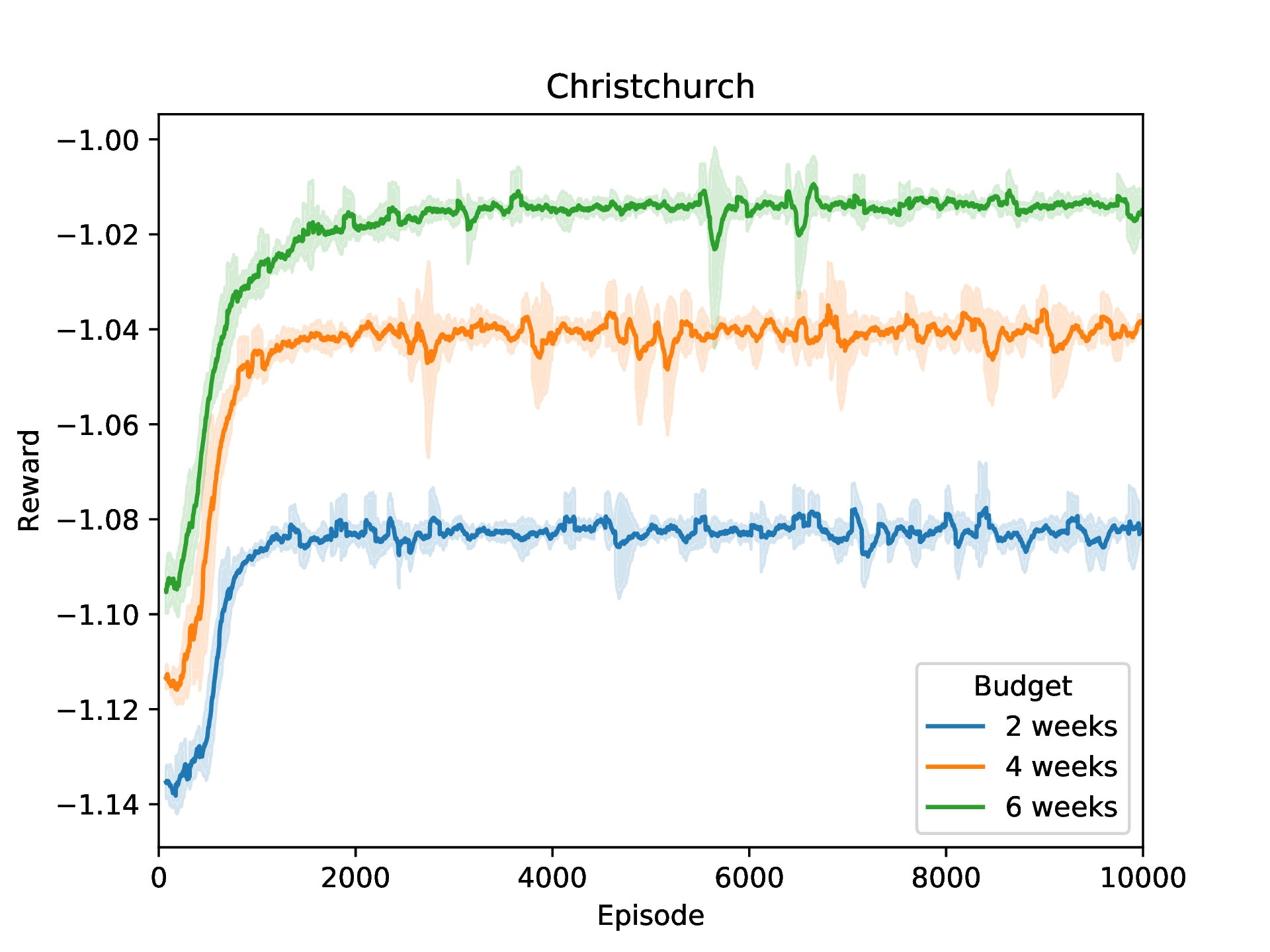}
	\includegraphics[width=.45\textwidth]{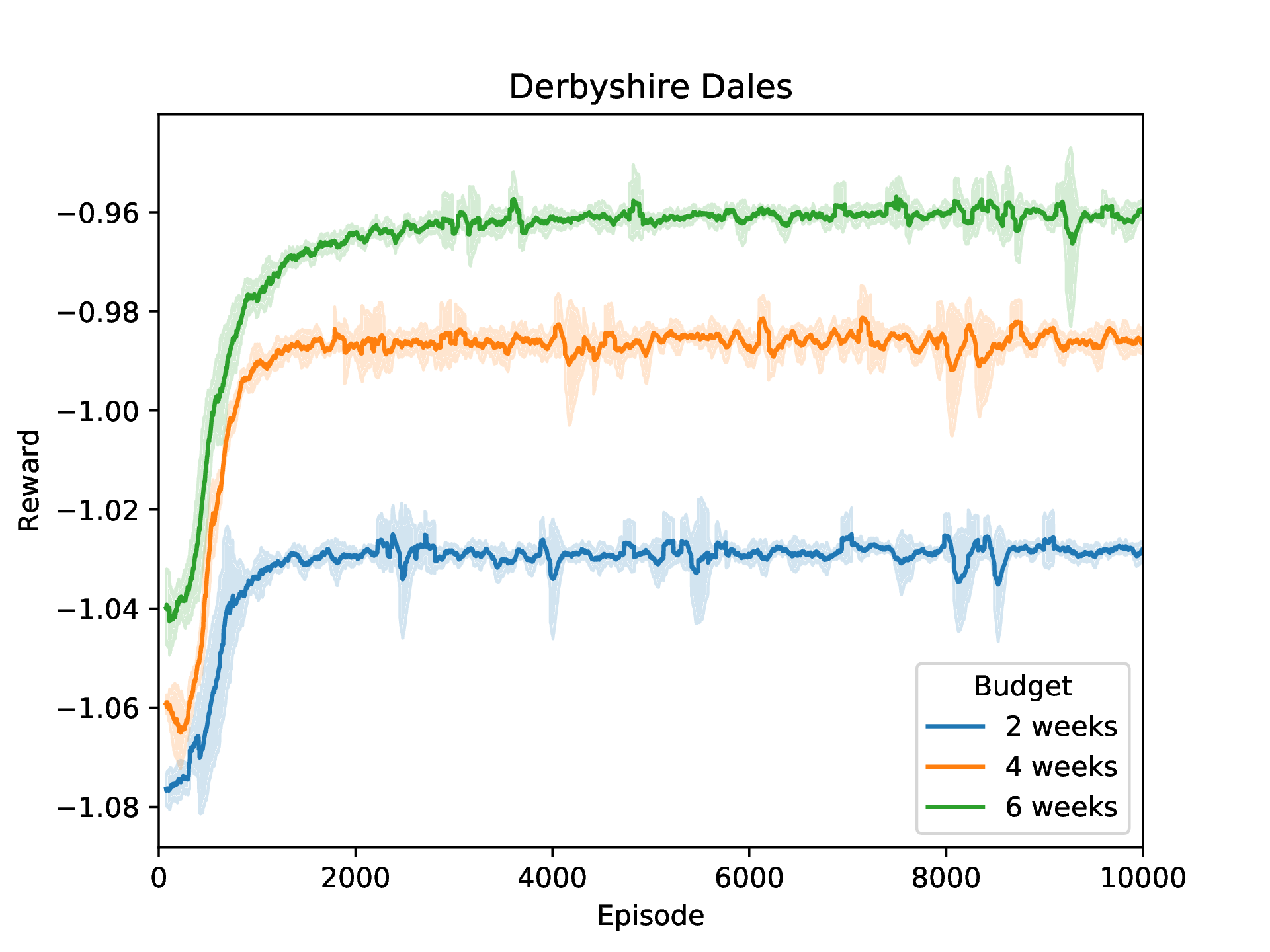}
	\end{minipage}
	\begin{minipage}{0.8\textwidth}
	\centering
	\includegraphics[width=.45\textwidth]{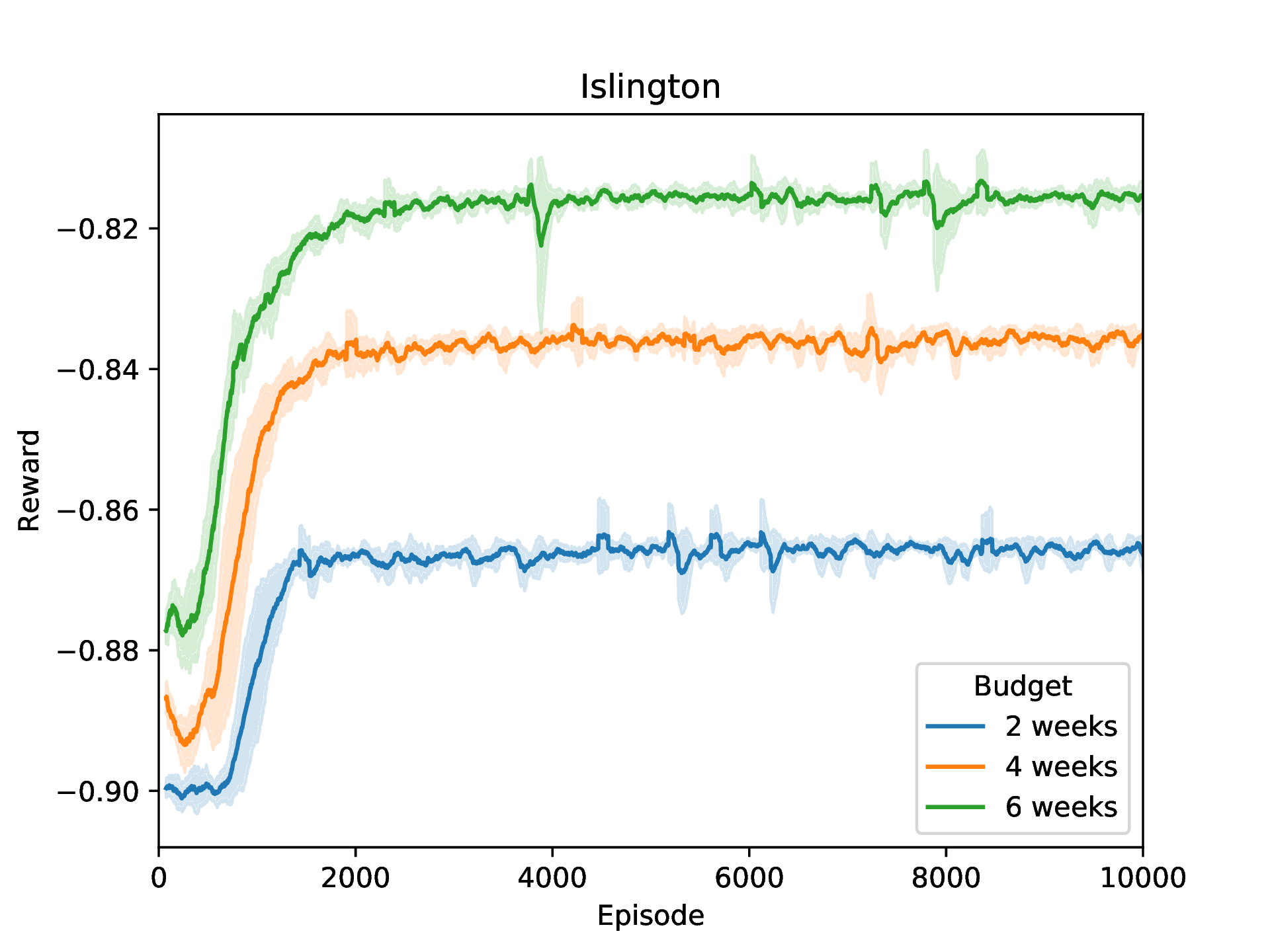}
	\includegraphics[width=.45\textwidth]{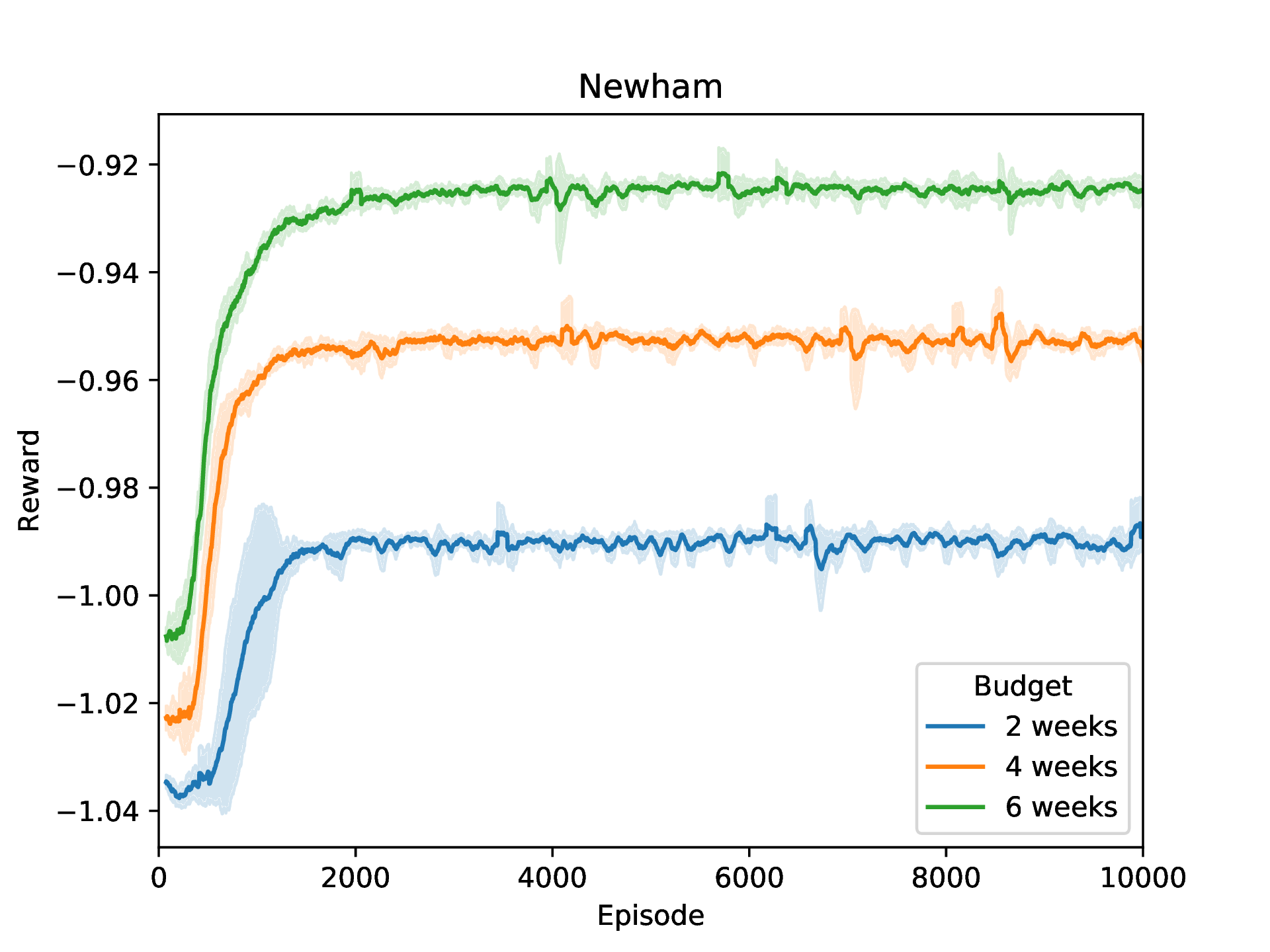}
	\end{minipage}
	\begin{minipage}{0.8\textwidth}
	\centering
	\includegraphics[width=.45\textwidth]{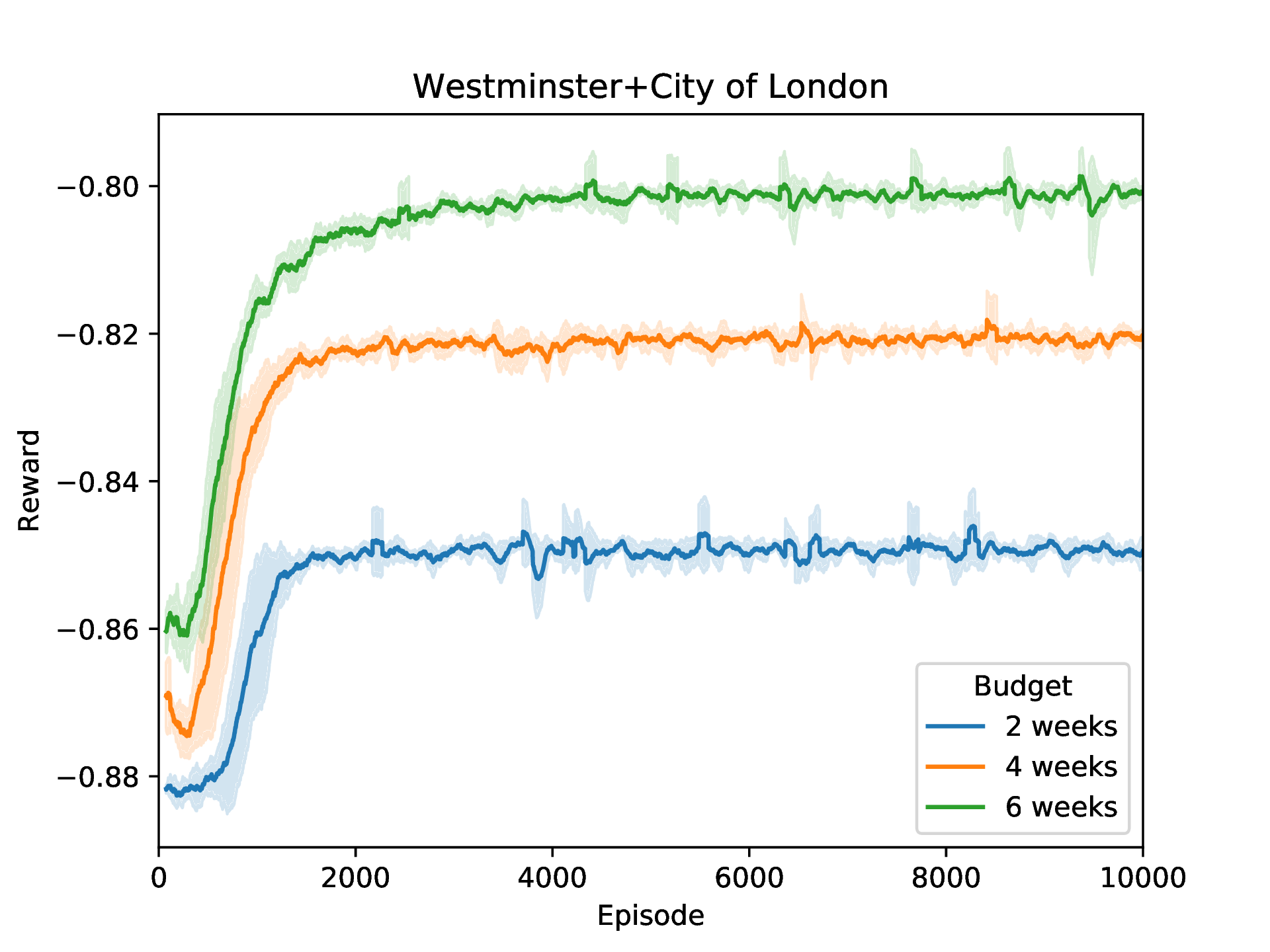}
	\includegraphics[width=.45\textwidth]{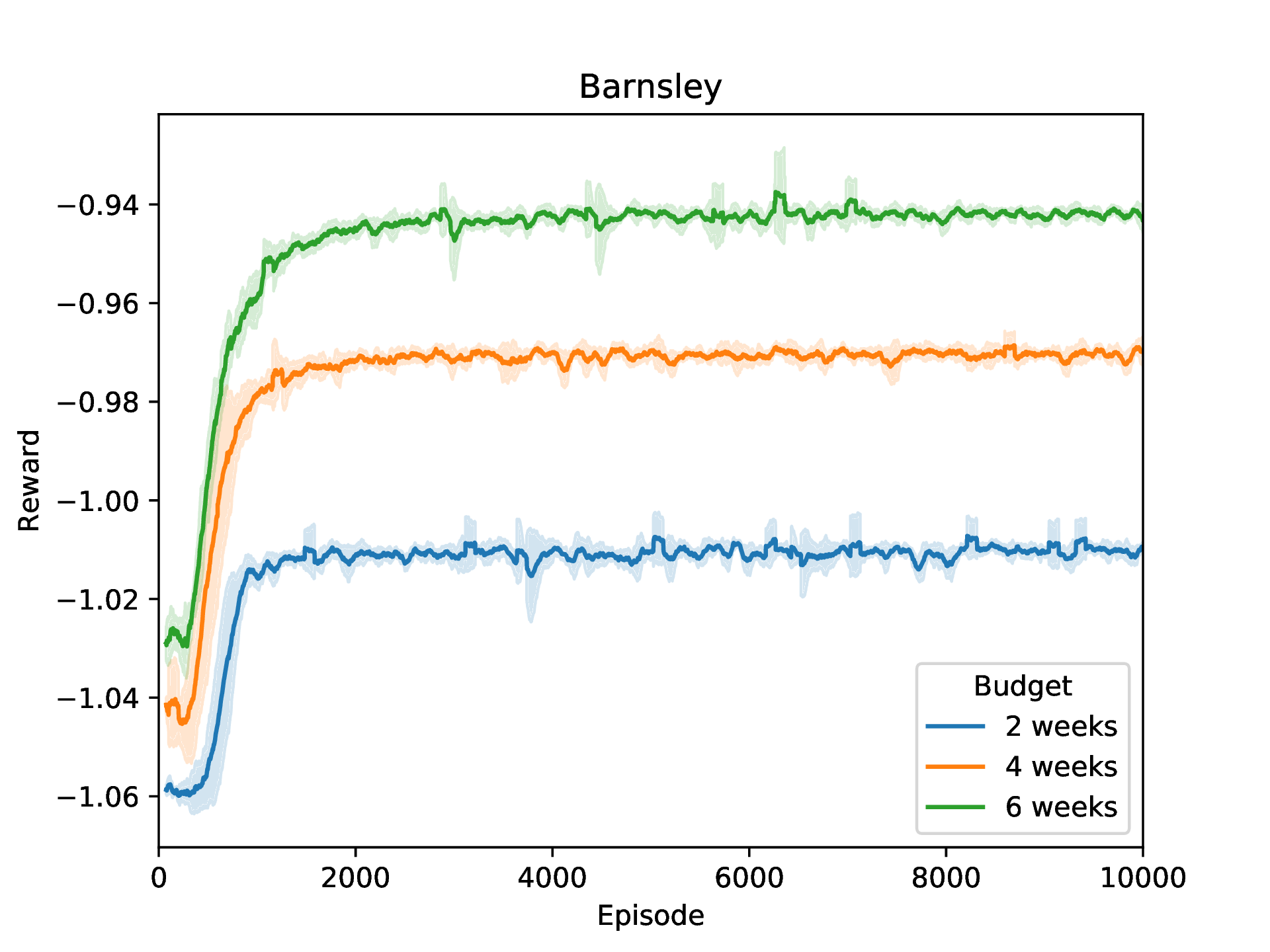}
	\end{minipage}
\caption{PPO learning curves for $R_0=2.4$.}
\end{figure}
%cd ~/research/spatial-pandemic-experiments/rl-chapter/spatial/census-experiment
%./compare_ppo_ground_truth.sh ar
\section{Comparing PPO to the ground truth (attack rate)}
\label{sec:apx_ar_ground_truth}
\begin{figure}[H]
	\begin{minipage}{0.9\textwidth}
	\centering
	\includegraphics[width=.45\textwidth]{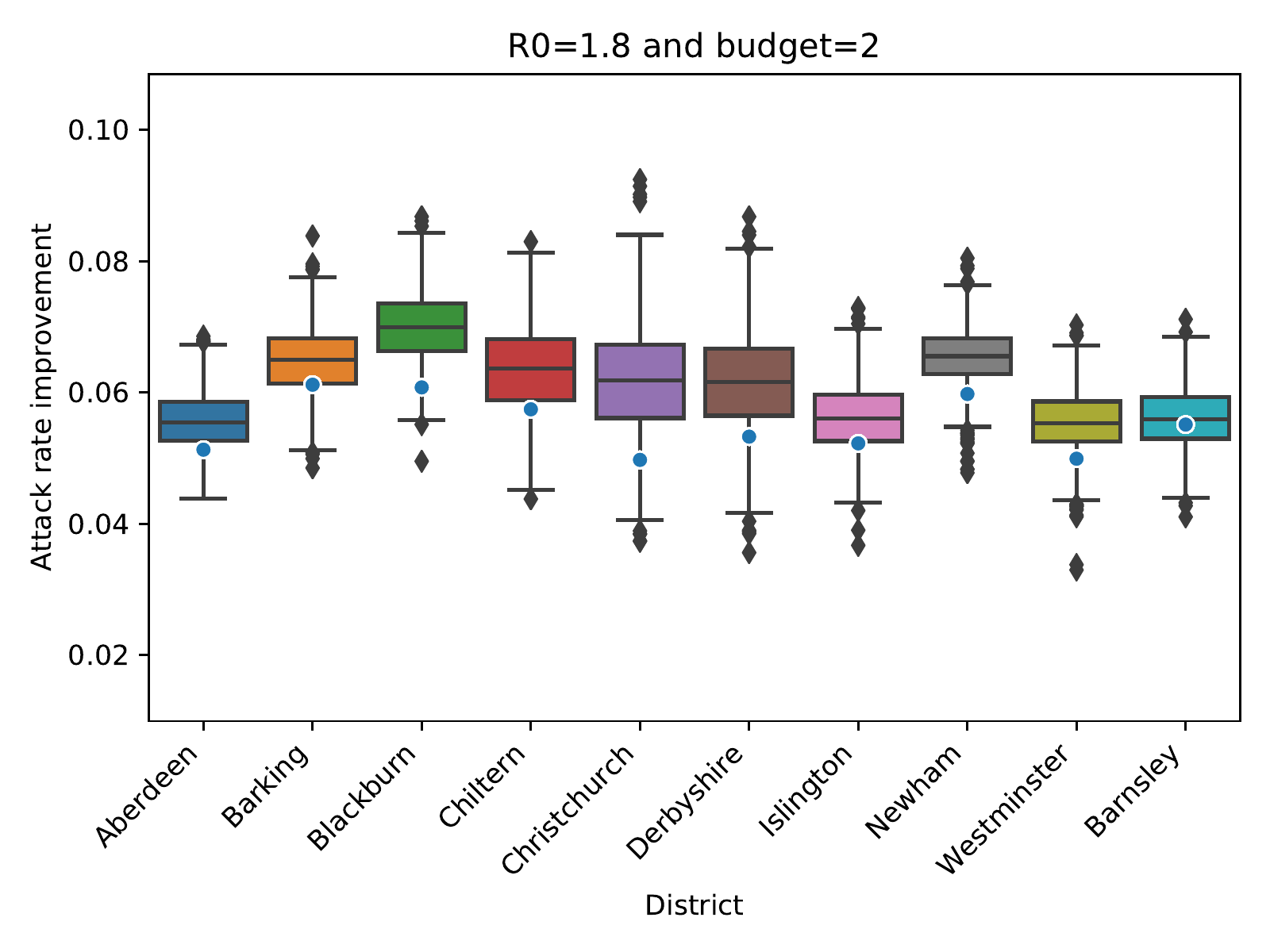}
	\includegraphics[width=.45\textwidth]{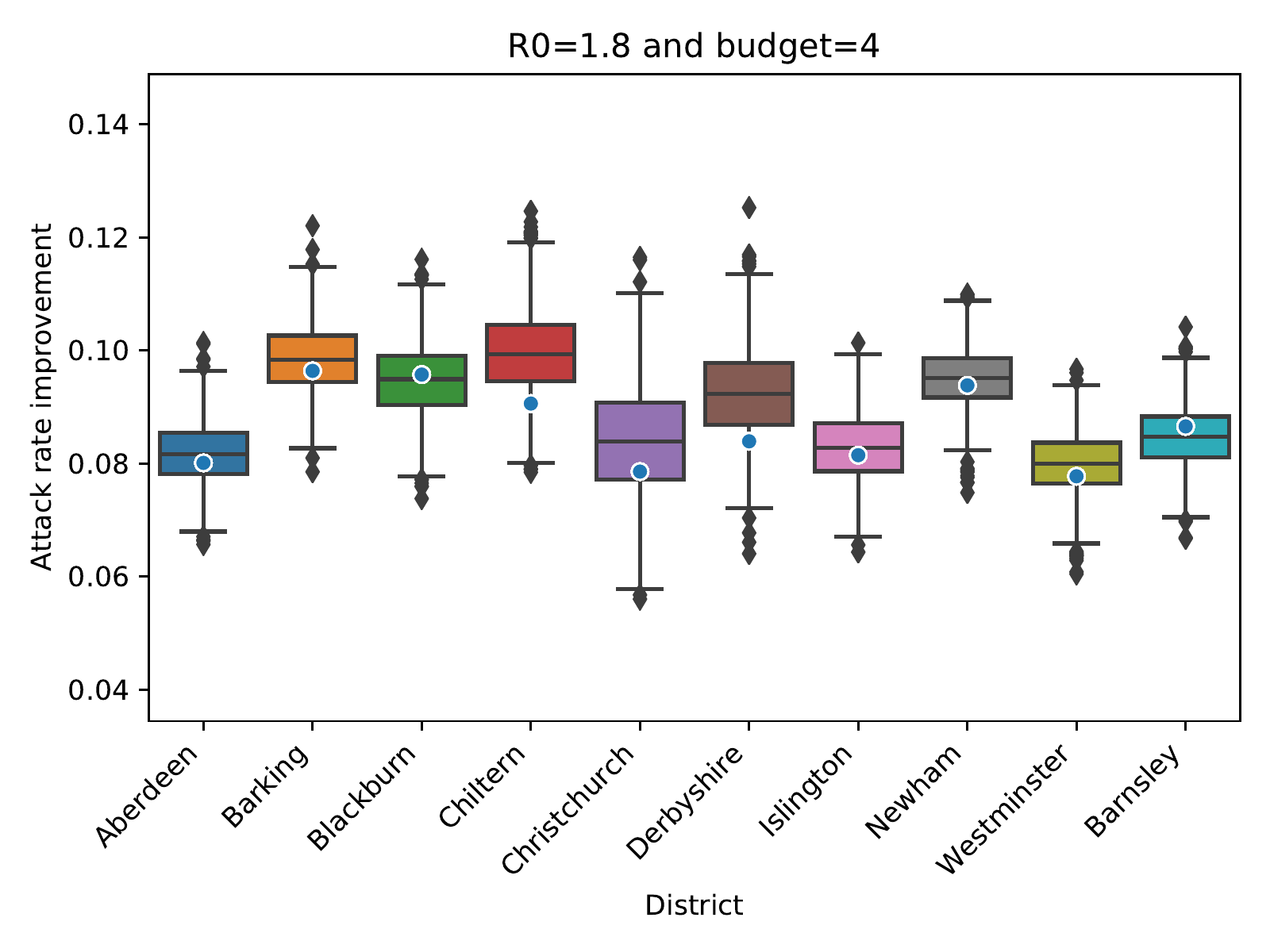}
	\end{minipage}
	\begin{minipage}{0.9\textwidth}
	\centering
	\includegraphics[width=.45\textwidth]{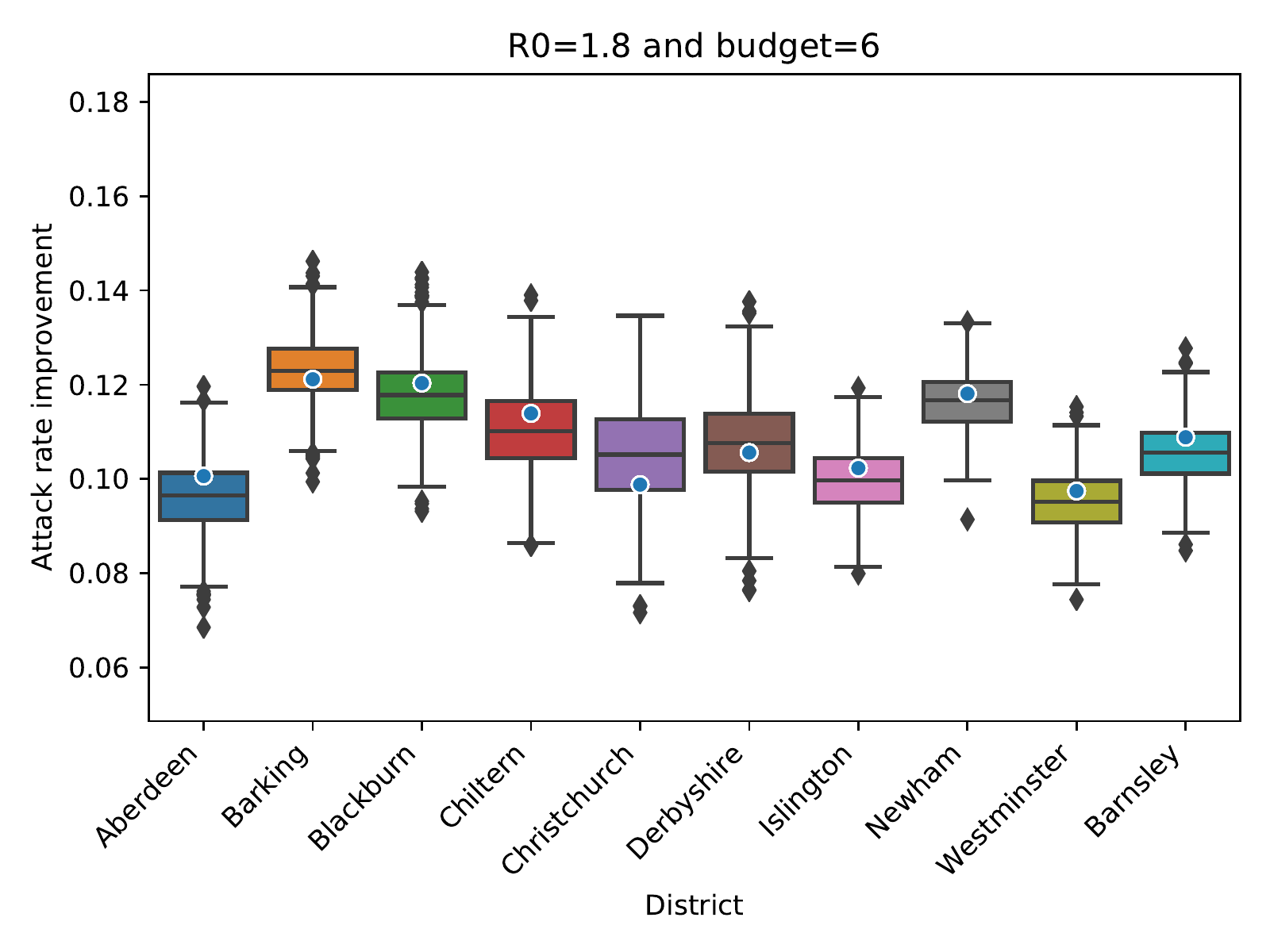}
	\includegraphics[width=.45\textwidth]{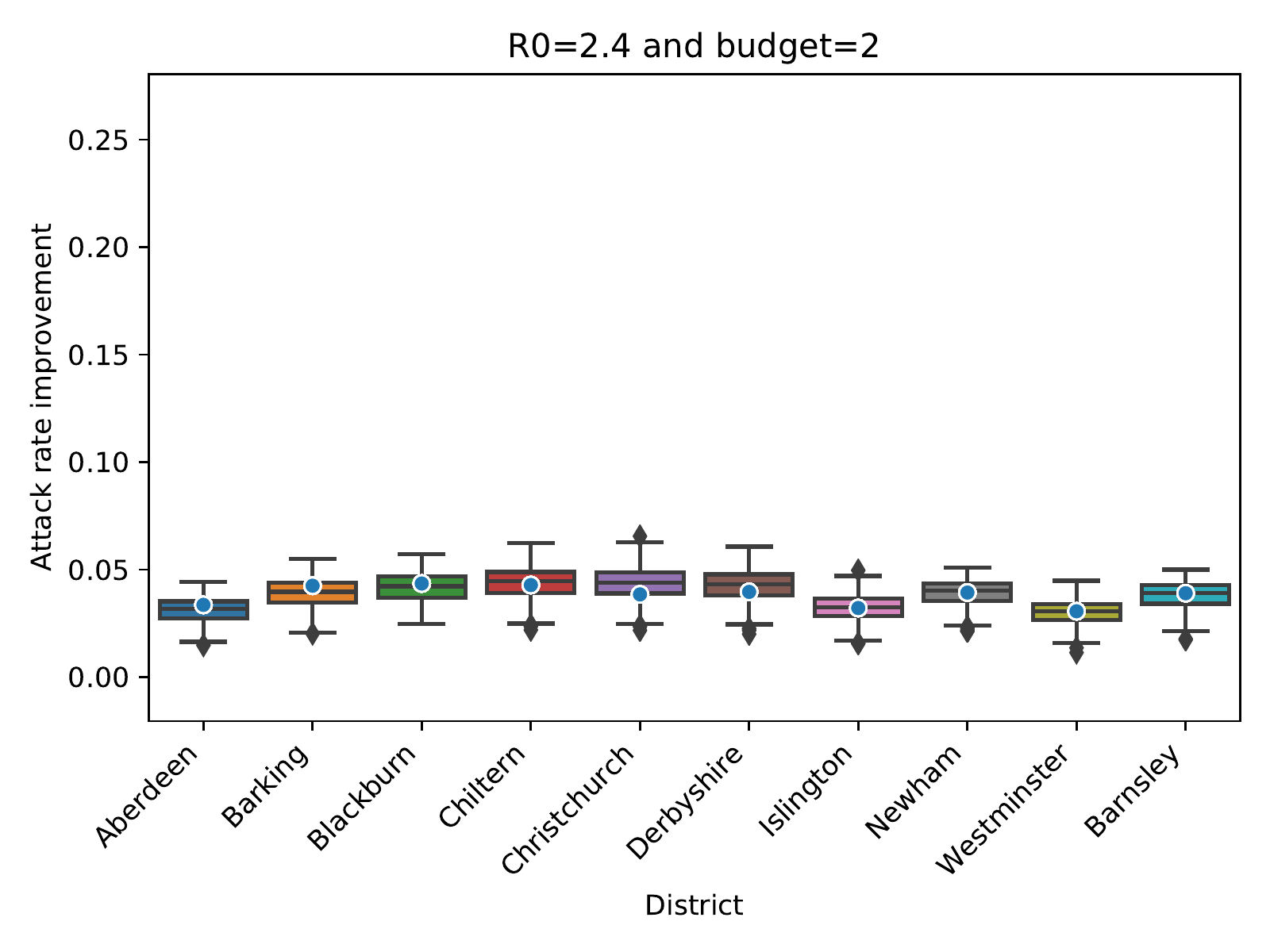}
	\end{minipage}
	\begin{minipage}{0.9\textwidth}
	\centering
	\includegraphics[width=.45\textwidth]{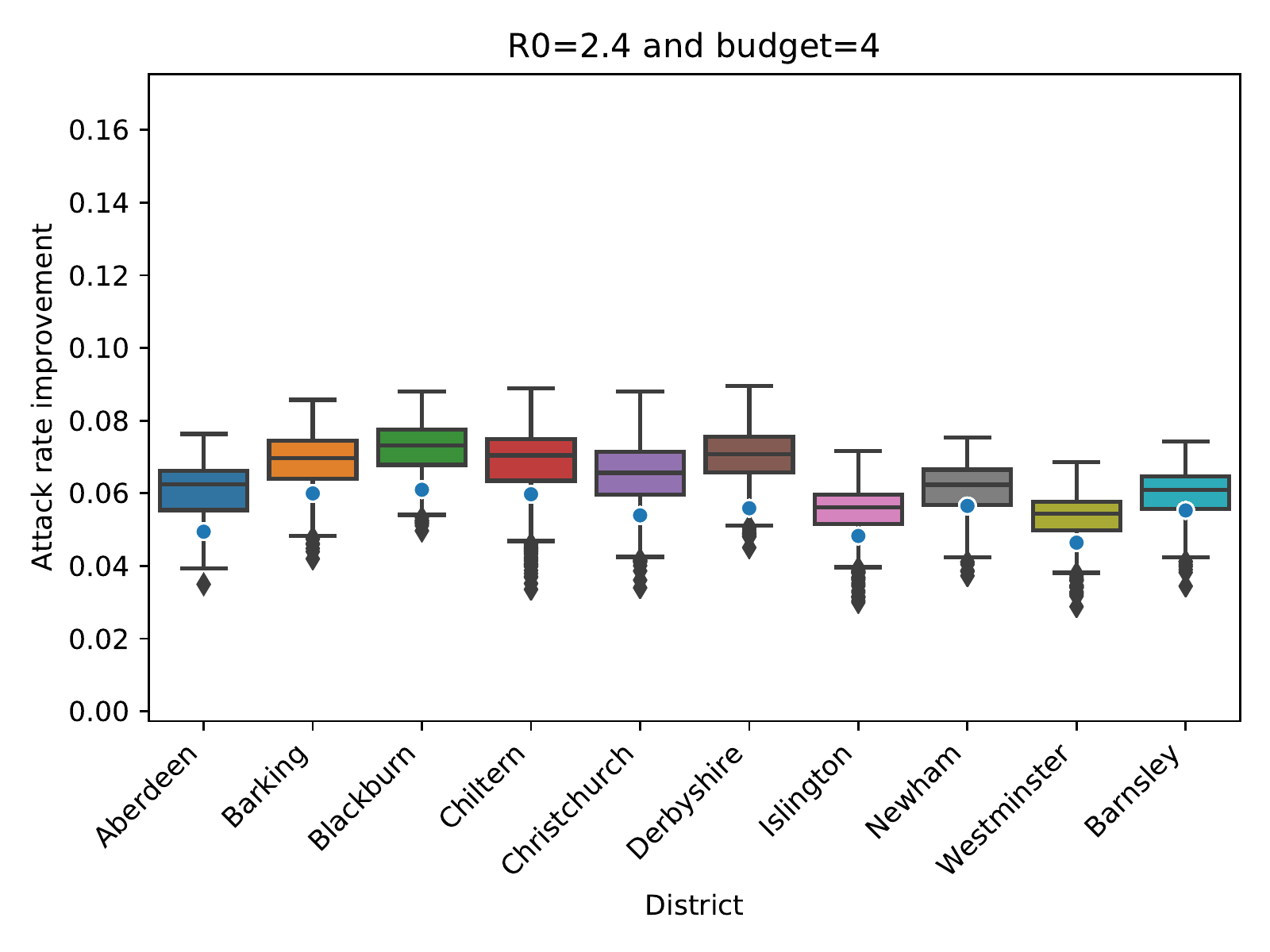}
	\includegraphics[width=.45\textwidth]{figures/spatial-rl/compositional-census/compare_ground_truth/compare_ar_24_6.pdf}
	\end{minipage}

\caption{Comparing PPO to the ground truth for $R_0 \in \{1.8,2.4\}$ and $\schoolclosurebudget \in \{2,4,6\}$.}
\end{figure}

\section{QMIX reward curves}
\label{sec:qmix_reward_curves}
\begin{figure}[H]
  \begin{minipage}{0.9\textwidth}
  \centering
  \includegraphics[width=.8\textwidth]{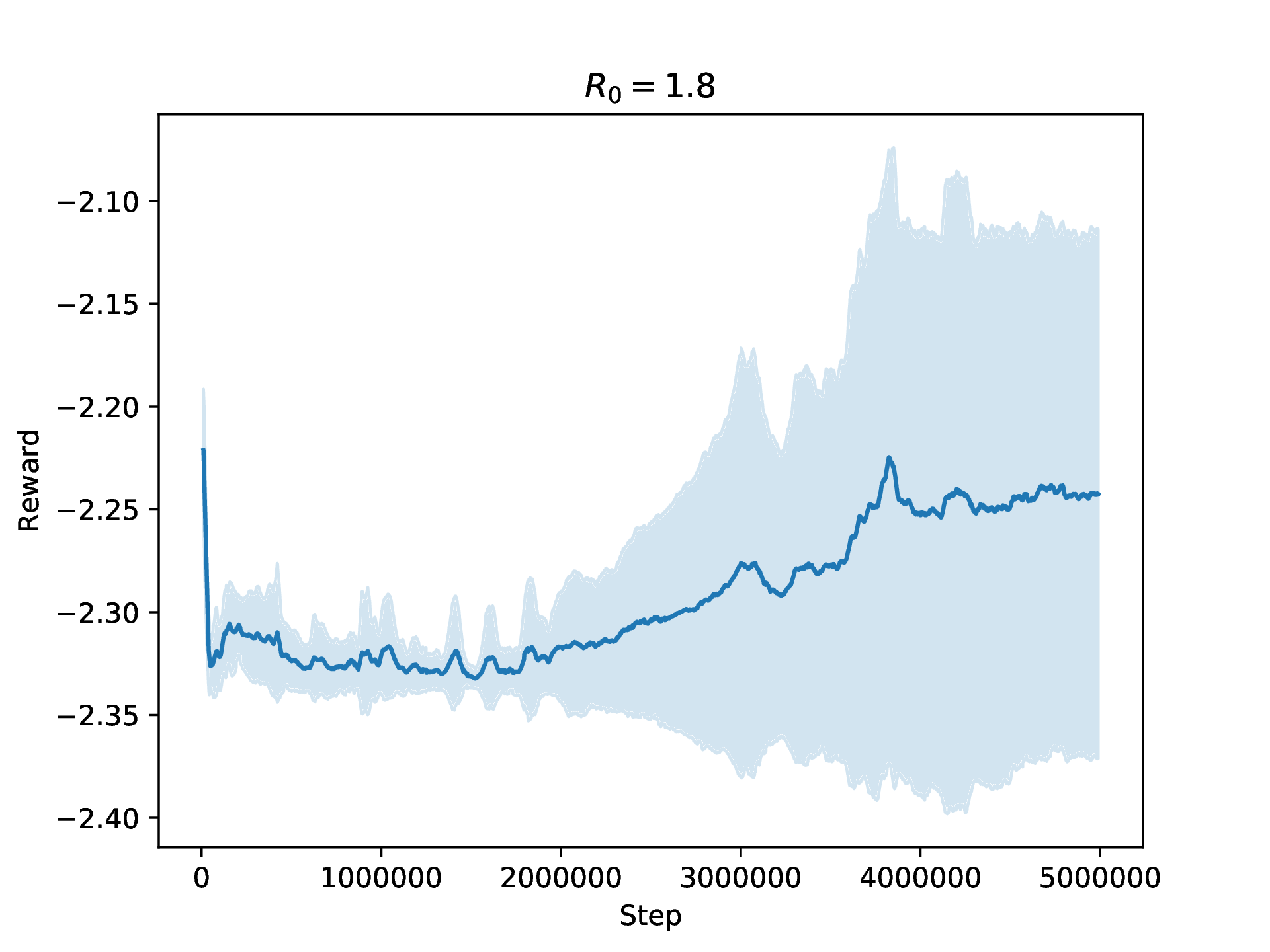}
  \includegraphics[width=.8\textwidth]{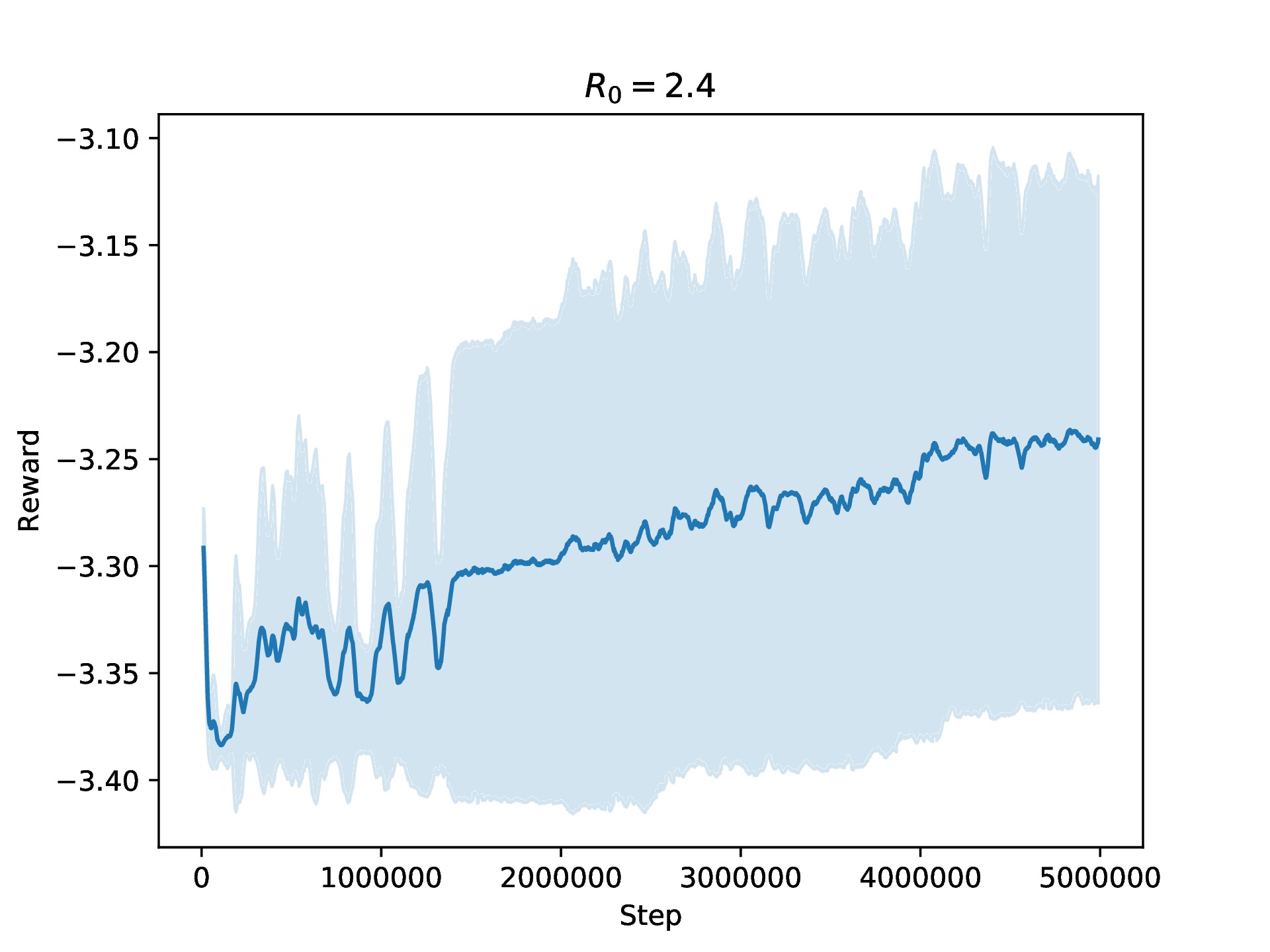}
  \end{minipage}
\caption{We show the reward curves for multi-district QMIX for $R_0=1.8$ (top panel) and $R_0=2.4$ (bottom panel). The shaded area shows the standard deviation of the reward signal, over 5 multi-district QMIX runs.}
\end{figure}

\newpage
\section{PPO hyper-parameters}
\label{sec:ppo_hyper_parameters}
\begin{itemize}
  \item Number of local steps: $1024$
  \item Batch size: $128$ (i.e., $8$ minibatches)
  \item Clipped Surrogate Objective epsilon $\epsilon$: $0.2$
  \item Number of epochs: $24$
  \item Entropy coefficient: $0.0059$
  \item \(\gamma\): \(0.99\)
  \item Generalized advantage estimation \(\lambda\): \(0.95\)
  \item Neural network of actor and critic:
  \begin{itemize}
    \item Number of hidden layers: $1$
    \item Number of units per hidden layer: $20$
    \item Non-linearity: hyperbolic tangent ($\mathrm{tanh}$)
    \item learning rate $\alpha$: $0.002$
    \item Optimizer: Adam~\cite{kingma2014adam}
    \item Gradient norm clipping threshold: $1.0$
  \end{itemize}
\end{itemize}

\newpage
\bibliographystyle{abbrv}  % do not change this line!
\bibliography{refs}  % put name of your .bib file here

\begin{thebibliography}{10}

\bibitem{aitchison1983principal}
J.~Aitchison.
\newblock Principal component analysis of compositional data.
\newblock {\em Biometrika}, 70(1):57--65, 1983.

\bibitem{aitchison1992criteria}
J.~Aitchison.
\newblock On criteria for measures of compositional difference.
\newblock {\em Mathematical Geology}, 24(4):365--379, 1992.

\bibitem{aitchison1994principles}
J.~Aitchison.
\newblock Principles of compositional data analysis.
\newblock {\em Lecture Notes-Monograph Series}, pages 73--81, 1994.

\bibitem{aitchison1997one}
J.~Aitchison and V.~Pawlowsky-Glahn.
\newblock The one-hour course in compositional data analysis or compositional
  data analysis is simple.
\newblock In {\em Proceedings of IAMG}, volume~97, pages 3--35, 1997.

\bibitem{allen2008construction}
E.~J. Allen, L.~J. Allen, A.~Arciniega, and P.~E. Greenwood.
\newblock Construction of equivalent stochastic differential equation models.
\newblock {\em Stochastic analysis and applications}, 26(2):274--297, 2008.

\bibitem{baguelin2010vaccination}
M.~Baguelin, A.~J. Van~Hoek, M.~Jit, S.~Flasche, P.~J. White, and W.~J.
  Edmunds.
\newblock Vaccination against pandemic influenza a/h1n1v in england: a
  real-time economic evaluation.
\newblock {\em Vaccine}, 28(12):2370--2384, 2010.

\bibitem{balcan2009seasonal}
D.~Balcan, H.~Hu, B.~Goncalves, P.~Bajardi, C.~Poletto, J.~J. Ramasco,
  D.~Paolotti, N.~Perra, M.~Tizzoni, W.~Van~den Broeck, et~al.
\newblock Seasonal transmission potential and activity peaks of the new
  influenza a (h1n1): a monte carlo likelihood analysis based on human
  mobility.
\newblock {\em BMC medicine}, 7(1):45, 2009.

\bibitem{barthelemy2010fluctuation}
M.~Barth{\'e}lemy, C.~Godreche, and J.-M. Luck.
\newblock Fluctuation effects in metapopulation models: percolation and
  pandemic threshold.
\newblock {\em Journal of theoretical biology}, 267(4):554--564, 2010.

\bibitem{Basta2009}
N.~E. Basta, D.~L. Chao, M.~E. Halloran, L.~Matrajt, and I.~M. Longini.
\newblock {Strategies for pandemic and seasonal influenza vaccination of
  schoolchildren in the United States}.
\newblock {\em American journal of epidemiology}, 170(6):679--686, 2009.

\bibitem{bertsekas2002introduction}
D.~P. Bertsekas and J.~N. Tsitsiklis.
\newblock {\em Introduction to probability}, volume~1.
\newblock Athena Scientific Belmont, MA, 2002.

\bibitem{brown2011would}
S.~T. Brown, J.~H. Tai, R.~R. Bailey, P.~C. Cooley, W.~D. Wheaton, M.~A.
  Potter, R.~E. Voorhees, M.~LeJeune, J.~J. Grefenstette, D.~S. Burke, et~al.
\newblock Would school closure for the 2009 h1n1 influenza epidemic have been
  worth the cost?: a computational simulation of pennsylvania.
\newblock {\em BMC public health}, 11(1):353, 2011.

\bibitem{ciavarella2016school}
C.~Ciavarella, L.~Fumanelli, S.~Merler, C.~Cattuto, and M.~Ajelli.
\newblock School closure policies at municipality level for mitigating
  influenza spread: a model-based evaluation.
\newblock {\em BMC infectious diseases}, 16(1):576, 2016.

\bibitem{cinlar2013introduction}
E.~Cinlar.
\newblock {\em Introduction to stochastic processes}.
\newblock Courier Corporation, 2013.

\bibitem{de2018impact}
G.~De~Luca, K.~Van~Kerckhove, P.~Coletti, C.~Poletto, N.~Bossuyt, N.~Hens, and
  V.~Colizza.
\newblock The impact of regular school closure on seasonal influenza epidemics:
  a data-driven spatial transmission model for belgium.
\newblock {\em BMC infectious diseases}, 18(1):29, 2018.

\bibitem{de2009preliminary}
U.~De~Silva, J.~Warachit, S.~Waicharoen, and M.~Chittaganpitch.
\newblock A preliminary analysis of the epidemiology of influenza a (h1n1) v
  virus infection in thailand from early outbreak data, june-july 2009.
\newblock {\em Eurosurveillance}, 14(31):19292, 2009.

\bibitem{diekmann2009construction}
O.~Diekmann, J.~Heesterbeek, and M.~Roberts.
\newblock The construction of next-generation matrices for compartmental
  epidemic models.
\newblock {\em Journal of the Royal Society Interface}, page rsif20090386,
  2009.

\bibitem{eames2012measured}
K.~T. Eames, N.~L. Tilston, E.~Brooks-Pollock, and W.~J. Edmunds.
\newblock Measured dynamic social contact patterns explain the spread of h1n1v
  influenza.
\newblock {\em PLoS computational biology}, 8(3):e1002425, 2012.

\bibitem{eggo2010spatial}
R.~M. Eggo, S.~Cauchemez, and N.~M. Ferguson.
\newblock Spatial dynamics of the 1918 influenza pandemic in england, wales and
  the united states.
\newblock {\em Journal of the Royal Society Interface}, 8(55):233--243, 2010.

\bibitem{ferguson2006strategies}
N.~M. Ferguson, D.~A. Cummings, C.~Fraser, J.~C. Cajka, P.~C. Cooley, and D.~S.
  Burke.
\newblock Strategies for mitigating an influenza pandemic.
\newblock {\em Nature}, 442(7101):448, 2006.

\bibitem{fraser2009pandemic}
C.~Fraser, C.~A. Donnelly, S.~Cauchemez, W.~P. Hanage, M.~D. Van~Kerkhove,
  T.~D. Hollingsworth, J.~Griffin, R.~F. Baggaley, H.~E. Jenkins, E.~J. Lyons,
  et~al.
\newblock Pandemic potential of a strain of influenza a (h1n1): early findings.
\newblock {\em science}, 324(5934):1557--1561, 2009.

\bibitem{fumanelli2012inferring}
L.~Fumanelli, M.~Ajelli, P.~Manfredi, A.~Vespignani, and S.~Merler.
\newblock Inferring the structure of social contacts from demographic data in
  the analysis of infectious diseases spread.
\newblock {\em PLoS computational biology}, 8(9):e1002673, 2012.

\bibitem{germann2019school}
T.~C. Germann, H.~Gao, M.~Gambhir, A.~Plummer, M.~Biggerstaff, C.~Reed, and
  A.~Uzicanin.
\newblock School dismissal as a pandemic influenza response: When, where and
  for how long?
\newblock {\em Epidemics}, page 100348, 2019.

\bibitem{gog2014spatial}
J.~R. Gog, S.~Ballesteros, C.~Viboud, L.~Simonsen, O.~N. Bjornstad, J.~Shaman,
  D.~L. Chao, F.~Khan, and B.~T. Grenfell.
\newblock Spatial transmission of 2009 pandemic influenza in the us.
\newblock {\em PLoS computational biology}, 10(6):e1003635, 2014.

\bibitem{gunning2019darpa}
D.~Gunning and D.~W. Aha.
\newblock Darpa's explainable artificial intelligence program.
\newblock {\em AI Magazine}, 40(2):44--58, 2019.

\bibitem{haber2007effectiveness}
M.~J. Haber, D.~K. Shay, X.~M. Davis, R.~Patel, X.~Jin, E.~Weintraub,
  E.~Orenstein, and W.~W. Thompson.
\newblock Effectiveness of interventions to reduce contact rates during a
  simulated influenza pandemic.
\newblock {\em Emerging infectious diseases}, 13(4):581, 2007.

\bibitem{halder2010developing}
N.~Halder, J.~K. Kelso, and G.~J. Milne.
\newblock Developing guidelines for school closure interventions to be used
  during a future influenza pandemic.
\newblock {\em BMC infectious diseases}, 10(1):221, 2010.

\bibitem{halloran2008modeling}
M.~E. Halloran, N.~M. Ferguson, S.~Eubank, I.~M. Longini, D.~A.~T. Cummings,
  B.~Lewis, S.~Xu, C.~Fraser, A.~Vullikanti, T.~C. Germann, and Others.
\newblock {Modeling targeted layered containment of an influenza pandemic in
  the United States}.
\newblock {\em Proceedings of the National Academy of Sciences},
  105(12):4639--4644, 2008.

\bibitem{hernandez2019survey}
P.~Hernandez-Leal, B.~Kartal, and M.~E. Taylor.
\newblock A survey and critique of multiagent deep reinforcement learning.
\newblock {\em Autonomous Agents and Multi-Agent Systems}, pages 1--48, 2019.

\bibitem{king2015avoidable}
A.~A. King, M.~Domenech~de Cell{\`e}s, F.~M. Magpantay, and P.~Rohani.
\newblock Avoidable errors in the modelling of outbreaks of emerging pathogens,
  with special reference to ebola.
\newblock {\em Proceedings of the Royal Society B: Biological Sciences},
  282(1806):20150347, 2015.

\bibitem{kingma2014adam}
D.~P. Kingma and J.~Ba.
\newblock Adam: A method for stochastic optimization.
\newblock {\em arXiv preprint arXiv:1412.6980}, 2014.

\bibitem{kissler2019geographic}
S.~M. Kissler, J.~R. Gog, C.~Viboud, V.~Charu, O.~N. Bj{\o}rnstad, L.~Simonsen,
  and B.~T. Grenfell.
\newblock Geographic transmission hubs of the 2009 influenza pandemic in the
  united states.
\newblock {\em Epidemics}, 26:86--94, 2019.

\bibitem{klepac2018contagion}
P.~Klepac, S.~Kissler, and J.~Gog.
\newblock Contagion! the bbc four pandemic--the model behind the documentary.
\newblock {\em Epidemics}, 2018.

\bibitem{kubiak20122009}
R.~J. Kubiak and A.~R. McLean.
\newblock Why was the 2009 influenza pandemic in england so small?
\newblock {\em PloS one}, 7(2):e30223, 2012.

\bibitem{lam2015numba}
S.~K. Lam, A.~Pitrou, and S.~Seibert.
\newblock Numba: A llvm-based python jit compiler.
\newblock In {\em Proceedings of the Second Workshop on the LLVM Compiler
  Infrastructure in HPC}, page~7. ACM, 2015.

\bibitem{leicht2008community}
E.~A. Leicht and M.~E. Newman.
\newblock Community structure in directed networks.
\newblock {\em Physical review letters}, 100(11):118703, 2008.

\bibitem{lewis2011network}
T.~G. Lewis.
\newblock {\em Network science: Theory and applications}.
\newblock John Wiley \& Sons, 2011.

\bibitem{liu2019multiagent}
Y.~Liu, W.~Wang, Y.~Hu, J.~Hao, X.~Chen, and Y.~Gao.
\newblock Multi-agent game abstraction via graph attention neural network.
\newblock {\em arXiv preprint arXiv:1810.09202}, 2019.

\bibitem{longini2005containing}
I.~M. Longini, A.~Nizam, S.~Xu, K.~Ungchusak, W.~Hanshaoworakul, D.~A.
  Cummings, and M.~E. Halloran.
\newblock Containing pandemic influenza at the source.
\newblock {\em Science}, 309(5737):1083--1087, 2005.

\bibitem{markel2007nonpharmaceutical}
H.~Markel, H.~B. Lipman, J.~A. Navarro, A.~Sloan, J.~R. Michalsen, A.~M. Stern,
  and M.~S. Cetron.
\newblock Nonpharmaceutical interventions implemented by us cities during the
  1918-1919 influenza pandemic.
\newblock {\em Jama}, 298(6):644--654, 2007.

\bibitem{Medlock2009}
J.~Medlock and A.~P. Galvani.
\newblock {Optimizing influenza vaccine distribution.}
\newblock {\em Science}, 325(5948):1705--1708, 2009.

\bibitem{miller2010incidence}
E.~Miller, K.~Hoschler, P.~Hardelid, E.~Stanford, N.~Andrews, and M.~Zambon.
\newblock Incidence of 2009 pandemic influenza a h1n1 infection in england: a
  cross-sectional serological study.
\newblock {\em The Lancet}, 375(9720):1100--1108, 2010.

\bibitem{mills2014spatial}
H.~L. Mills and S.~Riley.
\newblock The spatial resolution of epidemic peaks.
\newblock {\em PLoS computational biology}, 10(4):e1003561, 2014.

\bibitem{milne2013cost}
G.~J. Milne, N.~Halder, and J.~K. Kelso.
\newblock The cost effectiveness of pandemic influenza interventions: a
  pandemic severity based analysis.
\newblock {\em PloS one}, 8(4), 2013.

\bibitem{nishiura2010pros}
H.~Nishiura, G.~Chowell, M.~Safan, and C.~Castillo-Chavez.
\newblock Pros and cons of estimating the reproduction number from early
  epidemic growth rate of influenza a (h1n1) 2009.
\newblock {\em Theoretical Biology and Medical Modelling}, 7(1):1, 2010.

\bibitem{numpy2006}
T.~Oliphant.
\newblock {\em Guide to NumPy}.
\newblock 01 2006.

\bibitem{osaki2012applied}
S.~Osaki.
\newblock {\em Applied stochastic system modeling}.
\newblock Springer Science \& Business Media, 2012.

\bibitem{Paules2017}
C.~Paules and K.~Subbarao.
\newblock {Influenza}.
\newblock {\em The Lancet}, pages 697--708, 2017.

\bibitem{probert2019context}
W.~J. Probert, S.~Lakkur, C.~J. Fonnesbeck, K.~Shea, M.~C. Runge, M.~J.
  Tildesley, and M.~J. Ferrari.
\newblock Context matters: using reinforcement learning to develop
  human-readable, state-dependent outbreak response policies.
\newblock {\em Philosophical Transactions of the Royal Society B},
  374(1776):20180277, 2019.

\bibitem{radivojevic2017community}
M.~Radivojevi{\'c} and J.~Gruji{\'c}.
\newblock Community structure of copper supply networks in the prehistoric
  balkans: An independent evaluation of the archaeological record from the 7th
  to the 4th millennium bc.
\newblock {\em Journal of Complex Networks}, 6(1):106--124, 2017.

\bibitem{pmlr-v80-rashid18a}
T.~Rashid, M.~Samvelyan, C.~Schroeder, G.~Farquhar, J.~Foerster, and
  S.~Whiteson.
\newblock {QMIX}: Monotonic value function factorisation for deep multi-agent
  reinforcement learning.
\newblock In {\em Proceedings of the 35th International Conference on Machine
  Learning}, volume~80 of {\em Proceedings of Machine Learning Research}, pages
  4295--4304, Stockholmsmässan, Stockholm Sweden, 10--15 Jul 2018. PMLR.

\bibitem{schulman2017proximal}
J.~Schulman, F.~Wolski, P.~Dhariwal, A.~Radford, and O.~Klimov.
\newblock Proximal policy optimization algorithms.
\newblock {\em arXiv preprint arXiv:1707.06347}, 2017.

\bibitem{stein1987large}
M.~Stein.
\newblock Large sample properties of simulations using latin hypercube
  sampling.
\newblock {\em Technometrics}, 29(2):143--151, 1987.

\bibitem{tomba2008simple}
G.~S. Tomba and J.~Wallinga.
\newblock A simple explanation for the low impact of border control as a
  countermeasure to the spread of an infectious disease.
\newblock {\em Mathematical biosciences}, 214(1-2):70--72, 2008.

\bibitem{towers2012social}
S.~Towers and Z.~Feng.
\newblock Social contact patterns and control strategies for influenza in the
  elderly.
\newblock {\em Mathematical biosciences}, 240(2):241--249, 2012.

\bibitem{traag2019louvain}
V.~A. Traag, L.~Waltman, and N.~J. van Eck.
\newblock From louvain to leiden: guaranteeing well-connected communities.
\newblock {\em Scientific reports}, 9, 2019.

\bibitem{tuite2010estimated}
A.~R. Tuite, A.~L. Greer, M.~Whelan, A.-L. Winter, B.~Lee, P.~Yan, J.~Wu,
  S.~Moghadas, D.~Buckeridge, B.~Pourbohloul, et~al.
\newblock Estimated epidemiologic parameters and morbidity associated with
  pandemic h1n1 influenza.
\newblock {\em Can Med Assoc J}, 182(2):131--136, 2010.

\bibitem{vynnycky2010introduction}
E.~Vynnycky and R.~White.
\newblock {\em An introduction to infectious disease modelling}.
\newblock OUP oxford, 2010.

\bibitem{wang2018characterizing}
L.~Wang and J.~T. Wu.
\newblock Characterizing the dynamics underlying global spread of epidemics.
\newblock {\em Nature communications}, 9(1):218, 2018.

\bibitem{webby2003we}
R.~J. Webby and R.~G. Webster.
\newblock Are we ready for pandemic influenza?
\newblock {\em Science}, 302(5650):1519--1522, 2003.

\bibitem{yaesoubi2011dynamic}
R.~Yaesoubi and T.~Cohen.
\newblock Dynamic health policies for controlling the spread of emerging
  infections: influenza as an example.
\newblock {\em PloS one}, 6(9), 2011.

\bibitem{yaesoubi2013identifying}
R.~Yaesoubi and T.~Cohen.
\newblock Identifying dynamic tuberculosis case-finding policies for hiv/tb
  coepidemics.
\newblock {\em Proceedings of the National Academy of Sciences},
  110(23):9457--9462, 2013.

\bibitem{yaesoubi2016identifying}
R.~Yaesoubi and T.~Cohen.
\newblock Identifying cost-effective dynamic policies to control epidemics.
\newblock {\em Statistics in medicine}, 35(28):5189--5209, 2016.

\bibitem{yang2009transmissibility}
Y.~Yang, J.~D. Sugimoto, M.~E. Halloran, N.~E. Basta, D.~L. Chao, L.~Matrajt,
  G.~Potter, E.~Kenah, and I.~M. Longini.
\newblock The transmissibility and control of pandemic influenza a (h1n1)
  virus.
\newblock {\em Science}, 326(5953):729--733, 2009.

\bibitem{yu2018towards}
Y.~Yu.
\newblock Towards sample efficient reinforcement learning.
\newblock In {\em IJCAI}, pages 5739--5743, 2018.

\bibitem{zhu2020novel}
N.~Zhu, D.~Zhang, W.~Wang, X.~Li, B.~Yang, J.~Song, X.~Zhao, B.~Huang, W.~Shi,
  R.~Lu, et~al.
\newblock A novel coronavirus from patients with pneumonia in china, 2019.
\newblock {\em New England Journal of Medicine}, 2020.

\end{thebibliography}

\end{document}